\pdfoutput=1                    
\documentclass{article}

\usepackage{PRIMEarxiv}

\usepackage[utf8]{inputenc}     
\usepackage[T1]{fontenc}        
\usepackage{url}                
\usepackage{booktabs}           
\usepackage{amsmath}
\usepackage{amssymb}
\usepackage{amsfonts}
\usepackage{bm}
\usepackage{multirow}
\usepackage{tabularx}
\usepackage{graphicx}
\usepackage{subcaption}
\usepackage{algorithm}
\usepackage{algpseudocode}
\usepackage{float}
\usepackage{placeins}            
\usepackage{xcolor}
\usepackage{microtype}
\usepackage[numbers,sort&compress]{natbib}
\usepackage[colorlinks=true,linkcolor=blue,citecolor=blue,urlcolor=blue]{hyperref}

\graphicspath{{imgs/}{./}}

\newcommand{\rev}[1]{#1}
\newcommand{\revcolor}{}

\newcommand{\orcid}[1]{}

\def\eg{\emph{e.g.}} 
\def\etal{\emph{et al.}}

\title{Flexible Motion Generation \\ from Language and Style References
}

\author{
  Kai Weixian Lan$^{1,4}$ \quad Bodie Criswell$^{2,4}$ \quad Briana Fedkiw$^{3,4}$ \\
  \bfseries Zhan Zhang$^{1,4}$ \quad Joseph Teran$^{1,4}$ \quad Daniel Holden$^{4}$ \\[4pt]
  {\normalfont\normalsize $^1$University of California, Davis, USA \quad
   $^2$University of Utah, USA} \\
  {\normalfont\normalsize $^3$Brown University, USA \quad
   $^4$Epic Games, USA}
}

\begin{document}
\maketitle

\vspace*{-0.3in}
\setlength{\intextsep}{6pt plus 2pt minus 2pt}
\begin{figure}[H]
 \centering
 \includegraphics[width=0.63\linewidth]{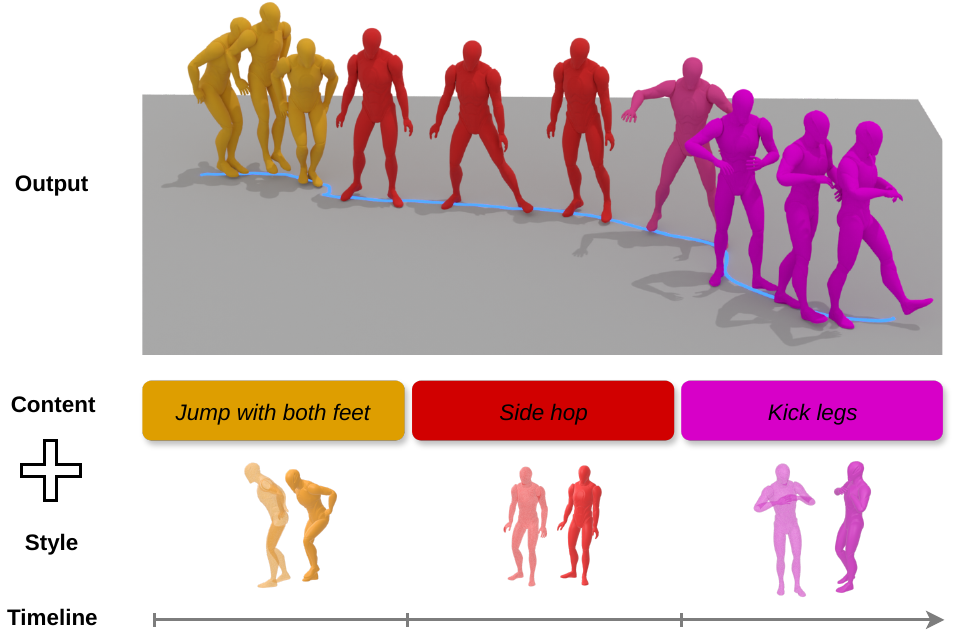}
 \caption{\textbf{FlexMoGen} generates long, stylized human motions conditioned on
time-varying text prompts and style reference clips.
Given three sequential content prompts
(``Jump with both feet'', ``Side hop'', ``Kick legs'')
and corresponding style references (bottom row), our framework synthesizes
a single temporally coherent motion sequence (top row) that faithfully
reflects both the intended action and the target style within each segment.
Color coding indicates the correspondence between style references,
content prompts, and generated frames.}
 \label{fig:teaser}
\end{figure}
\setlength{\intextsep}{12pt plus 2pt minus 2pt}   

\begingroup
\setlength{\leftmargini}{1em}
\begin{abstract}
We introduce \textbf{FlexMoGen}, a novel framework for flexible human motion synthesis conditioned on both natural language descriptions and motion style references.
Text prompts are effective at defining semantic content, but they are often limited in capturing fine-grained style details such as timing, limb articulation, and expressive dynamics.
A style example clip supplements the text by conveying these nuanced motion characteristics directly, enabling the model to preserve high-level intent while reproducing the desired stylistic traits.
Given a text prompt and a style example clip, \textbf{FlexMoGen} generates high-quality motions that preserve semantic content while faithfully reflecting the target style, offering users greater control over the animation generation process.
Unlike prior methods that rely on discrete style labels and do not generalize to long or multi-style generation, \textbf{FlexMoGen} learns a variational style encoder without style supervision and supports long, time-varying, multi-style synthesis.
Our framework jointly pre-trains the style encoder and a text-to-motion latent diffusion model within a unified architecture, modulating motion style through a lightweight adaptation module. It integrates an efficient relative positional encoding scheme and is trained on both stylized and non-stylized datasets, enabling strong generalization to unseen text--style combinations.
\rev{Experiments show that \textbf{FlexMoGen} achieves the best balance between content fidelity and style reflection.}
\end{abstract}

\begin{quote}
\end{quote}
\endgroup

\clearpage

\section{Introduction}
\label{sec:intro}

Character animation is a fundamental component of modern computer graphics applications. Traditional methods, such as hand animation via key-framing or using motion capture to capture motions from real subjects, are both costly and time-consuming. Data-driven character motion synthesis has emerged as a powerful alternative, enabling the automatic generation of animations. Recent advances demonstrate that animation can be generated from text prompts, style examples, trajectories, and other control signals.

For text-conditioned generation, generative models have become widely used. Early approaches to motion generation from text descriptions employed variational autoencoders (VAEs) \cite{kingma2013auto}, as seen in works such as Temos~\cite{petrovich2022temos} and text2motion~\cite{Guo_2022_CVPR}. More recently, diffusion models \cite{ho2020denoising, song2020denoising} have gained popularity for generating diverse human motions, with several studies adopting this framework \cite{tevet2023human, chen2023executing, dabral2023mofusion, zhang2024motiondiffuse}.
Another active area in animation is motion stylization, which focuses on transferring style from one motion to another \cite{aberman2020unpaired, Jang_2022, tao2022style, mason2022local, xia:2015:style}.

In this work, we focus on integrating content, represented by text prompts, and style, represented by example clips, to generate novel human motions, extending the formulation introduced by SMooDi~\cite{zhong2024smoodi}.
In this formulation, text prompts define the semantic content of the action (\eg, \textit{walking}, \textit{running}, or \textit{dancing}), while style captures the expressive characteristics of the motion, such as \textit{happy}, \textit{sad}, \textit{energetic}, or \textit{lethargic}.
Text alone is often insufficient to express the full range of stylistic nuances in motion; descriptors like rhythm, energy, and joint-level articulation can be ambiguous or hard to specify precisely.
A motion style reference clip provides these details directly, making it easier to communicate how a motion should look and feel while the text prompt preserves the high-level semantic intent.
Although one could, in principle, describe both content and style using natural language and input them into a text-to-motion (T2M) model, the nuanced and highly personalized nature of style is difficult to fully capture with text alone. Consistent with the findings of SMooDi, we also adopt motion examples as a more expressive representation of style.
A straightforward solution for text- and style-conditioned motion generation is to first synthesize the content using a T2M model and then apply style transfer as a post-processing step. However, existing style-transfer methods~\cite{mason2022local, aberman2020unpaired, xia:2015:style} are typically trained only on stylized datasets, which are limited in diversity and cannot support a broad range of content motions. In addition, a two-stage pipeline often introduces temporal inconsistencies and artifacts.
To overcome these drawbacks, SMooDi~\cite{zhong2024smoodi} introduces the idea of adding style control directly into a pretrained T2M model via ControlNet~\cite{zhang2023adding}.
Building on this formulation, subsequent works such as Sawdayee \etal~\cite{sawdayee2025dance} and Wu \etal~\cite{10.1145/3721238.3730641} also integrate example-based style conditioning into pretrained T2M models.

Although these works employed new techniques to add style control to a T2M model, common drawbacks are as follows. Their approach is limited to short animation sequences and offers limited flexibility in controlling the generated motions. Additionally, \rev{most of these frameworks require} explicit style class labels, which are often unavailable in practice\rev{, the exception being Wu \etal~\cite{10.1145/3721238.3730641}, who also learn style without labels but still target short, single-style sequences}.
To address these limitations, we propose a model that supports separate control signals and style examples for different segments of an animation. Our method employs a style encoder trained without reliance on explicit style class annotations. To enable the synthesis of temporally coherent and arbitrarily long motion sequences, we introduce a sliding-window transformer architecture inspired by relative position encoding from language modeling \cite{shaw-etal-2018-self, wennberg-henter-2021-case, su2024roformer} and music generation \cite{huang_music_transformer}. This architecture facilitates smooth state transitions and fine-grained temporal control over both semantics and style, as well as generating longer sequences.
Furthermore, drawing inspiration from skeleton-aware networks \cite{Jang_2022, park2021diverse, kim2024most}, we incorporate per-body-part style embeddings into our encoder to disentangle localized stylistic features. This enhances style transfer fidelity and enables targeted manipulation of individual body parts.

\rev{We demonstrate that our method advances the state of the art in flexibility and versatility, and achieves the best balance between content fidelity and style reflection.} Furthermore, we provide comprehensive supplementary videos and materials to substantiate the effectiveness of our approach.
Our main contributions are:
\begin{itemize}
\item A variational style encoder that learns rich motion style representations from example clips without requiring any explicit style class labels.
\item A lightweight style adaptation module implemented as learnable bias vectors injected into the key and value projections of attention layers, enabling efficient and effective modulation of motion style.
\item Temporal attention mechanisms that provide fine-grained temporal control, supporting temporally varying semantics and styles, and enabling the generation of motions significantly longer than those seen during training.
\end{itemize}

\section{Related work}
\label{sec:rel_work}

\subsection{Text-conditioned human motion generation}

Previously, many works~\cite{petrovich2022temos, Guo_2022_CVPR} relied on variational autoencoders (VAEs) for generating motions from text prompts.
Meanwhile, methods such as MotionCLIP~\cite{tevet2022motionclip} and TEMOS~\cite{petrovich2022temos} focused on aligning the text embedding space with the motion space to facilitate motion generation and retrieval.
With the advent of diffusion-based models\cite{ho2020denoising}, text-to-motion generation has seen rapid progress, starting with models like MDM~\cite{tevet2023human}, FLAME~\cite{kim2023flame}, and MotionDiffuse~\cite{zhang2024motiondiffuse}.
Since then, further improvements have been introduced. MoFusion~\cite{dabral2023mofusion}, for example, proposed a new diffusion framework capable of incorporating diverse control signals, such as music.
Subsequent works, including MLD~\cite{chen2023executing}, encode motions into a latent space using a VAE and train diffusion models directly in this compact representation.
Other approaches, such as T2M-GPT~\cite{zhang2023generating} and MoMask~\cite{guo2024momask}, utilize discrete motion representations obtained via VQ-VAE to improve robustness and sequence modeling.
Additionally, MotionGPT~\cite{jiang2023motiongpt} expands the scope of motion-related tasks, supporting not only text-driven motion generation but also motion captioning and motion in-betweening, further illustrating the growing versatility of unified motion–language models.
\rev{Work on the backbone itself~\cite{meng2024rethinking} is complementary to ours, since our backbone
stays frozen and a stronger text-to-motion model can be substituted without touching the style pathway.}

\rev{A separate line targets sequences longer than a single prompt whose content changes over time.
TEACH~\cite{athanasiou2022teach} generates a list of actions one after another,
PriorMDM~\cite{shafir2024human} reuses a short-clip model as a prior and denoises the transition
intervals in a second pass, and FlowMDM~\cite{barquero2024seamless} blends absolute and relative
positional encodings to remove the seams. Multi-Track Timeline Control~\cite{petrovich2024multi} extends
the single ordered list into parallel tracks, so overlapping text intervals can drive different body
parts. All schedule \emph{content} along a timeline; we take the same view but schedule \emph{style}
along it, per body part as well as over time.}

\subsection{Human motion stylization}
Hsu \etal~\cite{hsu:2005:style} was one of the first works to examine stylization in character animation by aligning motions and a linear model for stylistic differences, which allows for real-time translation of styles.
Ma \etal~\cite{ma:2010:style} further decomposed motions into style and variation components, dividing the character rig into joint groups associated with latent variation parameters modeled by a Bayesian network.
Subsequent works, such as \cite{8013475, 10.1145/2897824.2925975}, leveraged Gram matrices \cite{7780634} to transfer motion style by measuring correlations between different feature maps. With advances in machine learning, increasingly diverse techniques have emerged for motion style transfer.
Adaptive Instance Normalization (AdaIN)~\cite{huang2017arbitrary}, initially proposed for image style transfer, has gained popularity for motion style transfer as well~\cite{aberman2020unpaired, Jang_2022, guo2024generative, tang2024decoupling}. AdaIN disentangles content and style by aligning the mean and variance of the content feature map with those of the style feature map. Adversarial networks have also been employed for unpaired style transfer~\cite{aberman2020unpaired, tao2022style}.
Several recent approaches leverage the power of pretrained diffusion models for motion style transfer~\cite{chen2024pay, raab2024monkey, guo2024generative}. ZeroEGGS~\cite{ghorbani2022zeroeggs} generates stylized gestures from speech in a zero-shot manner using a VAE. Time-series models~\cite{mason2022local, tao2022style, xia:2015:style, jang2023mocha} enable real-time style transfer. Some autoregressive models are based on generative flows~\cite{9578056}, and more recently diffusion-based methods, such as SinMDM~\cite{raab2024single}, use diffusion models as priors for single-instance learning. Additional works~\cite{hu2024diffusion, song2024style, zhang2024generative, guo2024generative} further exploit the latent space of motion and diffusion models to enhance style extraction.
The latest research focuses on stylized text-to-motion, where models generate stylized motions guided by textual descriptions. Recent works, including Zhong \etal~\cite{zhong2024smoodi}, Sawdayee \etal~\cite{sawdayee2025dance}, and Wu \etal~\cite{10.1145/3721238.3730641}, extend pretrained text-to-motion diffusion models with additional style control using techniques such as ControlNet~\cite{zhang2023adding}, LoRA~\cite{hu2022lora}, or custom style adapters.
\rev{Among these, Wu \etal~\cite{10.1145/3721238.3730641} also learn style from reference clips
without class labels, injecting it with their Semantic-Aware Style Injection (SASI) module, which
uses the text prompt as a semantic mediator between the content and reference frames.
}
\rev{StyleMotif~\cite{guo2025stylemotif} likewise conditions on a reference rather than a label, and
broadens it to several modalities, but its style feature carries no time index and is fused once at a
single denoising block, so a reference conditions the sequence as a whole and multiple references are
interpolated uniformly rather than in sequence. Our style code is resolved along time and per body part,
which is what makes Sections~\ref{subsec:time-varying_style} and~\ref{sec:bodypart_mixing} possible.
The move away from discrete labels also appears in adjacent domains: Lyu
\etal~\cite{lyu2025tempo} replace genre labels in music-driven dance with a continuous tempo cue, on
the same grounds that lead us to avoid style labels --- they are coarse where they exist and usually
absent.}

\section{Method}
\label{sec:method}

\begin{figure}[t]
  \centering
   \includegraphics[width=\textwidth]{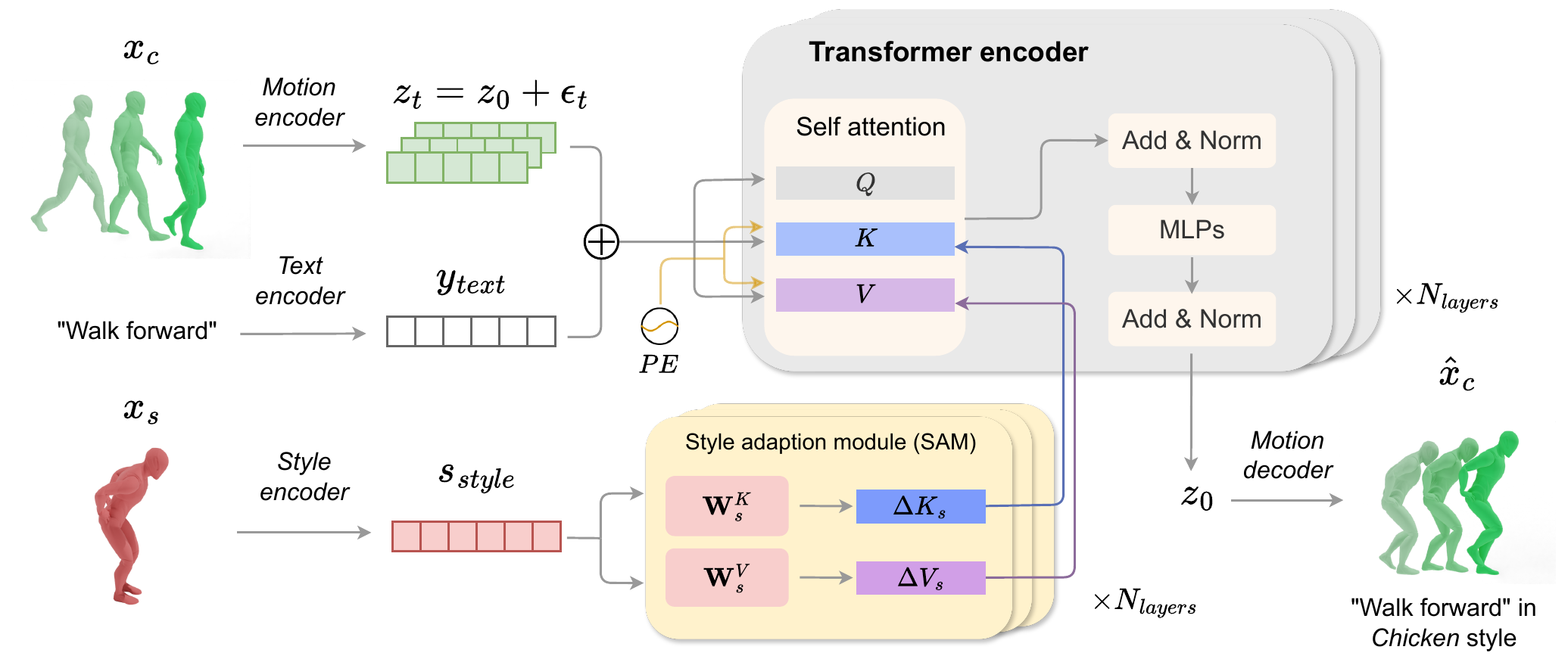}
   \caption{
    Overview of our style-adapted text-to-motion framework.
    The input content motion $x_c$ is encoded by a pretrained motion encoder, and Gaussian noise is added to its latent embedding.
    The text-to-motion (T2M) model (top right) consists of $7$ Transformer encoder layers that take motion, text, and time embeddings as inputs.
    Each self-attention block incorporates relative positional encoding into the key and value projections.
    During style finetuning, the T2M backbone is frozen, and only the weight matrices $W_s^{K}$ and $W_s^{V}$ in the Style Adaptation Module (SAM) are trained.
    SAM encodes a style example $x_s$ and injects its outputs as additive biases to the $K$ and $V$ vectors in the attention layers, guiding the network to preserve the semantics of $x_c$ while reflecting the style of $x_s$.
    }
   \label{fig:controlnet}
\end{figure}
In this section, we describe the core components of our framework. We first provide an overview, emphasizing the Style Adaptation Module (SAM), which injects style embeddings into a pretrained text-to-motion (T2M) diffusion model as learnable key and value biases in the attention layers. We then detail the pretraining of the style encoder and the T2M model, including architectural choices such as combining relative and global positional encodings to enable fine-grained temporal control over the generated motions.

\subsection{Framework overview}

Our framework, illustrated in Figure~\ref{fig:controlnet}, aims to generate motions that follow the semantics of a text prompt (\eg, \textit{walk forward}) while reflecting the style of a reference clip (\eg, \textit{Chicken} style). To achieve this, we pre-train a text-to-motion (T2M) diffusion model and a variational style encoder, and introduce a lightweight Style Adaptation Module (SAM) that injects style embeddings as key and value biases into the attention layers of the pretrained T2M model.
Previous works have explored similar directions using ControlNet~\cite{zhang2023adding} (\eg, SMooDi~\cite{zhong2024smoodi}) and LoRA~\cite{hu2022lora} (\eg, LoRA-MDM~\cite{sawdayee2025dance}), but these approaches either duplicate the model backbone or couple style with textual tokens.
In contrast, our SAM provides explicit, example-based style conditioning with minimal computational overhead and full compatibility with pretrained T2M architectures.

\paragraph*{Style Adaptation Module (SAM).}
\label{par:sam}
Instead of duplicating the entire T2M network as in ControlNet-style approaches, our Style Adaptation Module (SAM) introduces two lightweight learnable matrices, $W_s^{K}$ and $W_s^{V}$ (shown in red blocks in Figure~\ref{fig:controlnet}), which project the style embedding $s$ into key and value biases $\Delta K_s=W_s^{K}s$ and $\Delta V_s=W_s^{V}s$. The attention operation is then modified as:
\begin{equation}
\mathbf{A} = \mathrm{softmax}\left(\frac{Q (K + W_s^{K}s)^{\top}}{\sqrt{d}}\right) (V + W_s^{V}s).
\end{equation}
\rev{The style embedding is temporally resolved, so the biases are formed per token: token $t$ receives
$\Delta K_{s,t}=W_s^{K}s_t$ and $\Delta V_{s,t}=W_s^{V}s_t$, with $s_t$ the style code in force at that
frame, broadcast over the attention heads and the batch but not over time. The key bias would cancel in
the softmax only if $\Delta K_{s,t}$ were constant across an entire attention window, since $q\cdot\Delta
K_s$ would then be a per-query constant; under sliding-window attention with a style that varies along
the sequence this does not occur, and the bias reshapes the attention weights. Section~\ref{sec:ablation} ablates the two biases separately.}
This simple yet effective module conditions the pretrained T2M model on example-based style embeddings with minimal computational overhead. \rev{With a $384$-dimensional style code and a backbone of width $512$ and $7$ layers, SAM amounts to $6.44$M trainable parameters, $31\%$ of the deployed model and the only part of it that is trained; the count is derived in Appendix~\ref{sup:params}.} As demonstrated in our experiments (Section~\ref{sec:exp}), SAM achieves strong performance across diverse stylization and control tasks.

\begin{figure}[t]
  \centering
  \begin{minipage}[t]{0.53\linewidth}
    \centering
    \includegraphics[width=\linewidth]{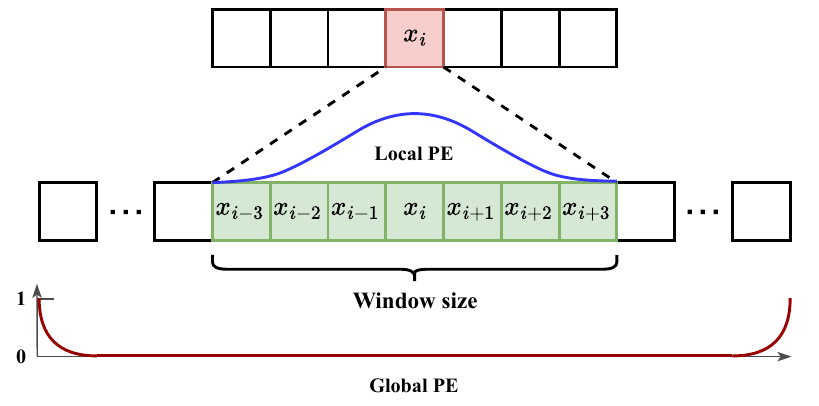}
    \caption{Illustration of positional encoding schemes. \textbf{Top:} Relative position encoding (RPE), where each frame attends to its neighbors within a local window. \textbf{Bottom:} Global position encoding (GPE), a single-channel signal marking sequence boundaries with smooth transitions between $1$s and $0$s.}
    \label{fig:relative_pe}
  \end{minipage}
  \hfill
  \begin{minipage}[t]{0.44\linewidth}
    \centering
    \includegraphics[width=\linewidth]{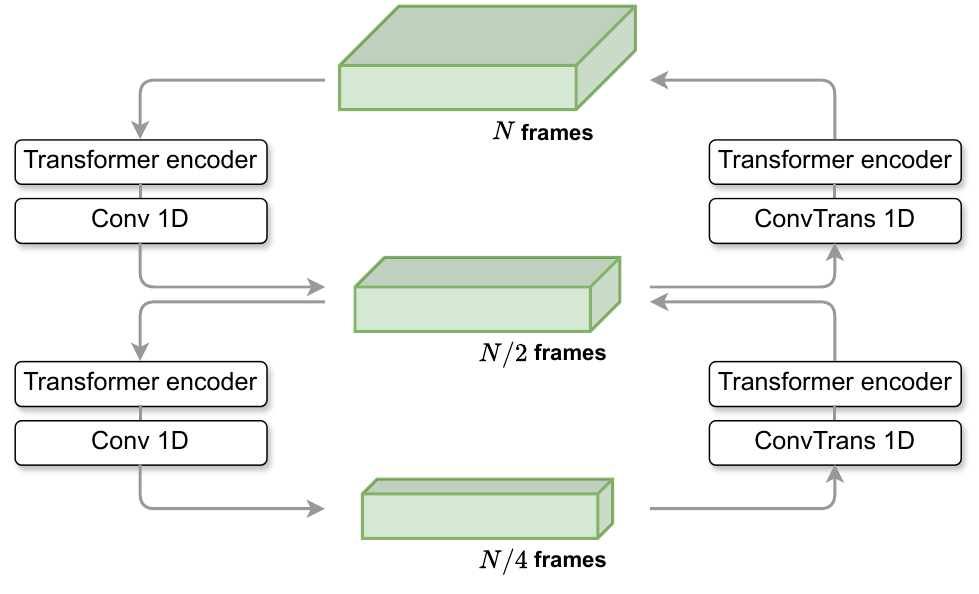}
    \caption{Architecture of the motion autoencoder. \textbf{Left:} The encoder $E_m$ applies two Transformer blocks and 1D convolutions, halving the frame count at each stage. \textbf{Right:} The decoder $D_m$ uses two 1D transposed convolutions followed by a Transformer block, doubling the frame count per stage.}
    \label{fig:motion_encoder}
  \end{minipage}
\end{figure}

\subsection{Pretraining text-to-motion model}

Motion latent diffusion models have been shown to outperform models that operate directly on raw motion data in works such as MLD~\cite{chen2023executing}, MoMask~\cite{guo2024momask}, and MotionLCM~\cite{motionlcm}.
However, collapsing motion features from $N$ frames into a single latent vector results in the loss of fine-grained temporal control.
To \rev{retain a temporally resolved representation}, we employ a temporal-preserving motion VAE, consisting of an encoder $E_m$ with two Transformer encoder blocks and two temporal downsampling operations, and a decoder $D_m$ with two Transformer blocks and two temporal upsampling operations.
An overview of the architecture is shown in Figure~\ref{fig:motion_encoder}.
\rev{Each downsampling operation halves the frame count, so the latent sequence is four times shorter
than the motion: one latent token per four frames, or $0.13$\,s at $30$\,fps. This is the granularity
at which content and style can be scheduled, and it is what we mean by \emph{temporally resolved}
control in the remainder of the paper. It is far finer than the clip- or sequence-level conditioning of
prior work, but it is not literally per frame, and we avoid the term \emph{frame-level} for it.}
Our text-to-motion diffusion model operates in this temporally preserved latent space, allowing the diffusion process to be conditioned on rich temporal representations for more coherent generation.

\paragraph*{Temporal attention.}

A limitation of many prior works is their inability to adjust content or style during generation. In contrast, our model enables \rev{temporally resolved} control, allowing different text prompts or style examples to be applied across time intervals. We employ sliding-window attention with relative position encoding (RPE)~\cite{shaw-etal-2018-self, wennberg-henter-2021-case, huang_music_transformer, su2024roformer} to achieve fine-grained temporal control. To ensure global semantic coherence (cf. FlowMDM~\cite{barquero2024seamless}), we further introduce a global position encoding (GPE), represented as a simple numeric signal indicating the start and end of the animation. Since the style encoder and motion encoder do not require global context, we apply RPE only to these modules, while using the combined RPE and GPE scheme in the T2M network.

As illustrated in Figure~\ref{fig:relative_pe}, Relative position encoding (RPE) restricts attention to a local temporal window $W$ of size $2K{+}1$.
Formally, for a sequence of frame embeddings $\mathbf{x}_1, \ldots, \mathbf{x}_N$, the attention operation is defined as:
\begin{align}
    &\mathbf{q}_i = \mathbf{W}^q \mathbf{x}_i,\;
    \mathbf{k}^{(i)}_j = \mathbf{W}^k [\,\mathbf{x}_j \Vert \mathbf{p}_{j-i}\,], \\
    &\mathbf{v}^{(i)}_j = \mathbf{W}^v [\,\mathbf{x}_j \Vert \mathbf{p}_{j-i}\,], \\
    &\mathbf{y}_i = \sum_{j = i-K}^{i+K}
        \mathrm{softmax}\!\left(
        \frac{\mathbf{q}_i^\top \mathbf{k}^{(i)}_j}{\sqrt{d}}
        \right)
        \mathbf{v}^{(i)}_j,
\end{align}
where $[\cdot \Vert \cdot]$ denotes vector concatenation, and $\mathbf{W}^q \!\in\! \mathbb{R}^{d\times d}$,
$\mathbf{W}^k, \mathbf{W}^v \!\in\! \mathbb{R}^{d\times (d{+}p)}$ are the learnable projection matrices.
The relative position vector $\mathbf{p}_{j-i}$ encodes the temporal offset between frames $i$ and $j$, allowing the model to reason about motion continuity and local transitions independent of absolute frame indices.
We adapted the \textit{shifting bits} techniques proposed in Huang \etal~\cite{huang_music_transformer} for efficient and scalable implementation.
The GPE signal is $1$ at sequence boundaries and transitions smoothly to $0$ mid-sequence; it is projected and combined with text and style embeddings for conditioning. Together, RPE and GPE support variable-length training, longer generations at inference, and distinct text and style conditions over time.

\subsection{Pretraining style encoder}\label{pretraining_se}

\begin{figure}[t]
  \centering
   \includegraphics[width=0.7\linewidth]{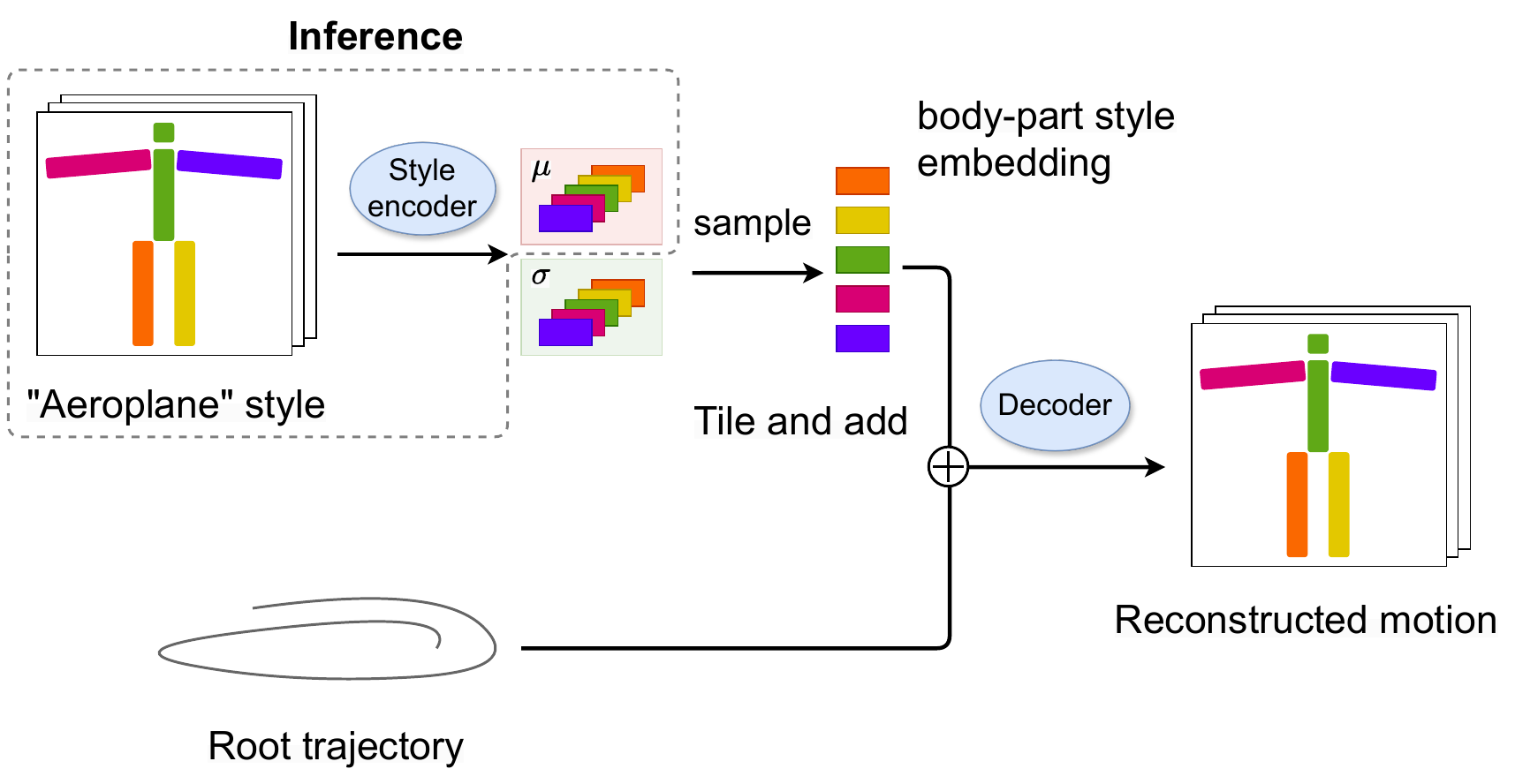}
\caption{Overview of style encoder pretraining. The encoder processes five body parts (legs, spine, and left/right arms) to produce per-part style embeddings. During training, style embeddings are sampled from the latent distribution and concatenated with the root trajectory for motion reconstruction through a decoder. At inference, only the encoder is used to extract the mean latent style code.}
   \label{fig:style_encoder}
\end{figure}

We train the style encoder on the 100STYLE dataset~\cite{mason2022local}, which contains 100 motion styles and 8 gait types. Our goal is to learn a latent style representation without relying on explicit style labels.
Inspired by ZeroEGGS~\cite{ghorbani2022zeroeggs}, we treat style as the motion information not explained by the root trajectory: the encoder learns to capture it by reconstructing the full motion from the root trajectory and a latent style code, effectively disentangling trajectory from style (Figure~\ref{fig:style_encoder}).
\rev{This proxy holds because all pretraining clips share the same underlying action: 100STYLE is locomotion performed in different manners, and we exclude the entries whose ``style'' is itself an action with its own limb articulation, such as punching or kicking, keeping 63 non-action styles (the classification of styles is given in Table~\ref{tab:classification_styles}). Any articulation left unexplained by the root trajectory is therefore stylistic rather than semantic, and at generation time the division is maintained architecturally: the action enters through the frozen backbone's text-conditioned queries, while style enters only as key and value biases (Section~\ref{par:sam}).}

\paragraph*{Architecture.}
We implement the encoder as a variational autoencoder (VAE) that produces the mean and variance of a Gaussian style distribution.
To capture both global and local stylistic features, we divide the skeleton into five body parts (legs, spine, and left/right arms) and compute a separate style embedding (of dimension 64) per part; together with a root component, this yields a $6\times64=384$-dimensional style code.
Prior work~\cite{Jang_2022, park2021diverse, kim2024most} shows that exploiting skeletal structure improves style transfer and motion blending; however, joint-level graph approaches~\cite{Jang_2022} are computationally expensive. We follow the lighter body-part modulator of Kim \etal~\cite{kim2024most}, achieving comparable skeletal awareness at lower cost.
The encoder comprises four transformer layers with temporal attention, operating independently on each body part.
\rev{Interactions between parts are masked according to a learned part-dependence graph, so each embedding stays tied to its own region; the graph and the skeleton division are given in Appendix~\ref{sup:bodypart}. A variant trained without the per-part division reaches comparable quantitative scores, but it separates styles less cleanly in the latent space: under t-SNE over 20 styles, the per-part code forms markedly tighter clusters with wider margins between them (Appendix~\ref{sup:tsne}). Because this code is what SAM injects at every layer, the tighter encoding leaves less ambiguity between styles at injection time, which is what the visibly higher style fidelity in the supplementary video reflects; the variant without the division also cannot express the per-part compositions of Section~\ref{sec:bodypart_mixing}.}

\paragraph*{Training.}
To encourage robustness to partial observations, we adopt the masking strategy of~\cite{Jang_2022}: body parts are randomly noised with probability $0.3$, and the model reconstructs the full motion from the remaining parts and the root trajectory.
Training minimizes a reconstruction loss on the output motion together with a KL divergence loss on the latent distribution, preventing posterior collapse and promoting a well-structured embedding space.
At inference, only the encoder is used; the mean of the posterior is taken as the deterministic style code, discarding the variance.

\subsection{Further implementation details}
\paragraph*{Classifier-free guidance.}
Classifier-free guidance (CFG)~\cite{ho2022classifier} allows a diffusion model to trade off between sample quality and diversity. In our case, we use CFG to enhance flexibility by balancing content and style during motion generation. During training, we randomly mask out text and style embeddings with a probability of $10\%$ each.
At sampling time, we follow a similar strategy to SMooDi~\cite{zhong2024smoodi}, using separate CFG weights to control the strength of content and style:
\begin{equation}
\begin{aligned}
G_{\theta}(\bm{z}_t, t, \bm{c}, \bm{s})
&= G_{\theta}(\bm{z}_t, t, \emptyset, \emptyset) \\
&+ \underbrace{w_c \left( G_{\theta}(\bm{z}_t, t, \bm{c}, \emptyset) - G_{\theta}(\bm{z}_t, t, \emptyset, \emptyset) \right)}_{\text{Classifier-free Content Guidance}} \\
&\quad + \underbrace{w_s \left( G_{\theta}(\bm{z}_t, t, \emptyset, \bm{s}) - G_{\theta}(\bm{z}_t, t, \emptyset, \emptyset) \right)}_{\text{Classifier-free Style Guidance}},
\end{aligned}\label{eq:cfg}
\end{equation}
where $G_{\theta}$ is the denoising network. It takes the noisy motion latent $\bm{z}_t$, the diffusion step $t$, the text embedding $\bm{c}$ and the style embedding $\bm{s}$, and predicts the clean latent; $w_c$ and $w_s$ control the relative strength of content and style guidance.

\paragraph*{Losses.}
During the style finetuning stage, we employ a motion reconstruction loss and a style preservation loss to optimize the diffusion model. Although cycle consistency loss~\cite{choi2020stargan, lee2018diverse, CycleGAN2017} was used in SMooDi~\cite{zhong2024smoodi}, it did not yield improvements in our experiments and is therefore omitted.
Let $\mathbf{m}$ and $\mathbf{x}$ denote the content and the style motion, and let $\bm{z} = E_m(\mathbf{m})$ and $\bm{z}' = E_m(\mathbf{x})$ be their VAE latents, with $\bm{z}_t$ and $\bm{z}'_t$ the corresponding latents at diffusion step $t$.
We write $\bm{c}_\mathbf{m}$ and $\bm{c}_\mathbf{x}$ for the text embeddings associated with $\mathbf{m}$ and $\mathbf{x}$, respectively.
The style encoder $SE$ operates on motion rather than on latents: it maps a motion to a Gaussian over style codes, so the style embedding of $\mathbf{x}$ is $\bm{s}_\mathbf{x} \sim \mathcal{N}(\bm{\mu}_\mathbf{x}, \bm{\sigma}_\mathbf{x})$ with $(\bm{\mu}_\mathbf{x}, \bm{\sigma}_\mathbf{x}) = SE(\mathbf{x})$, and likewise $\bm{s}_\mathbf{m}$ for $\mathbf{m}$.
The reconstruction loss encourages accurate reconstruction of both content and style motions when each serves as both content and style input:
\begin{equation}\label{eq:loss}
    \begin{aligned}
    \mathcal{L}_{recon} =  &\mathbb{E}_{\mathbf{m}, t} \left[ \Vert G_{\theta}(\bm{z}_{t}, t, \bm{c}_\mathbf{m}, \bm{s}_\mathbf{m} ) - \bm{z} \Vert^2\right] \\
    & + \mathbb{E}_{\mathbf{x}, t} \left[ \Vert G_{\theta}(\bm{z}'_{t}, t, \bm{c}_\mathbf{x}, \bm{s}_\mathbf{x}) - \bm{z}' \Vert^2\right].
    \end{aligned}
\end{equation}

To align the generated motion’s style with that of the reference motion $\mathbf{x}$, we apply an $\ell_2$-based style preservation loss:
\begin{equation}
\label{eq:style_loss}
    \mathcal{L}_{sty} =
    \Vert \bm{s}_{\Tilde{\mathbf{m}}} - \bm{s}_\mathbf{x} \Vert,
\end{equation}
where $\Tilde{\mathbf{m}} = D_m\big(G_{\theta}(\bm{z}_t, t, \bm{c}_\mathbf{m}, \bm{s}_\mathbf{x})\big)$ is the motion decoded from the generated latent, and $\bm{s}_{\Tilde{\mathbf{m}}} \sim \mathcal{N}(\bm{\mu}_{\Tilde{\mathbf{m}}}, \bm{\sigma}_{\Tilde{\mathbf{m}}})$ with $(\bm{\mu}_{\Tilde{\mathbf{m}}}, \bm{\sigma}_{\Tilde{\mathbf{m}}}) = SE(\Tilde{\mathbf{m}})$ is its style embedding.

The final objective is:
\begin{equation}
    \mathcal{L} = \mathcal{L}_{recon} + \lambda_{sty}\ \mathcal{L}_{sty},
\end{equation}
with $\lambda_{sty}=0.01$ in our experiments.

\section{Experiments}
\label{sec:exp}

We evaluate our model on two tasks: \((1)\) stylized text-to-motion generation with a single style for long-sequence generation (Subsection~\ref{subsec:long_stylized_t2m}), and \((2)\) stylized text-to-motion generation with multiple time-varying styles (Subsection~\ref{subsec:time-varying_style}).
In the following sections, we describe the implementation details, experimental setup, evaluation metrics, and both quantitative and qualitative results.

\paragraph*{Datasets.}
We use an internal high-quality motion capture (MoCap) dataset comprising 3,089 motion clips, primarily locomotion and various daily or sports activities, totaling approximately 6.4 hours of data (see Appendix~\ref{sup:dataset} for clip counts, durations, annotation counts and example activity categories).
In addition, we employ the 100STYLE dataset~\cite{mason2022local}, which contains about 20 hours of locomotion data across 100 distinct motion styles.
Both datasets were carefully annotated with natural language descriptions covering fine-grained motion details and high-level semantics.
For pretraining the text-to-motion (T2M) model, we combine the internal dataset and 100STYLE, using an 8:2 split for training and validation. All motion sequences are downsampled to 30~FPS. We also augmented the dataset with mirroring.
To pretrain the style encoder, we use 63 non-action styles from 100STYLE, and during training, animation sequences are randomly clipped to lengths between 30 and 200 frames to increase temporal diversity. The chosen styles for training, together with the 6 out-of-distribution styles held out for evaluation, are listed in Appendix~\ref{sup:dataset}.

\paragraph*{Motion representation.} \label{motion_rep}
Many recent works adopt the motion feature representation introduced in HumanML3D~\cite{guo2020action2motion, guo2024generative}, which encodes each frame as a 263-vector consisting of root linear and angular velocities, joint rotations, joint positions, joint velocities, and foot-contact labels.
However, the joint rotation representation in HumanML3D is incompatible with common 3D modeling software, making post-processing and visualization cumbersome.
To address this limitation, we adapt our motion features from ZeroEGGS~\cite{ghorbani2022zeroeggs}.
Each frame $i$ is represented as $\mathbf{x}_i = [\dot{\mathbf{r}}_a, \dot{\mathbf{r}}_p, \mathbf{h}_p, \mathbf{j}_r, \mathbf{f}_c]$,
where $\dot{\mathbf{r}}_a\in\mathbb{R}^1$ and $\dot{\mathbf{r}}_p\in\mathbb{R}^2$ denote the character root angular velocity around the vertical axis and root linear velocity relative to the facing direction, respectively;
$\mathbf{h}_p\in\mathbb{R}^3$ is the hip translation local to the root;
$\mathbf{j}_r\in\mathbb{R}^{6J}$ represents joint rotations in the 2-column format~\cite{10.1145/3197517.3201366, DBLP:journals/corr/abs-1812-07035};
and $\mathbf{f}_c\in\mathbb{R}^2$ encodes foot-contact labels.
We set $J=25$ to denote the number of skeletal joints, with the facing direction defined as the forward axis of the pelvis.
A motion sequence with $N$ frames is thus represented as $[\mathbf{x}_0, \mathbf{x}_1, \ldots, \mathbf{x}_N]$.
This representation is directly convertible to BVH files, facilitating seamless integration with 3D modeling and animation software.

\begin{table}[t]
\centering
\caption{Comparison of text–style motion frameworks.
($\checkmark$: supported, $\times$: not supported.)}
\label{tab:baselines}
\begin{tabular*}{\linewidth}{@{\extracolsep{\fill}}lcccc@{}}
\toprule
\textbf{Model} & \textbf{Style Injection} & \textbf{Label-free} & \textbf{Temporal Control} & \textbf{Long Sequence} \\
\midrule
T2M+MP & AdaIN & \checkmark & \checkmark & $\times$ \\
SMooDi & ControlNet & $\times$ & $\times$ & $\times$ \\
LoRA-MDM & Token concatenation & $\times$ & $\times$ & $\times$ \\
Ours & Attention bias & \checkmark & \checkmark & \checkmark \\
\bottomrule
\end{tabular*}
\end{table}

\paragraph*{Baselines.}

We compare our method against three baselines: SMooDi~\cite{zhong2024smoodi}, our T2M model combined with MotionPuzzle~\cite{Jang_2022} (T2M+MP), and LoRA-MDM~\cite{sawdayee2025dance}.
All models are trained under identical settings whenever applicable to ensure fair comparison.
For stylized text-to-motion generation, SMooDi, LoRA-MDM, and our model directly synthesize motions conditioned on both a text prompt and a style example.
In contrast, MotionPuzzle performs style transfer from a reference clip to a given motion; therefore, we pair it with our T2M model—first generating motion from text using T2M, followed by a style-transfer stage from the style example to the T2M output motion. Table~\ref{tab:baselines} summarizes the baselines in terms of their style injection strategy, reliance on style labels, and their ability to handle long sequences or time-varying style inputs.

\paragraph*{Evaluation metrics.}
We evaluate all models on three aspects: content preservation, style reflection, and motion quality.
For content preservation, we adopt standard T2M metrics~\cite{Guo_2022_CVPR, guo2020action2motion}, including \textit{R-Precision}, \textit{Fréchet Inception Distance (FID)}, and \textit{Multimodal Distance (MM Dist)}.
Our T2M evaluator is trained on both content and style datasets using the same architecture as~\cite{Guo_2022_CVPR}.
We further compute the \textit{CLIP Score}~\cite{hessel2021clipscore} to measure text–motion alignment, using a CLIP-based evaluator trained following~\cite{meng2024rethinking}.
For style reflection, we report the \textit{Style Reflection Accuracy (SRA)}~\cite{Jang_2022}, measured by a style classifier trained on the 100STYLE dataset.
All models are trained on 63 styles and evaluated on 69 styles, including 6 out-of-distribution styles, with non-overlapping evaluation examples. We report the top-3 \textit{SRA}.
\rev{
However, \textit{SRA} has two limitations. It is defined only over the styles the classifier knows, so fidelity
to an arbitrary reference cannot be quantified and references outside 100STYLE are shown qualitatively
instead; and it can be raised by copying poses from the reference rather than by transferring style,
which is visible as T2M+MP's competitive \textit{SRA} alongside the weakest content scores of any
baseline. We know of no metric that avoids either, so we read \textit{SRA} together with the content
metrics and the segment-level measurements of Section~\ref{subsec:time-varying_style} rather than on
its own. Note also that our out-of-distribution styles are unseen by the \emph{generator} but not by
the \emph{classifier}, which covers all of 100STYLE by construction; out-of-distribution \textit{SRA}
may therefore equal or exceed the in-domain value, and the meaningful comparison there is between
methods rather than between the two columns.
}
For motion quality evaluation, we measure the foot-skating ratio, which is defined as the proportion of frames where either foot slides while in contact with the ground.

\begin{table}[p]
\centering
\caption{Comparison of stylized text-to-motion generation against baselines T2M+MotionPuzzle (MP), SMooDi, and LoRA-MDM, evaluated on motion sequences ranging from 30 to 400 frames and style examples sampled from 69 styles of the 100STYLE dataset outside the training set. The top-3 \textit{R-Precision} and \textit{SRA} scores are reported. $\uparrow$ indicates higher is better, $\downarrow$ indicates lower is better. We use \textbf{bold} for the best score and \underline{underscore} for the second best.}
\label{tab:metric}
\small
\begin{tabular*}{\linewidth}{@{\extracolsep{\fill}}lcccccccc@{}}
\toprule
\multirow{2}{*}{\textbf{Models}}
& \multirow{2}{*}{\textbf{MM Dist$\downarrow$}}
& \multirow{2}{*}{\textbf{R-precision$\uparrow$}}
& \multirow{2}{*}{\textbf{FID$\downarrow$}}
& \multirow{2}{*}{\textbf{CLIP score$\uparrow$}}
& \multicolumn{3}{c}{\textbf{SRA$\uparrow$}}
& \multirow{2}{*}{\textbf{Foot Skating Ratio$\downarrow$}} \\
\cmidrule(lr){6-8}
&  &  &  &
& \textbf{Top-1} & \textbf{Top-2} & \textbf{Top-3}
&  \\
\midrule
 Real   & 1.1120  & 0.9613 & 0.4157 & 0.7654 & 0.9988 & 0.9988  & 0.9988 & 0.0001 \\
 \midrule
 T2M+MP  & 5.1688 &  0.2319 & 9.6582  & 0.2452 & \textbf{0.9056} & \textbf{0.9494} & \textbf{0.9650}  & 0.8435 \\
 SMooDi  & 3.9109  & 0.4681 & \underline{2.9284}  & 0.4197 & 0.6619 & 0.6981 & 0.7125  & 0.1164 \\
 LoRA-MDM & \textbf{1.7546} &  \textbf{0.8756} & \textbf{0.6377} & \textbf{0.6954} & 0.0156 & 0.0325 & 0.0475 & \textbf{0.0797}\\
 \midrule
 FlexMoGen (Ours)     & \underline{3.5289}  & \underline{0.5756} & 3.0782  & \underline{0.5387} & \underline{0.6863} & \underline{0.7969} & \underline{0.8356} &  \underline{0.0971} \\
\bottomrule
\end{tabular*}
\end{table}
\begin{figure}[p]
    \centering
    \begin{subfigure}[b]{0.19\textwidth}
        \includegraphics[width=0.88\textwidth]{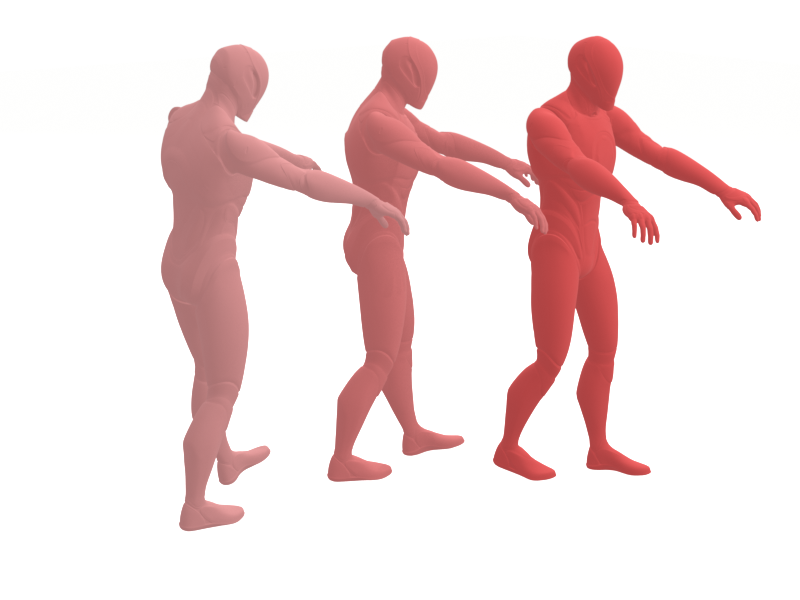}
        \caption{\textit{Zombie} style}
    \end{subfigure}
    \hfill
    \begin{subfigure}[b]{0.19\textwidth}
        \includegraphics[width=0.88\textwidth]{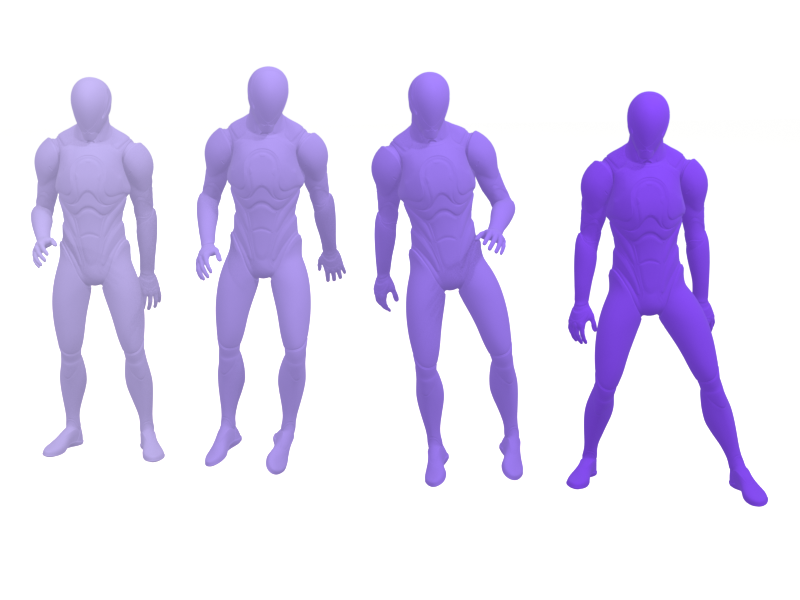}
        \caption{LoRA-MDM}
    \end{subfigure}
    \hfill
    \begin{subfigure}[b]{0.19\textwidth}
        \includegraphics[width=0.88\textwidth]{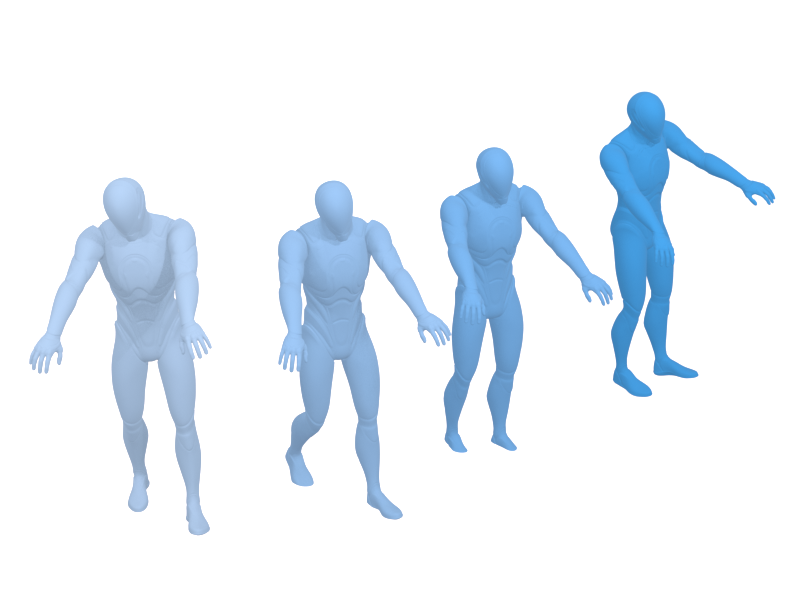}
        \caption{T2M$+$MotionPuzzle}
    \end{subfigure}
    \hfill
    \begin{subfigure}[b]{0.19\textwidth}
        \includegraphics[width=0.88\textwidth]{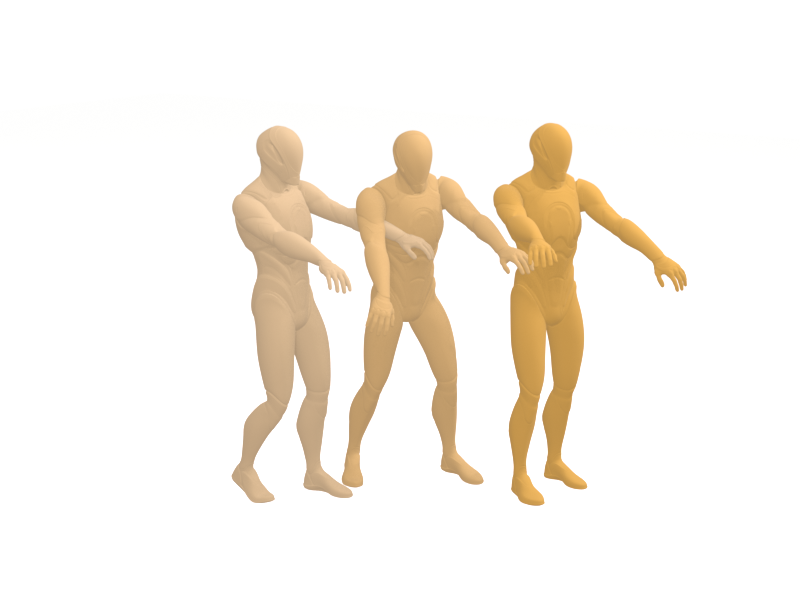}
        \caption{SMooDi}
    \end{subfigure}
    \hfill
    \begin{subfigure}[b]{0.19\textwidth}
        \includegraphics[width=0.88\textwidth]{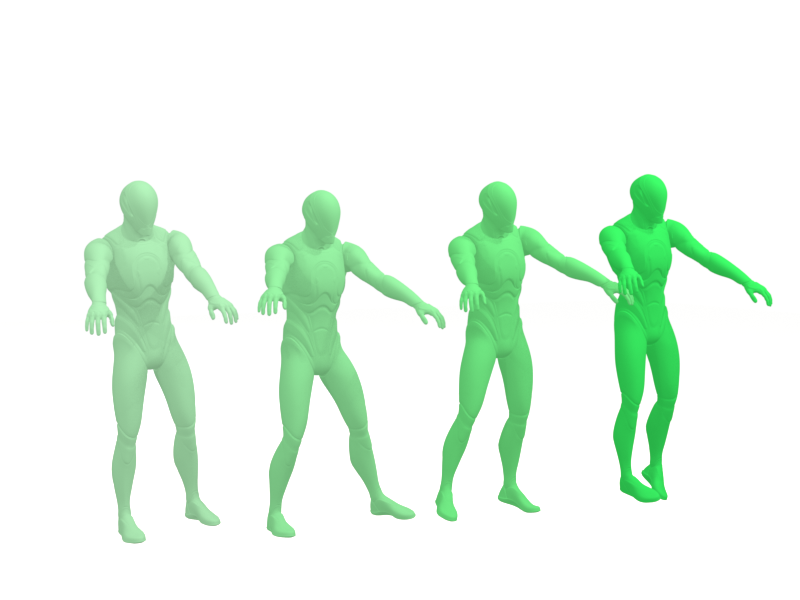}
        \caption{Ours}
    \end{subfigure}
    \caption{Comparison of different models on the text prompt ``Side hop to left'' with the \textit{Zombie} style (arms raised forward with a stiff body). Only our model produces a motion where the character hops to the left in a straight line with arms raised in front. The color transition from light to dark indicates earlier to later frames in the motion sequence.}
    \label{fig:sidehoptoleft_zombie}
\end{figure}

\begin{table}[p]
\centering
\caption{Comparison on time-varying style input among T2M+MotionPuzzle (MP) and SMooDi, evaluated on motion sequences ranging from 30 to 200 frames, with distinct style examples from the 100STYLE dataset outside the training set. The top-3 \textit{R-Precision} and \textit{SRA} scores are reported. We use \textbf{bold} for the best score.}
\label{tab:varying_style_metric}
\small
\begin{tabular*}{\linewidth}{@{\extracolsep{\fill}}lcccccccc@{}}
\toprule
\multirow{2}{*}{\textbf{Models}}
& \multirow{2}{*}{\textbf{MM Dist$\downarrow$}}
& \multirow{2}{*}{\textbf{R-precision$\uparrow$}}
& \multirow{2}{*}{\textbf{FID$\downarrow$}}
& \multirow{2}{*}{\textbf{CLIP score$\uparrow$}}
& \multicolumn{3}{c}{\textbf{SRA$\uparrow$}}
& \multirow{2}{*}{\textbf{Foot Skating Ratio$\downarrow$}} \\
\cmidrule(lr){6-8}
&  &  &  &
& \textbf{Top-1} & \textbf{Top-2} & \textbf{Top-3}
&  \\
\midrule
Real   & 1.2323  & 0.9519 & 0.3589 & 0.7432 & 0.9988 & 0.9997 & 1.0000 &  0.0000\\
\midrule
 T2M+MP  & 4.9206 & 0.2375 & 10.1706   & 0.2554 & 0.4262 & 0.6081 & 0.6791 & 0.8292\\
 SMooDi  & 4.2410 & 0.4100 & \textbf{3.0211}   & 0.3234 & 0.2272 & 0.2863 & 0.3228 & 0.0940 \\
 \midrule
 FlexMoGen (Ours)     & \textbf{3.3034} & \textbf{0.6337} & 3.4981 & \textbf{0.5612} & \textbf{0.5328} & \textbf{0.6637} & \textbf{0.7272} & \textbf{0.0818} \\
\bottomrule
\end{tabular*}
\end{table}

\begin{figure}[p]
    \centering
    \begin{subfigure}[b]{0.24\textwidth}
        \includegraphics[width=0.88\textwidth]{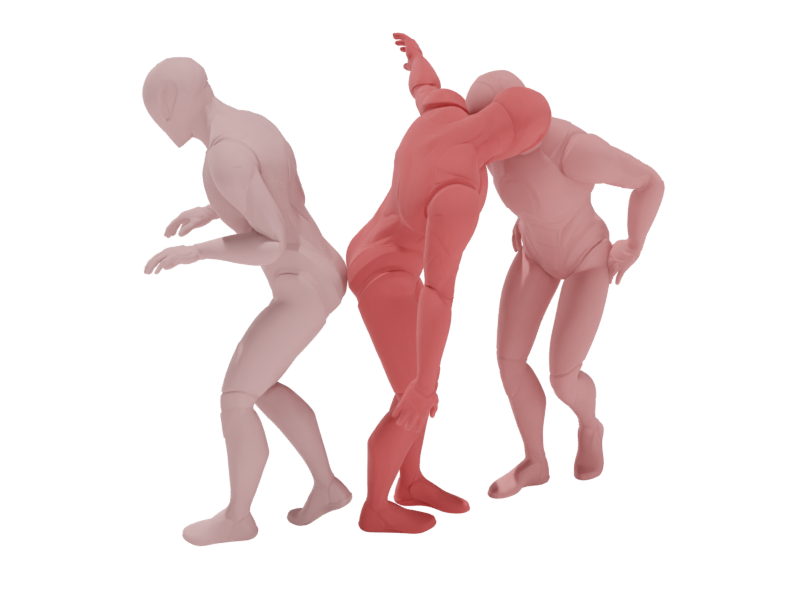}
        \caption{Transitions from \textit{Dinosaur} to \textit{Chicken} to \textit{Aeroplane} styles}
    \end{subfigure}
    \hfill
    \begin{subfigure}[b]{0.24\textwidth}
        \includegraphics[width=0.88\textwidth]{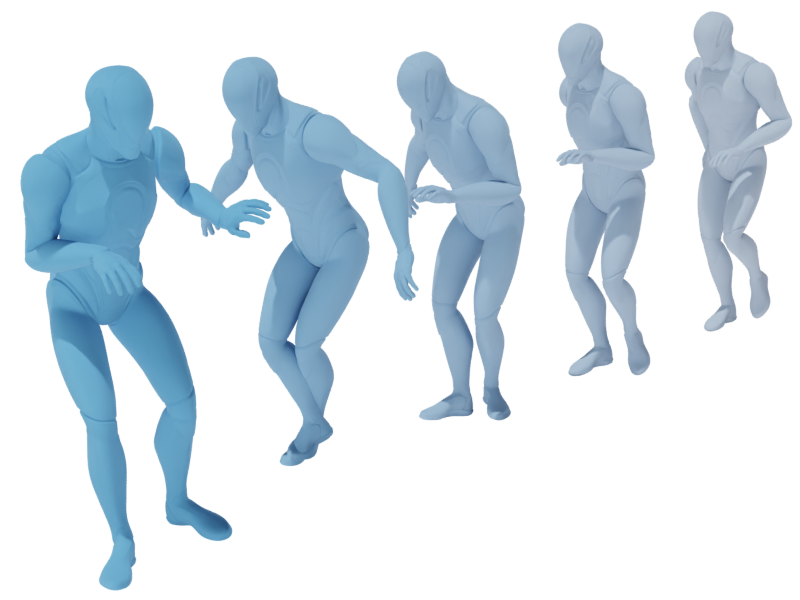}
        \caption{T2M+MotionPuzzle}\label{fig:multistyle_mp}
    \end{subfigure}
    \hfill
    \begin{subfigure}[b]{0.24\textwidth}
        \includegraphics[width=0.88\textwidth]{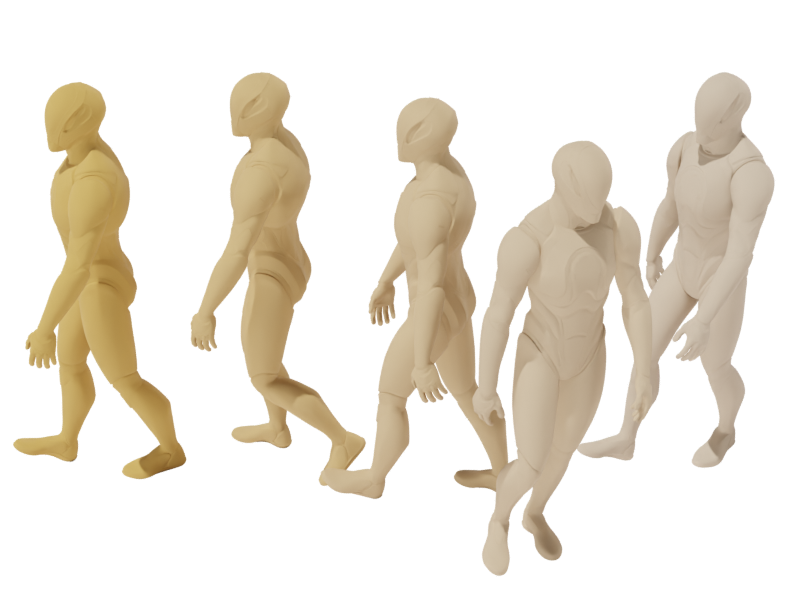}
    \caption{SMooDi}\label{fig:multistyle_smoodi}
    \end{subfigure}
    \hfill
    \begin{subfigure}[b]{0.24\textwidth}
        \includegraphics[width=0.88\textwidth]{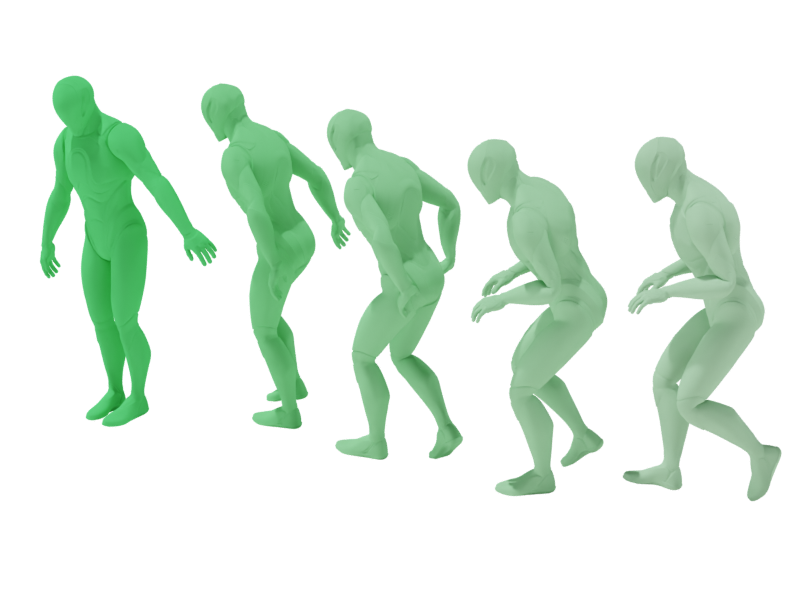}
        \caption{Ours}\label{fig:multistyle_ours}
    \end{subfigure}

    \caption{Comparison of different models on the text prompt ``Walking forward then stopping and turning left'', with styles transitioning from \textit{Dinosaur} to \textit{Chicken} to \textit{Aeroplane}. Only our model accurately captures the intended style transitions while performing all actions: \textit{walking forward}, \textit{stopping, and turning left}.}
    \label{fig:multistyle}
\end{figure}

\subsection{Long sequence generation}\label{subsec:long_stylized_t2m}

Although all models are trained on motion sequences of 30–200 frames, we evaluate them on sequences up to 400 frames to assess their generalization to longer motions.
All experiments use DDIM~\cite{song2020denoising} accelerated sampling with 100 diffusion steps. For our model, the classifier-free guidance weights $w_c$ and $w_s$ (Equation~\ref{eq:cfg}) are set to $2.0$ and $1.0$, respectively. Results are reported in Table~\ref{tab:metric}.

Generating long sequences poses significant challenges—most baseline models fail to maintain plausible poses beyond 200 frames. SMooDi and LoRA-MDM, which rely solely on global position encoding, tend to lose temporal coherence and produce drifting poses.
MotionPuzzle constrains the output to follow the trajectory of the content motion, but this often results in inconsistent limb movements and local artifacts.

In contrast, our model combines relative and global position encodings, enabling localized temporal reasoning while maintaining global semantic consistency.
\rev{As a result, it ranks second on each of the T2M content metrics---\textit{MM Dist.},
\textit{R-Precision} and \textit{CLIP Score}---and second in \textit{SRA}, and it is the only method
placed this highly on both at once.}
While SMooDi benefits from classifier-based guidance, its \textit{SRA} is not as high as our model’s.
Its slightly lower \textit{FID} can be attributed to both its classifier-based guidance and its use of a single global motion latent, which aligns more closely with the T2M evaluator’s embedding representation.
In contrast, our model’s temporally structured latent space captures richer \rev{temporally resolved} dynamics and offers finer flexibility, at the expense of a marginally higher \textit{FID}.
T2M+MP attains high \textit{SRA} by directly copying from the style motion but fails to align with the content semantics, while LoRA-MDM, originally designed for fine-tuning a small number of styles, loses its ability to reflect styles when trained on all 63 styles.

\rev{Stylized text-to-motion requires content preservation \emph{and} style reflection together, so we
do not read Table~\ref{tab:metric} as a per-column ranking: a model that leads every content metric
with a top-1 \textit{SRA} of $0.0156$, or leads \textit{SRA} with an \textit{R-Precision} of $0.2319$,
has not performed the task.}
Overall, our model achieves the best balance among content fidelity, stylistic accuracy, and motion quality across all methods.

Through the visualization in Figure~\ref{fig:sidehoptoleft_zombie}, we verify that only our model generates a motion that performs the desired action of \textit{side hop to the left} while faithfully reflecting the \textit{Zombie} style, characterized by arms raised in front.
LoRA-MDM performs the content motion but fails to preserve the style.
For T2M+MP, the character’s foot placement—one foot forward and the other backward—suggests a \textit{walking} motion rather than \textit{hopping}, indicating that the model largely copies poses from the style example (which performs walking), despite following a roughly correct trajectory.
SMooDi captures the \textit{Zombie} style but produces only a single step to the left before stopping. This is because their model is unpredictable when generating animations longer than training data.
Animated results for these examples are included in the supplementary video.

\subsection{Temporally-varying style generation} \label{subsec:time-varying_style}
We further evaluate our model’s ability to generate motions conditioned on temporally varying style examples. In this experiment, multiple style clips are provided as input, while the output motion length is fixed to 200 frames, which is consistent with the training setup.

For our model, to generate a stylized sequence of $N$ frames, we first extract a style embedding for each reference clip. Each embedding is repeated to match its assigned time segment, and all are concatenated to form a time-dependent sequence of style embeddings of length $N$.
In contrast, SMooDi~\cite{zhong2024smoodi} represents motion and style using single global vectors. To simulate multiple styles, we concatenate all style clips with inertialization~\cite{BolloInertialize} and compute a single style embedding.
Similarly, MotionPuzzle~\cite{Jang_2022} uses the concatenated style motion as a single style input motion.
LoRA-MDM~\cite{sawdayee2025dance}, which relies on text-based style control, cannot handle temporally varying style inputs and is therefore excluded from this comparison.

Results are shown in Table~\ref{tab:varying_style_metric}, with one example visualized in Figure~\ref{fig:multistyle}.
Our model achieves the best scores on \textit{Multimodal Distance}, \textit{R-Precision}, and \textit{CLIP Score}, and also attains the highest \textit{SRA}, thanks to its temporally resolved style embeddings.
\rev{\textit{FID} is the one exception, for the same reason as in Table~\ref{tab:metric}. Here, unlike
the single-style setting, the two requirements are not split between competing baselines: our model
leads on content and on style at the same time.}
As illustrated in Figure~\ref{fig:multistyle_ours}, our results exhibit accurate action transitions and correct temporal ordering of styles.

SMooDi, constrained by a single style vector, fails to represent multiple styles accurately, leading to degraded content and style consistency even when classifier-based style guidance is strengthened.
As shown in Figure~\ref{fig:multistyle_smoodi}, it performs the actions \textit{walk forward}, \textit{turn left}, and \textit{stop}, but in the wrong temporal order, and without reflecting any of the intended styles.
T2M+MP also misaligns styles with their intended segments: as shown in Figure~\ref{fig:multistyle_mp}, styles such as \textit{Dinosaur}, \textit{Chicken}, and \textit{Aeroplane} appear on random frames rather than in the correct sequence.
Animated results for all examples are provided in the supplementary video.

\subsection{Style generalization test}\label{sec:generalization}
Among the three baselines, T2M+MP tends to ignore the content motion, while LoRA-MDM fails to capture the target style. SMooDi is the strongest baseline, and we therefore focus on comparing its generalization ability with our model.
To evaluate generalization to unseen styles, we report Style Reflection
Accuracy (SRA) on the $63$ in-domain and the $6$ out-of-domain styles separately, on the same test set
as the two evaluations of Sections~\ref{subsec:long_stylized_t2m}
and~\ref{subsec:time-varying_style}, so that the two generation settings are compared under the same
style split; Table~\ref{tab:classification_styles} in Appendix~\ref{sup:dataset} lists which styles
fall on each side of it. Table~\ref{tab:style-comparison-long-general} covers long-sequence generation
and Table~\ref{tab:style-comparison-multistyle-general} time-varying style generation.

\begin{table}[ht]
\centering
\caption{Comparison on in-domain and out-of-domain styles on long sequences}
\begin{tabular*}{\linewidth}{@{\extracolsep{\fill}}l ccc ccc@{}}
\toprule

\multirow{2}{*}{\textbf{Models}}
& \multicolumn{3}{c}{\textbf{SRA (In-domain)$\uparrow$}}
& \multicolumn{3}{c}{\textbf{SRA (Out-of-domain)$\uparrow$}} \\

\cmidrule(lr){2-4} \cmidrule(lr){5-7}

& \textbf{Top-1} & \textbf{Top-2} & \textbf{Top-3}
& \textbf{Top-1} & \textbf{Top-2} & \textbf{Top-3} \\

\midrule

 SMooDi & 0.6721 & 0.7047 & 0.7164 & 0.5467 & 0.6213 & 0.6674 \\
FlexMoGen (Ours)   & \textbf{0.6882} & \textbf{0.7974} & \textbf{0.8350} & \textbf{0.6634} & \textbf{0.7903} & \textbf{0.8508} \\

\bottomrule
\end{tabular*}
\label{tab:style-comparison-long-general}
\end{table}

\begin{table}[ht]
\centering
\caption{Comparison on in-domain and out-of-domain styles on time varying styles}
\begin{tabular*}{\linewidth}{@{\extracolsep{\fill}}l ccc ccc@{}}
\toprule

\multirow{2}{*}{\textbf{Models}}
& \multicolumn{3}{c}{\textbf{SRA (In-domain)$\uparrow$}}
& \multicolumn{3}{c}{\textbf{SRA (Out-of-domain)$\uparrow$}} \\

\cmidrule(lr){2-4} \cmidrule(lr){5-7}

& \textbf{Top-1} & \textbf{Top-2} & \textbf{Top-3}
& \textbf{Top-1} & \textbf{Top-2} & \textbf{Top-3} \\

\midrule

 SMooDi & 0.2306 & 0.2883 & 0.3240 & 0.1895 & 0.2643 & 0.3104 \\
FlexMoGen (Ours)   & \textbf{0.5359} & \textbf{0.6661} & \textbf{0.7278} & \textbf{0.4979} & \textbf{0.6408} & \textbf{0.7264} \\

\bottomrule
\end{tabular*}
\label{tab:style-comparison-multistyle-general}
\end{table}

\rev{SMooDi shows a noticeable drop in style accuracy on out-of-domain styles in the long-sequence
setting, losing $0.1254$ top-1 \textit{SRA} between the two style sets ($0.6721$ against $0.5467$)
against $0.0248$ for our model ($0.6882$ against $0.6634$), which therefore maintains consistently high
\textit{SRA} on styles it was never trained on. In the time-varying setting the two methods lose the
same amount ($0.0411$ and $0.0410$), but from very different levels: our out-of-domain top-1 of
$0.4979$ is more than twice SMooDi's in-domain $0.2306$, and we lead at every rank on both style sets.
The comparison is between withheld and non-withheld 100STYLE styles throughout; references from outside
the dataset are a separate matter, treated in Appendix~\ref{sup:vis-ood}.}

\rev{
\paragraph*{Styles outside 100STYLE.}
The six out-of-domain styles above are withheld from training, but they are still 100STYLE clips and
so share its capture setup, its subject and its locomotion-centred vocabulary. Our largest failure
mode appears once that shared origin is removed: on references from an internal capture of largely
upper-body and prop-handling motions, transfer is markedly less reliable, and it fails unevenly rather
than uniformly, some references being reflected about as clearly as an in-domain style and others
barely at all. The failures are graceful, in that the content prompt continues to be followed and what
is lost is the stylistic character of the reference rather than the motion itself. We attribute this to
the style encoder's pretraining distribution, which consists of locomotion styles from a single capture
source (Section~\ref{pretraining_se}), and accordingly claim generalization to unseen styles only
within that distribution. The setting admits no quantitative score, since \textit{SRA} is defined only
over the labelled 100STYLE taxonomy; three references spanning the above range are therefore examined
qualitatively in Appendix~\ref{sup:vis-ood} and animated in the supplementary video.
}

\rev{
\subsection{Evaluations on temporal localization and long sequences}
\label{sec:more_eval}
Two further evaluations extend the time-varying setting of
Section~\ref{subsec:time-varying_style}, and both are reported in full in the appendices. The first asks whether each style stays inside the interval it was assigned to; the second
asks whether temporal control and long-sequence generation hold at the same time.

\paragraph*{Temporal localization (Appendix~\ref{sup:localization}).}
Sequence-level \textit{SRA} cannot tell whether a style is confined to its interval: a model that
applies the first reference to the entire sequence already collects half of it. We therefore measure
localization directly, on the $200$-frame protocol of
Section~\ref{subsec:time-varying_style}, with segment-wise \textit{SRA}, a
leakage measure between the two intervals, the delay and the failure rate of the requested switch, and
a boundary check on transition smoothness. FlexMoGen is the only method whose two intervals score
alike ($0.5380$ against $0.5352$, in-domain); both baselines lose $0.17$ to $0.34$ between the first
and the second interval, that is, they keep producing the first style after the switch was requested.
Their leakage sits at $\approx0.49$, exactly the value obtained when a single style covers both
intervals, against $0.1064$ for ours, and the switch is detectable in $90.6\%$ of our sequences and
lands within $1.5$ frames ($0.05$s) of where it was requested, against failure rates of $34.4\%$ and
$53.4\%$ and offsets of $6$ to $14$ frames for the baselines. The margins and the timing are unchanged
on the out-of-domain styles.

\paragraph*{Long sequences with time-varying style (Appendix~\ref{sup:long-multistyle}).}
Each evaluation so far varies one factor only: Table~\ref{tab:varying_style_metric} and the
localization evaluation above fix the output at $200$ frames, the maximum training length, while
Section~\ref{subsec:long_stylized_t2m} reaches $400$ frames but under a single style. We therefore
repeat that whole protocol, sequence-level and segment-level alike, at $200$, $400$ and
$800$ frames, so that length and time-varying style are demanded of the model together. Since the two
style intervals split the sequence, at $800$ frames each of them is itself twice the training maximum.
Localization is unaffected by the length: the two intervals still score alike, leakage and the
detection rate stay at their $200$-frame values, and \textit{FID} and foot skating do not grow. The
one cost is a mild loss of style fidelity at the longest setting.
}

\rev{\subsection{Body-part style mixing}\label{sec:bodypart_mixing}}
\rev{Because the style encoder embeds the five body parts independently
(Section~\ref{pretraining_se}), the style code is not a single monolithic vector but a concatenation
of per-part components. This makes available a capability we have not used so far: the components can
be taken from \emph{different} reference clips and assembled into one code before injection, so that a
single generated motion carries one style in the legs, another in the spine and another in the arms.
This does not require any changes to the model or retraining
--- only the assembly of the style code is required.}

\rev{We demonstrate this with two compositions. The first draws three references, one per part group:
\textit{Chicken} legs, \textit{Aeroplane} arms and a \textit{Zombie} spine. The second draws four,
assigning \emph{different} styles to the left and the right arm (\textit{RaisedLeftArm} and
\textit{ArmsBehindBack}) over a \textit{Zombie} spine and \textit{LegsApart} legs; the generated
motion is correspondingly asymmetric between the two arms, a combination that no single style
reference provides, and which also indicates that the part components act locally rather than being
blended into one global style. Both are shown in Appendix~\ref{sup:vis-bodypart} and animated in the
supplementary video. We report them qualitatively: \textit{SRA} scores a whole sequence against the
100STYLE label taxonomy, and a mixed-part composition has no ground-truth label in that taxonomy, so
the setting admits no meaningful quantitative score. We regard this as a downstream use of the style
representation rather than a contribution of the generation framework itself, but it is what most
directly motivates the per-body-part design of Section~\ref{pretraining_se}, as whole-body stylization
alone would not require it.}

\FloatBarrier
\rev{\subsection{Ablation study}\label{sec:ablation}}

To validate our design choices, we conduct ablation experiments on three key components: (1) the choice of positional encoding and network architecture, (2) pretraining a motion VAE, and (3) the style injection mechanism.
Results are reported for both single-style (Table~\ref{tab:ablation_single_style}) and time-varying style (Table~\ref{tab:ablation_time_varying_style}) settings.

\begin{table}[t]
\centering
\caption{Ablation study on major design components for single style input. The top-3 \textit{R-Precision} and \textit{SRA} scores are reported. \rev{The \textit{SASI} row replaces our SAM with the style injection module of Wu \etal, retrained on the same backbone and data; its style guidance is set so that its top-1 \textit{SRA} matches ours in this setting. The last two rows restrict our own SAM to a key bias or to a value bias alone.}}
\label{tab:ablation_single_style}
\small
\begin{tabular*}{\linewidth}{@{\extracolsep{\fill}}lcccccccc@{}}
\toprule
\multirow{2}{*}{\textbf{Models}}
& \multirow{2}{*}{\textbf{MM Dist$\downarrow$}}
& \multirow{2}{*}{\textbf{R-precision$\uparrow$}}
& \multirow{2}{*}{\textbf{FID$\downarrow$}}
& \multirow{2}{*}{\textbf{CLIP score$\uparrow$}}
& \multicolumn{3}{c}{\textbf{SRA$\uparrow$}}
& \multirow{2}{*}{\textbf{Foot Skating Ratio$\downarrow$}} \\
\cmidrule(lr){6-8}
&  &  &  &
& \textbf{Top-1} & \textbf{Top-2} & \textbf{Top-3}
&  \\
\midrule
Real   & 1.112  & 0.9613 & 0.4157 & 0.7654 & 0.9988 & 0.9998 & 0.9988 &  0.0001\\
 Ours    & 3.5289 & 0.5756 & 3.0782 & 0.5387 & 0.6863 & 0.7969 & 0.8356 & 0.0971 \\
 Sinusoidal PE & 3.6786	&0.5556	& 3.8996 & 0.5198	& 0.6669	&0.7856	& 0.8144	&0.0963\\
 UNet & 3.503 & 0.5631 & 1.6786 & 0.4711 & 0.2331 & 0.3006 & 0.3456 & 0.0314 \\
 No VAE & 3.6713	&	0.5525 & 4.3438&	0.5306&	0.6469&	0.7737&	0.815&	0.0815 \\
 ControlNet & 3.599 & 0.5813 & 3.5267 & 0.528 & 0.6613 & 0.8069 & 0.8531 & 0.1009 \\
 Cross Attn & 3.586 & 0.55 & 2.93 & 0.5001 & 0.6663 & 0.7744 & 0.8194 & 0.1096 \\
 \rev{SASI} & \rev{3.2151} & \rev{0.6150} & \rev{1.0491} & \rev{0.4871} & \rev{0.6719} & \rev{0.7312} & \rev{0.7662} & \rev{0.0576} \\
 \rev{Key bias only} & \rev{1.2242} & \rev{0.9637} & \rev{0.4605} & \rev{0.7527} & \rev{0.0156} & \rev{0.0331} & \rev{0.0587} & \rev{0.0716} \\
 \rev{Value bias only} & \rev{3.5631} & \rev{0.5750} & \rev{3.1893} & \rev{0.5357} & \rev{0.6937} & \rev{0.8063} & \rev{0.8388} & \rev{0.0966} \\
\bottomrule
\end{tabular*}
\end{table}

\begin{table}[t]
\centering
\caption{Ablation study on major design components for time-varying style input. The top-3 \textit{R-Precision} and \textit{SRA} scores are reported. \rev{The \textit{SASI} row uses the same replacement and the same guidance setting as in Table~\ref{tab:ablation_single_style}. The last two rows restrict our own SAM to a key bias or to a value bias alone.}}
\label{tab:ablation_time_varying_style}
\small
\begin{tabular*}{\linewidth}{@{\extracolsep{\fill}}lcccccccc@{}}
\toprule
\multirow{2}{*}{\textbf{Models}}
& \multirow{2}{*}{\textbf{MM Dist$\downarrow$}}
& \multirow{2}{*}{\textbf{R-precision$\uparrow$}}
& \multirow{2}{*}{\textbf{FID$\downarrow$}}
& \multirow{2}{*}{\textbf{CLIP score$\uparrow$}}
& \multicolumn{3}{c}{\textbf{SRA$\uparrow$}}
& \multirow{2}{*}{\textbf{Foot Skating Ratio$\downarrow$}} \\
\cmidrule(lr){6-8}
&  &  &  &
& \textbf{Top-1} & \textbf{Top-2} & \textbf{Top-3}
&  \\
\midrule
Real   & 1.2323  & 0.9519 & 0.3589 & 0.7432 & 0.9988 & 0.9997 & 1.0000 &  0.0000\\
 Ours    & 3.3034 & 0.6337 & 3.4981 & 0.5612 & 0.5328 & 0.6637 & 0.7272 & 0.0818 \\
 Sinusoidal PE & 3.4016	&0.6094	& 3.9069 & 0.5485	& 0.5291	&0.6516	& 0.7066	&0.1093\\
 UNet & 4.17 & 0.4313 & 2.6791 & 0.3571 & 0.1603 & 0.2228 & 0.2678 & 0.0621 \\
 No VAE & 3.4596	&	0.6038 &	4.532&	0.5554&	0.4984&	0.6353&	0.6919&	0.0987 \\
 ControlNet & 3.3843 & 0.6362 & 3.9498 & 0.562 & 0.5144 & 0.6788 & 0.7534 & 0.062 \\
 Cross Attn & 3.3119 & 0.6206 & 3.0984 & 0.5316 & 0.3938 & 0.5159 & 0.5884 & 0.2855 \\
 \rev{SASI} & \rev{3.4665} & \rev{0.5531} & \rev{1.9359} & \rev{0.4327} & \rev{0.3309} & \rev{0.4022} & \rev{0.4522} & \rev{0.0532} \\
 \rev{Key bias only} & \rev{1.2761} & \rev{0.9631} & \rev{0.3851} & \rev{0.7348} & \rev{0.0125} & \rev{0.0244} & \rev{0.0369} & \rev{0.0836} \\
 \rev{Value bias only} & \rev{3.3082} & \rev{0.6337} & \rev{3.5184} & \rev{0.5607} & \rev{0.5416} & \rev{0.6728} & \rev{0.7306} & \rev{0.0764} \\
\bottomrule
\end{tabular*}
\end{table}

\paragraph*{Positional encoding and network architecture.}
There are two popular families of network architectures for motion diffusion models: transformer-based
and UNet-based.
Replacing our relative position encoding (RPE) with sinusoidal position encoding degrades performance consistently across both settings—\textit{FID} rises from $3.08$ to $3.90$ (single style) and from $3.50$ to $3.91$ (time-varying), while \textit{R-Precision} and \textit{SRA} also drop in both cases. Notably, the foot skating gap widens markedly in the time-varying setting ($0.1093$ vs.\ $0.0818$), where the model must generalize to sequence segments with independently shifting styles, confirming that RPE is especially important for temporal generalization.
The UNet variant achieves a lower \textit{FID} and reduced foot skating in both settings, but consistently collapses style fidelity: top-3 \textit{SRA} falls to $0.3456$ (single style) and $0.2678$ (time-varying). Its hierarchical downsampling discards the fine-grained temporal structure needed for accurate style reflection and, in the time-varying setting, also degrades content fidelity (\textit{R-Precision} drops to $0.4313$). These results indicate that preserving \rev{temporal} resolution throughout the network is critical for both style accuracy and temporal flexibility.

\paragraph*{Motion VAE pretraining.}
Training the T2M model directly on raw motion features—without a pretrained motion VAE—consistently degrades all metrics in both settings: \textit{FID} rises to $4.34$ (single style) and $4.53$ (time-varying), \textit{R-Precision} drops by roughly $0.02$, and \textit{SRA} top-3 decreases by $0.02$–$0.04$. The VAE latent space provides a compact, structured representation that makes the diffusion model's denoising task more tractable and supports richer content–style disentanglement. Because our VAE preserves \rev{a temporally resolved latent} rather than collapsing the sequence to a single vector, SAM can inject style independently at each \rev{latent token}, enabling \rev{temporally resolved} content--style control.

\paragraph*{Style injection mechanism.}
We compare our Style Adaptation Module (SAM), which injects style as additive key/value biases, against \rev{three} alternatives: ControlNet\rev{,} cross-attention\rev{, and the SASI module of Wu \etal}.
ControlNet achieves slightly higher \textit{SRA} in both settings (top-3 $0.8531$ vs.\ $0.8356$ for single style; $0.7534$ vs.\ $0.7272$ for time-varying), but at the cost of higher \textit{FID} ($3.53$ vs.\ $3.08$; $3.95$ vs.\ $3.50$), as duplicating the entire backbone introduces excess capacity that overfits to style at the expense of motion quality.
Cross-attention performs comparably to our model in the single-style setting, but degrades severely under time-varying style input: \textit{SRA} top-3 falls to $0.5884$ (vs.\ $0.7272$) and foot skating spikes to $0.2855$ (vs.\ $0.0818$). This indicates that cross-attention struggles to inject style in a temporally resolved manner, coupling style tokens across frames and disrupting motion dynamics.
\rev{
Finally, we replace SAM with SASI, which is close in spirit to ours in that it also conditions on the
reference clip frame by frame, but forms the injected weights as a text-mediated attention between
content and reference frames instead of as additive key/value biases. We retrain it on the same
backbone, data and schedule, with its style guidance set so that its single-style top-1 \textit{SRA}
($0.6719$) matches ours ($0.6863$), so the two rows are compared at equal style accuracy; it trains
$14.25$M parameters against SAM's $6.44$M (Appendix~\ref{sup:params}).
Under single-style input SASI is the stronger of the two on content and quality, clearly so on
\textit{FID} ($1.05$ vs.\ $3.08$), while trailing on top-3 \textit{SRA} ($0.7662$ vs.\ $0.8356$).
The ordering reverses once the style varies over time: its top-1 \textit{SRA} drops to $0.3309$ (ours:
$0.5328$), its leakage is $0.4994$, and the requested switch never becomes detectable in $70.3\%$ of
sequences (ours: $0.1064$ and $9.4\%$); its low foot skating here ($0.0532$) follows from the same
behaviour, as a sequence that never changes style has no transition to disturb.
Text-mediated weights thus route style well when a single reference is in force, but do not by
themselves confine a reference to its assigned interval, which is what the explicit per-token
conditioning in SAM provides; Appendix~\ref{sup:vis-sasi} shows three qualitative examples.
The SASI variant was trained from an earlier motion VAE checkpoint and at half the batch size of the
other rows; both differences work against SASI and so do not account for its single-style advantages.
}
Our SAM achieves the best overall balance across both settings: strong content preservation, competitive style accuracy, and low foot skating, while \rev{training only its own $6.44$M parameters (Appendix~\ref{sup:params}), and among the four mechanisms it is the only one whose style accuracy does not collapse when the style varies over time}.

\rev{
\paragraph*{Key versus value bias.}
SAM biases both the keys and the values of the frozen backbone's attention. Restricting it to either
half alone, retrained with the same data, schedule and guidance and evaluated in both settings (last
two rows of Tables~\ref{tab:ablation_single_style} and~\ref{tab:ablation_time_varying_style}), shows
that the two are not symmetric. A key bias alone transfers no style in either setting: top-1
\textit{SRA} is at chance ($0.0156$ and $0.0125$) and every content metric returns to ground-truth
level. This is what the mechanism predicts, since reweighting which frames attend to which cannot
introduce into the residual stream any content that the values do not already carry. A value bias alone
matches the full form on every metric in both settings, in each case within one standard deviation over
the five repetitions. The value bias is thus the component that carries the style; we keep both for
consistency with the rest of the paper.
}

Overall, these results confirm that our design choices—RPE-based temporal attention, motion VAE pretraining, and SAM-based style injection—collectively contribute to both higher motion quality and better content–style flexibility, with the advantages being especially pronounced in the more challenging time-varying setting.

\rev{\subsection{User study}\label{sec:user_study}}
\rev{Appendix~\ref{sup:user-study} describes a user study in which $22$ participants made $330$
pairwise comparisons between our results and those of the three baselines, on content preservation,
style reflection and overall motion quality; our method is preferred on nearly every criterion, and the
two criteria on which a baseline wins are the ones its metrics already predict.}

\rev{
\subsection{Failure cases}\label{sec:failure}
Our two failure modes share a common origin: the further a style reference lies from what the style
encoder saw during pretraining, the less predictable the outcome becomes. The larger of the two is the
one analysed in Section~\ref{sec:generalization}: for references drawn from outside 100STYLE
altogether, transfer is hit or miss rather than uniformly degraded, the style being muted while the
content prompt is still followed. The second is milder and concerns references \emph{within} 100STYLE
whose character is carried by rapid or abrupt limb motion, \eg\ \textit{BeatChest}. Style fidelity
drops for these and the outputs tend to be over-smoothed, suppressing exactly the high-frequency limb
dynamics that define the style; an example is shown in Appendix~\ref{sup:vis-abrupt}, where
the arms follow the reference's chest-striking path but at a visibly gentler pace.
This second mode is a limitation of what the model reproduces rather
than of what it recognizes: unlike the external references, these styles are represented in the style
latent space, but the diffusion backbone averages away their fastest components. The two therefore
call for different remedies --- widening the style encoder's pretraining distribution to
non-locomotion and multi-subject data in the first case, which is limited today by the scarcity of
stylized motion data of sufficient quality, and supervising motion in the frequency domain
(Section~\ref{sec:conclusion}) in the second.
}

\section{Conclusion}
\label{sec:conclusion}

We presented \textbf{FlexMoGen}, a framework for controllable stylized motion synthesis that generates high-quality human motions from text prompts and example style clips.
By combining a label-free variational style encoder, a lightweight Style Adaptation Module that conditions a pretrained T2M diffusion model via key–value attention biases, and relative position encoding for fine-grained temporal control, FlexMoGen supports time-varying style inputs and long-sequence generation that prior methods cannot handle.
\rev{Experiments confirm that FlexMoGen strikes the best balance between style fidelity and content preservation across our evaluation settings, and that it leads on nearly every metric in the time-varying setting that prior methods do not natively support.}

Current limitations include a tendency toward over-smoothed outputs that suppress rapid limb movements, and reduced style fidelity on fast or abrupt style examples (\eg, \textit{BeatChest})\rev{, together with the unreliable transfer of style references drawn from outside the style encoder's pretraining distribution (Section~\ref{sec:generalization}); Section~\ref{sec:failure} discusses both}.
To mitigate over-smoothing, a promising direction is to supervise motion in the frequency domain. This can be a spectral reconstruction loss that emphasizes high-frequency bands, or an explicit low/high-frequency decomposition so that rapid limb dynamics are preserved rather than averaged away.
Additionally, diffusion-based sampling is too slow for real-time use; flow-matching models, which achieve comparable quality with far fewer function evaluations, are a promising avenue for future work.

\section*{Acknowledgements}
The contributions of Kai Weixian Lan, Bodie Criswell, and Briana Fedkiw to this work were made during internships at Epic Games.

\clearpage
\bibliographystyle{plainnat}
\bibliography{refs}

\clearpage
\appendix

\begingroup
\centering
{\LARGE\sc Supplementary Material}\\[6pt]
\endgroup

The following appendices contain the supplementary material that accompanies the
paper. Appendix~\ref{sup:vis} collects additional qualitative results,
Appendices~\ref{sup:localization} and~\ref{sup:long-multistyle} the
temporal-localization and long-sequence evaluations summarized in
Section~\ref{sec:more_eval}, Appendix~\ref{sup:user-study} the user study, and
Appendices~\ref{sup:pseudocode} to~\ref{sup:bodypart} the pseudo code,
implementation details, dataset composition and skeleton definition.
A supplementary video accompanies these results.

\section{More visualizations}\label{sup:vis}
We include additional visualizations of the generated motions to further illustrate the behavior of our model. A supplementary video is also provided for clearer and more intuitive inspection of the results.
\rev{We group them by the setting they illustrate: a single style applied to the whole sequence
(Section~\ref{sup:vis-single}), a style that changes over time (Section~\ref{sup:vis-varying}),
style references drawn from outside the 100STYLE dataset (Section~\ref{sup:vis-ood}), and style codes
assembled per body part from several references (Section~\ref{sup:vis-bodypart}).
Section~\ref{sup:vis-sasi} shows the SASI variant of the injection-mechanism ablation instead of our
own model, and Section~\ref{sup:vis-abrupt} a failure case on a style carried by rapid limb motion.}

\rev{\subsection{Single style}\label{sup:vis-single}}
\rev{Each of the following figures applies one style to the whole sequence and compares all four
methods on the same text prompt.}

\begin{figure}[H]
    \centering
    \textbf{\small Text prompt:} \textit{Backpedaling}\\[2pt]
    \textbf{\small Style:} \textit{Dinosaur}\\[2pt]
    \begin{subfigure}[b]{0.19\textwidth}
        \includegraphics[width=\textwidth]{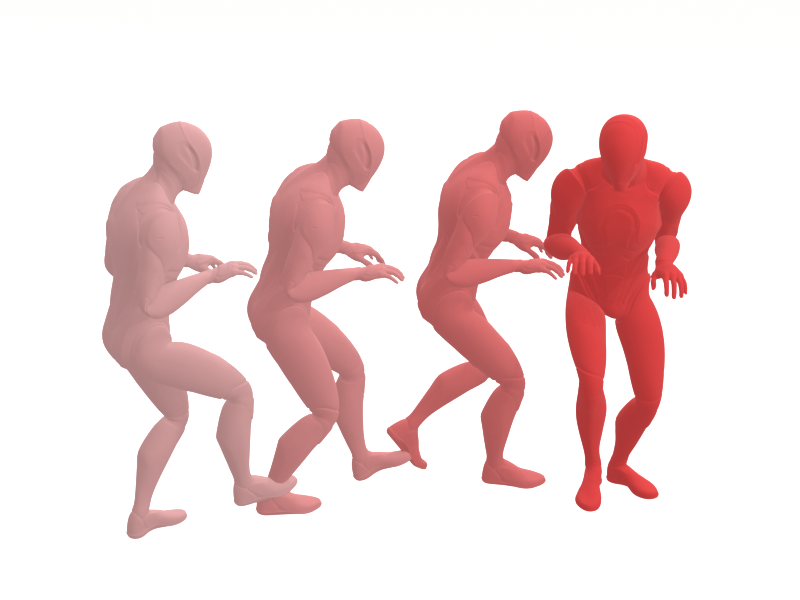}
        \caption{\textit{Dinosaur} style}
    \end{subfigure}
    \hfill
    \begin{subfigure}[b]{0.19\textwidth}
        \includegraphics[width=\textwidth]{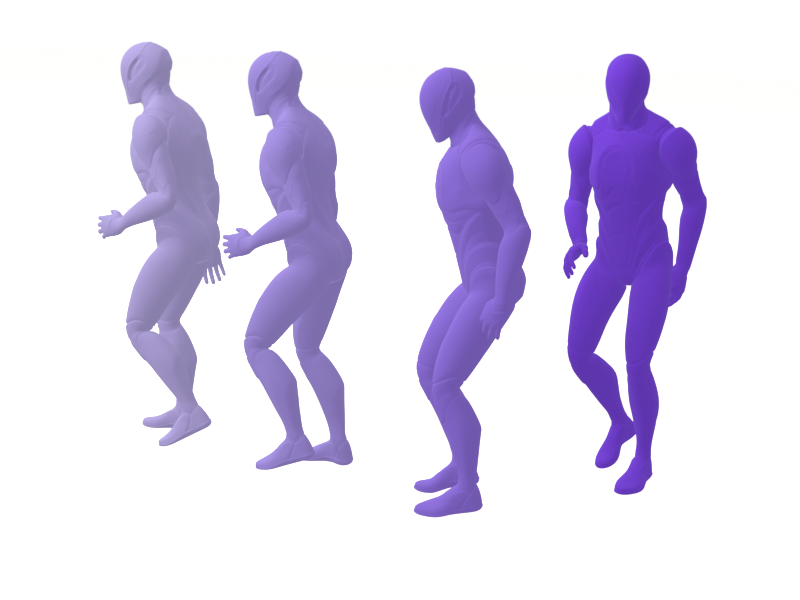}
        \caption{LoRA-MDM}
    \end{subfigure}
    \hfill
    \begin{subfigure}[b]{0.19\textwidth}
        \includegraphics[width=\textwidth]{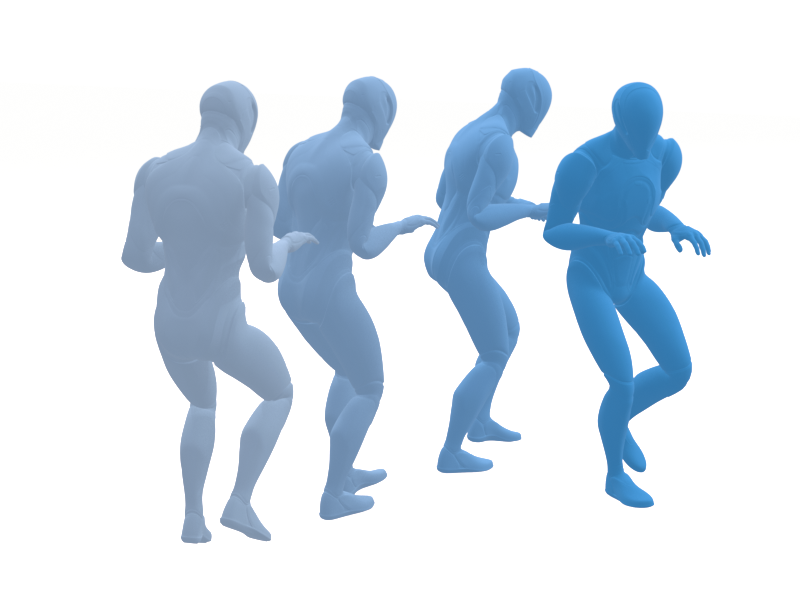}
        \caption{T2M$+$MotionPuzzle}
    \end{subfigure}
    \hfill
    \begin{subfigure}[b]{0.19\textwidth}
        \includegraphics[width=\textwidth]{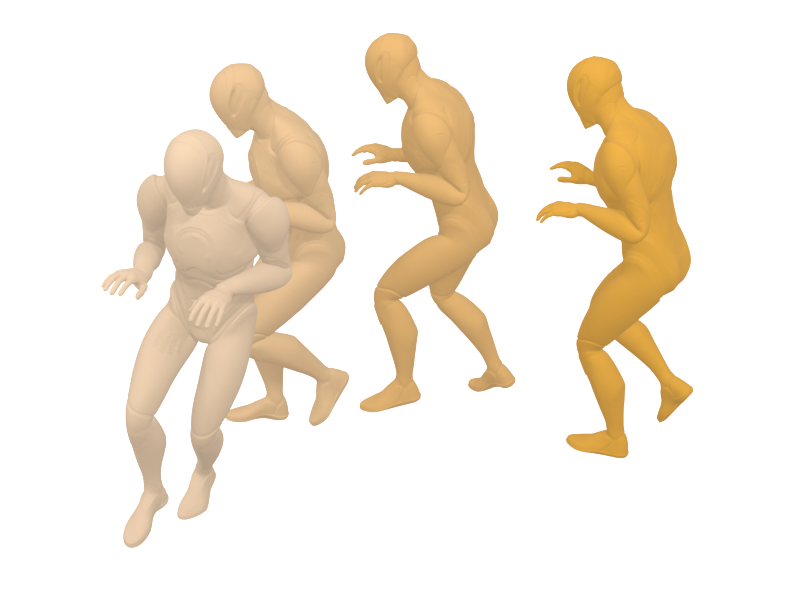}
        \caption{SMooDi}
    \end{subfigure}
    \hfill
    \begin{subfigure}[b]{0.19\textwidth}
        \includegraphics[width=\textwidth]{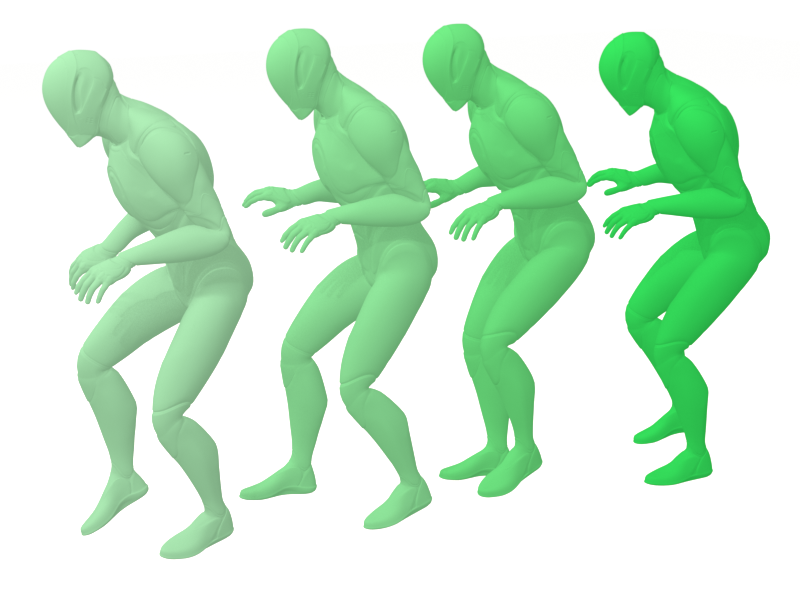}
        \caption{Ours}
    \end{subfigure}
    \caption{Comparison of different models on the text prompt ``backpedaling'' with the \textit{Dinosaur} style (bent arms raised in front of the chest, bent torso and knees). Only our model backpedals in the correct direction without erratic movements, while reflecting the style accurately. Colors transition from light to dark to indicate earlier to later frames in the sequence.}
    \label{fig:backpedal_dinosaur}
\end{figure}

\begin{figure}[H]
    \centering
    \textbf{\small Text prompt:} \textit{Spinning Clockwise}\\[2pt]
    \textbf{\small Style:} \textit{ArmsAboveHead}\\[2pt]
    \begin{subfigure}[b]{0.19\textwidth}
        \includegraphics[width=\textwidth]{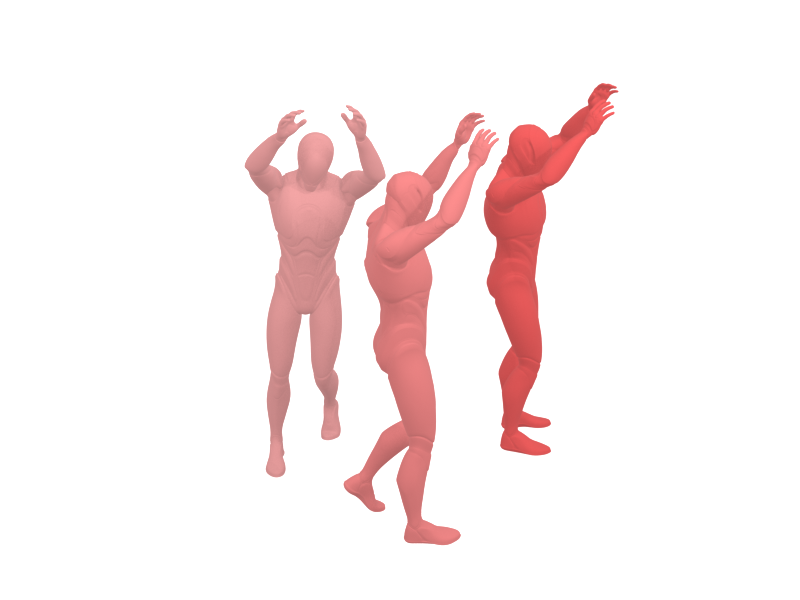}
        \caption{\textit{ArmsAboveHead} style}
    \end{subfigure}
    \hfill
    \begin{subfigure}[b]{0.19\textwidth}
        \includegraphics[width=\textwidth]{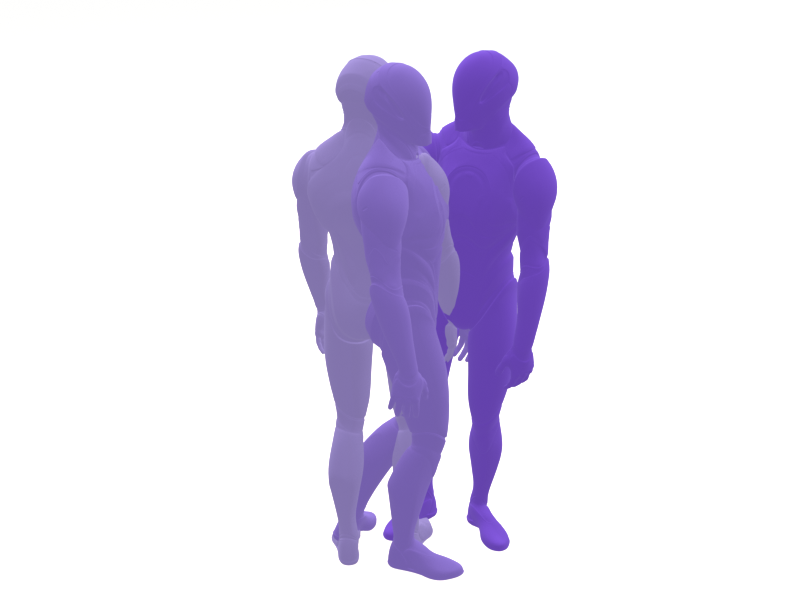}
        \caption{LoRA-MDM}
    \end{subfigure}
    \hfill
    \begin{subfigure}[b]{0.19\textwidth}
        \includegraphics[width=\textwidth]{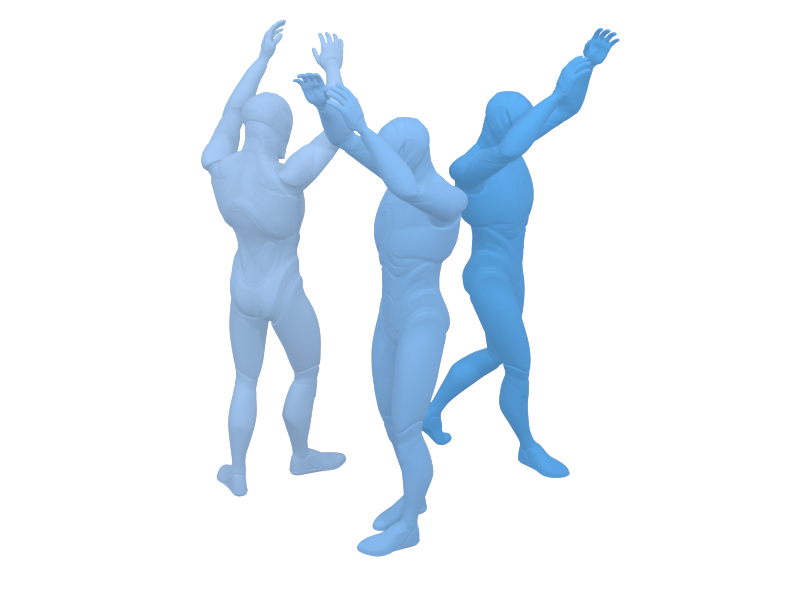}
        \caption{T2M$+$MotionPuzzle}
    \end{subfigure}
    \hfill
    \begin{subfigure}[b]{0.19\textwidth}
        \includegraphics[width=\textwidth]{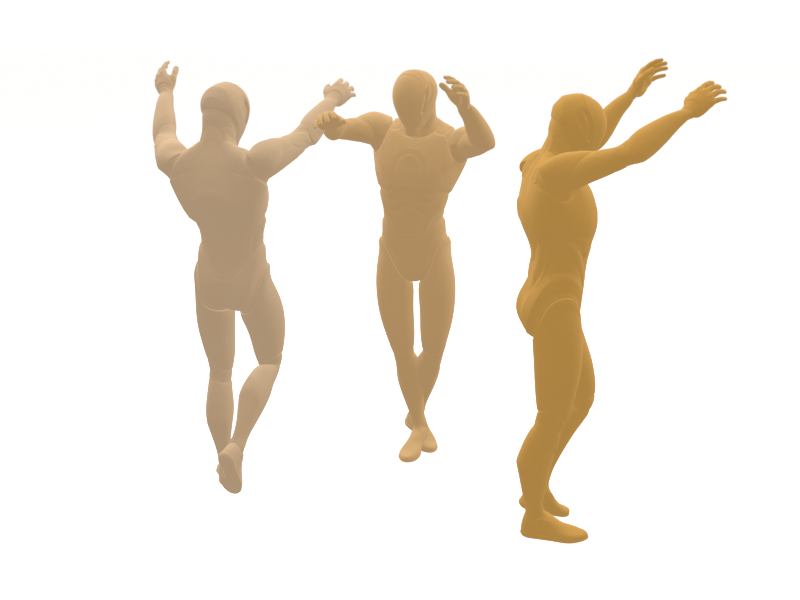}
        \caption{SMooDi}
    \end{subfigure}
    \hfill
    \begin{subfigure}[b]{0.19\textwidth}
        \includegraphics[width=\textwidth]{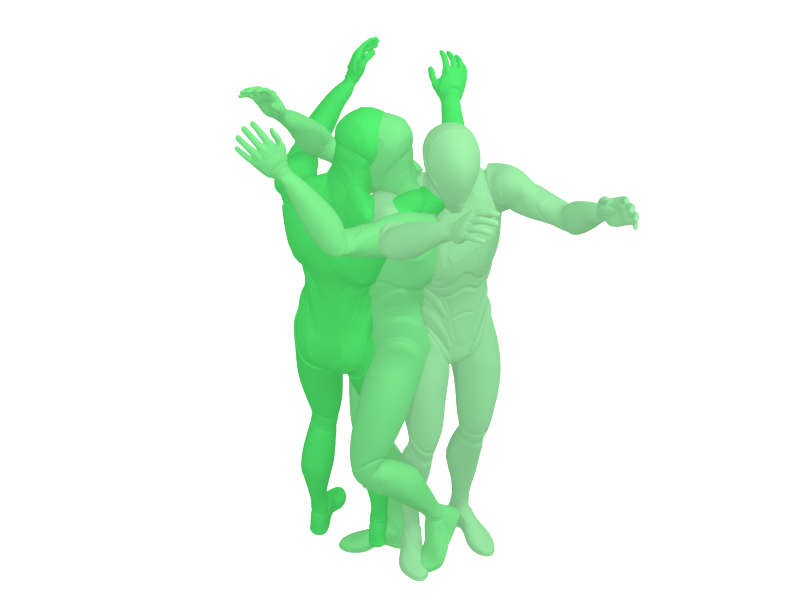}
        \caption{Ours}
    \end{subfigure}
   \caption{Comparison of different models on the text prompt ``Spinning clockwise'' with the \textit{ArmsAboveHead} style. Only our model generates a motion where the character spins clockwise with raised arms without introducing erratic artifacts such as unintended translation.}
    \label{fig:spin_armsabovehead}
\end{figure}
\begin{figure}[H]
    \centering
    \textbf{\small Text prompt:} \textit{Standing long jump}\\[2pt]
    \textbf{\small Style:} \textit{Chicken}\\[2pt]
    \begin{subfigure}[b]{0.19\textwidth}
        \includegraphics[width=\textwidth]{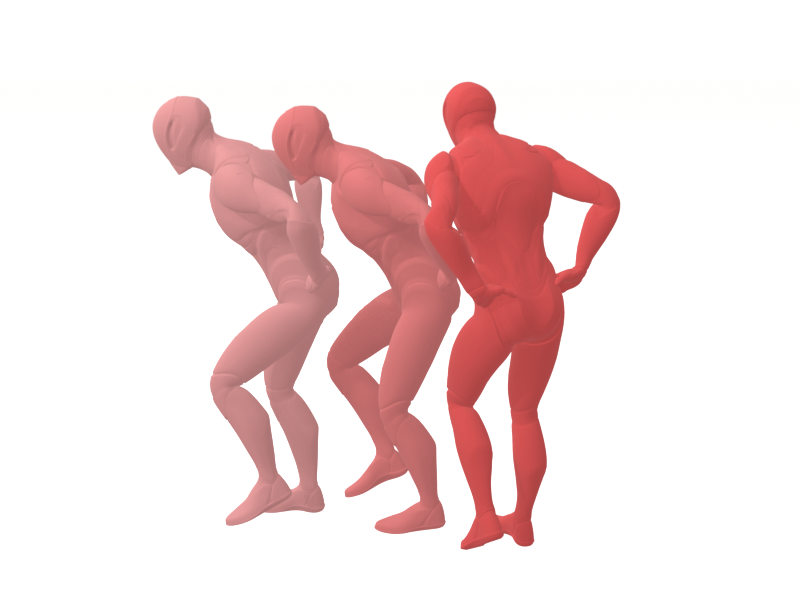}
        \caption{\textit{Chicken} style}
    \end{subfigure}
    \hfill
    \begin{subfigure}[b]{0.19\textwidth}
        \includegraphics[width=\textwidth]{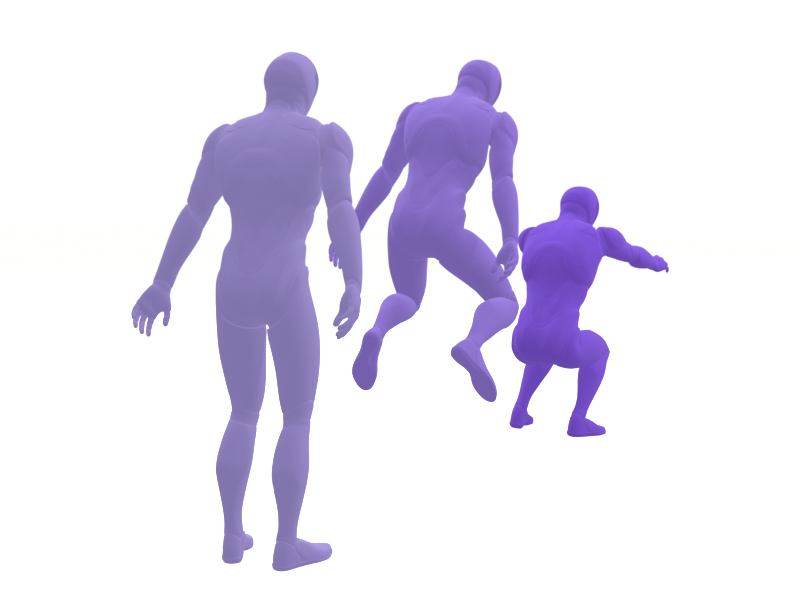}
        \caption{LoRA-MDM}
    \end{subfigure}
    \hfill
    \begin{subfigure}[b]{0.19\textwidth}
        \includegraphics[width=\textwidth]{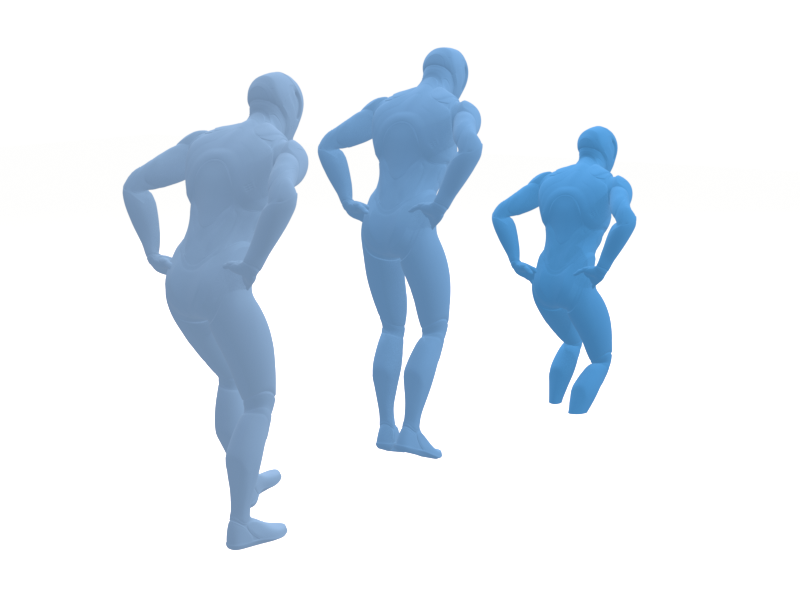}
        \caption{T2M$+$MotionPuzzle}
    \end{subfigure}
    \hfill
    \begin{subfigure}[b]{0.19\textwidth}
        \includegraphics[width=\textwidth]{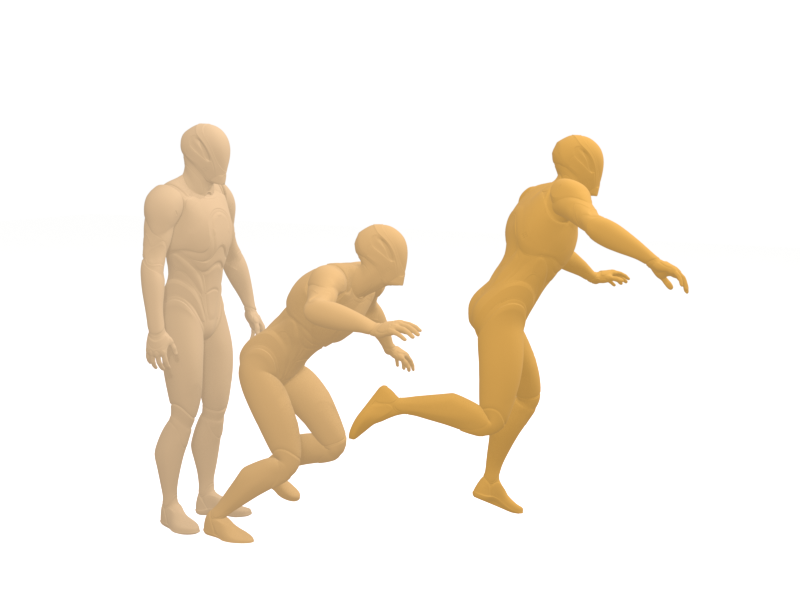}
        \caption{SMooDi}
    \end{subfigure}
    \hfill
    \begin{subfigure}[b]{0.19\textwidth}
        \includegraphics[width=\textwidth]{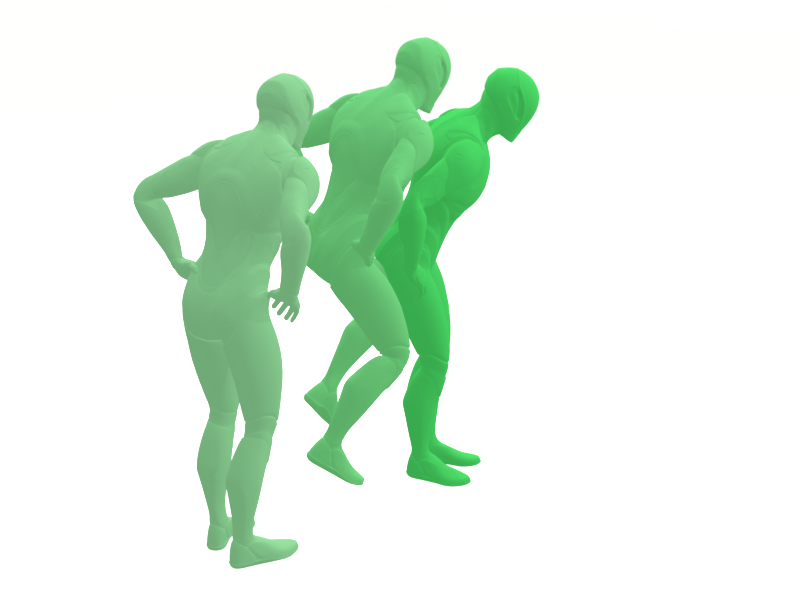}
        \caption{Ours}
    \end{subfigure}
   \caption{Comparison of different models on the text prompt ``Standing long jump'' with the \textit{chicken} style. Only our model produces a clean jump without visible artifacts such as ground penetration or unnecessary extra steps.}
    \label{fig:jump_chicken}

\end{figure}

\rev{\subsection{Time-varying style}\label{sup:vis-varying}}
\rev{Here the style changes partway through the sequence, alongside a change of action.}

\begin{figure}[H]
    \centering

    \begin{subfigure}{0.2\linewidth}
        \centering
        \includegraphics[width=\linewidth]{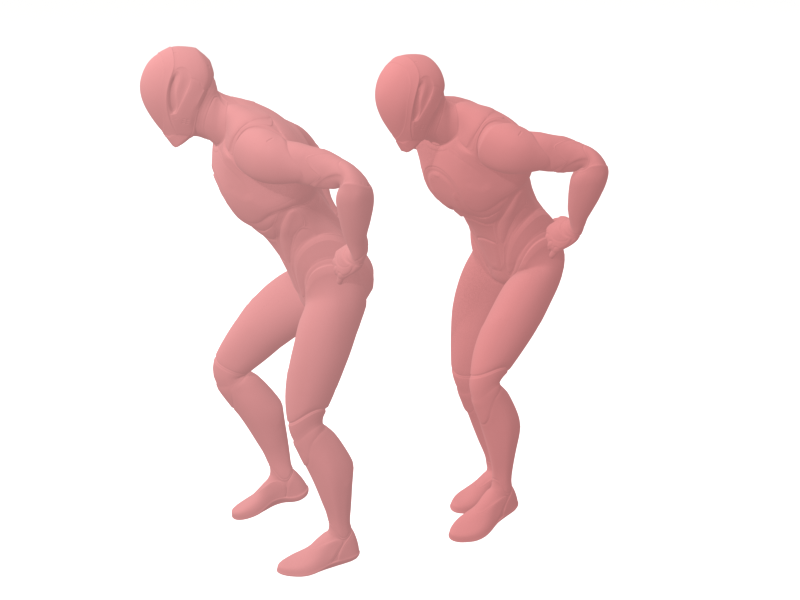}
        \caption{\textit{Chicken} style}
    \end{subfigure}
    \begin{subfigure}{0.2\linewidth}
        \centering
        \includegraphics[width=\linewidth]{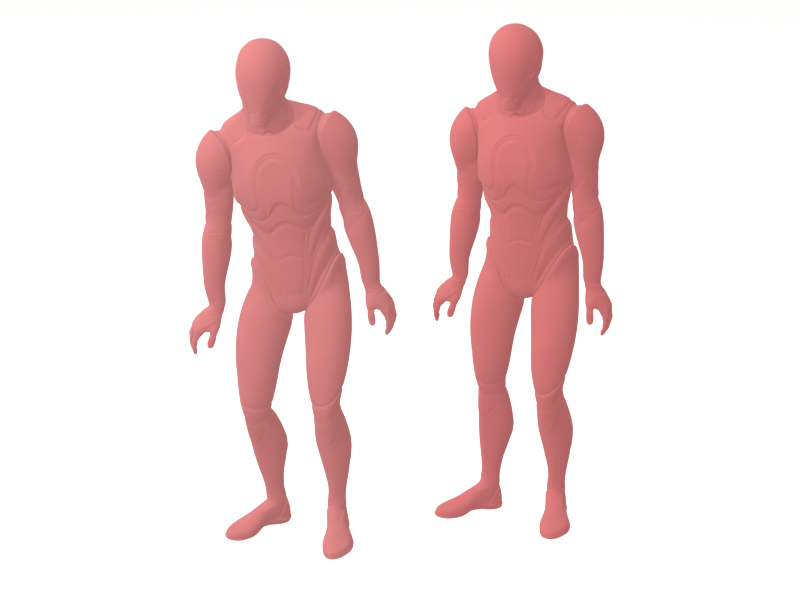}
        \caption{\textit{Penguin} style}
    \end{subfigure}
    \begin{subfigure}{0.2\linewidth}
        \centering
        \includegraphics[width=\linewidth]{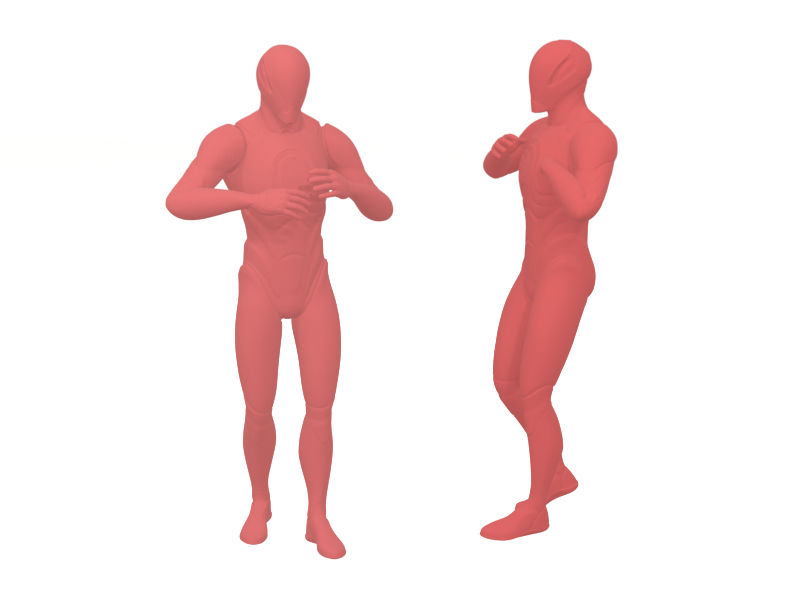}
        \caption{\textit{BeatChest} style}
    \end{subfigure}

    \begin{subfigure}{0.2\linewidth}
        \centering
        \includegraphics[width=\linewidth]{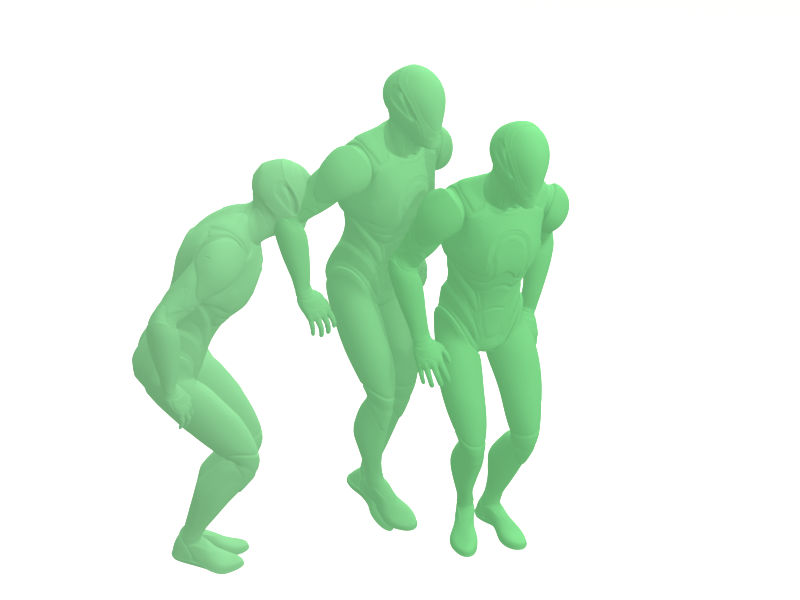}
    \end{subfigure}
    \begin{subfigure}{0.2\linewidth}
        \centering
        \includegraphics[width=\linewidth]{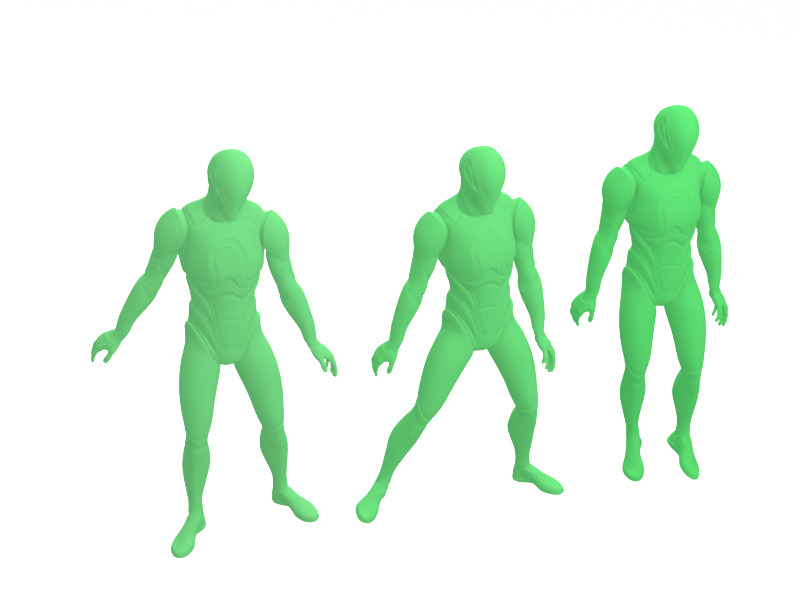}
    \end{subfigure}
    \begin{subfigure}{0.2\linewidth}
        \centering
        \includegraphics[width=\linewidth]{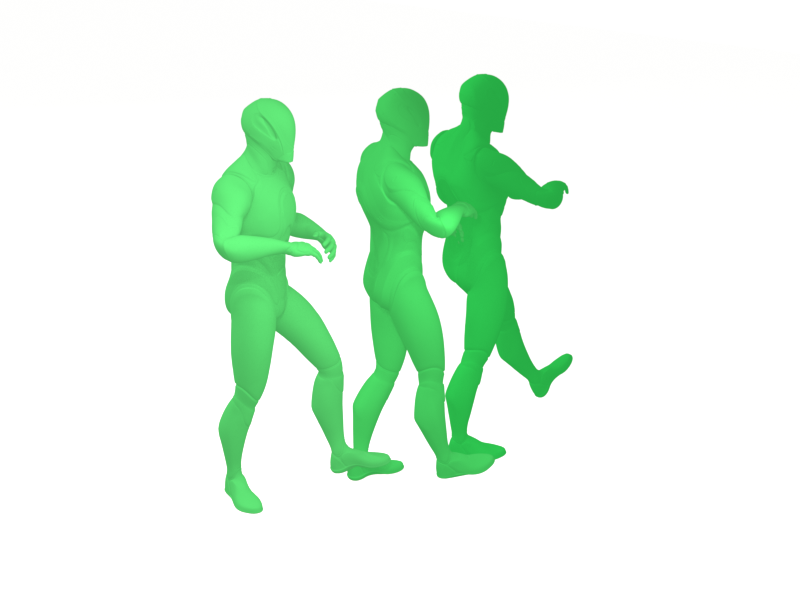}
    \end{subfigure}

    \caption{Simultaneous action and style transitions. \textbf{Top row:} style reference clips for \textit{Chicken} (arms tucked, bent torso), \textit{Penguin} (stiff upright posture, arms close to body), and \textit{BeatChest} (arms striking chest), applied sequentially. \textbf{Bottom row:} FlexMoGen output for the text prompt ``Jump with both feet, then side hop, then kick legs'', generated under the corresponding time-varying styles. Our model accurately performs all three actions in the correct temporal order while faithfully reflecting each style transition. \textit{BeatChest} is an out-of-distribution style unseen during training.}
    \label{fig:vis}
\end{figure}

\rev{\subsection{Styles outside 100STYLE}\label{sup:vis-ood}}
\rev{The references below are not from 100STYLE at all, and illustrate the failure mode summarized in
Section~\ref{sec:generalization}. They are taken from an internally captured set of stylized motions whose
subject, capture setup and vocabulary all differ substantially from 100STYLE, in particular in
consisting largely of upper-body and prop-handling actions rather than variations of walking. This is a
setting that cannot be scored, since \textit{SRA} is defined only over the labelled 100STYLE taxonomy
and no classifier exists for an arbitrary reference; we therefore report it qualitatively and draw no
quantitative conclusion from it.}

\begin{figure}[H]
    \centering
    \begin{subfigure}{0.48\linewidth}
        \centering
        \includegraphics[height=3.9cm]{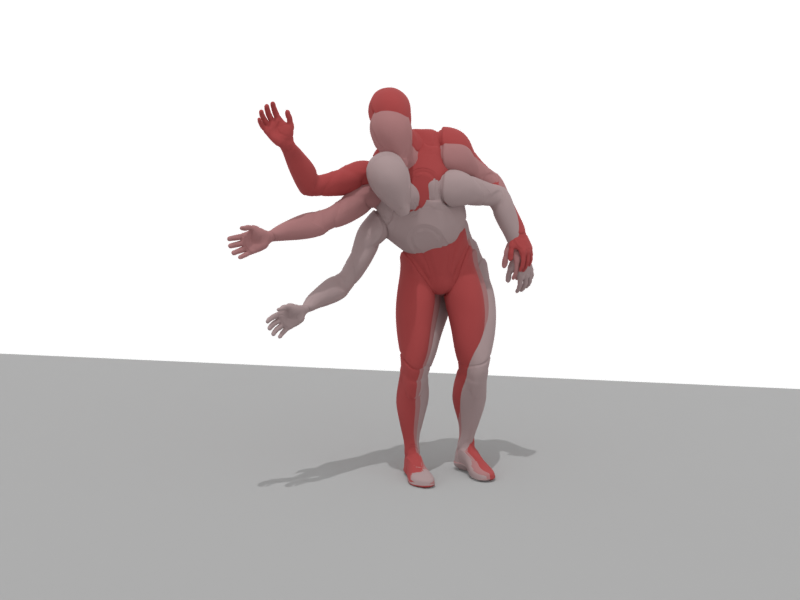}
        \caption{Out-of-distribution style\rev{: \textit{Reaching and waving}}}
    \end{subfigure}
    \hfill
    \begin{subfigure}{0.48\linewidth}
        \centering
        \includegraphics[height=3.9cm]{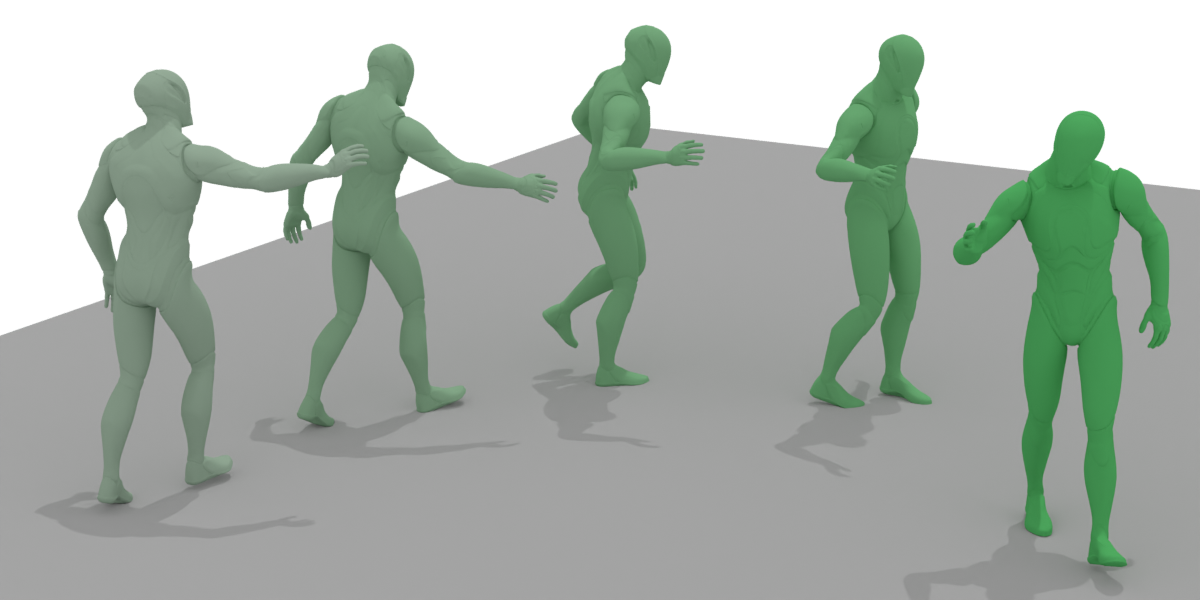}
        \caption{Ours}
    \end{subfigure}

    \vspace{4pt}

    \begin{subfigure}{0.48\linewidth}
        \centering
        \includegraphics[height=3.9cm]{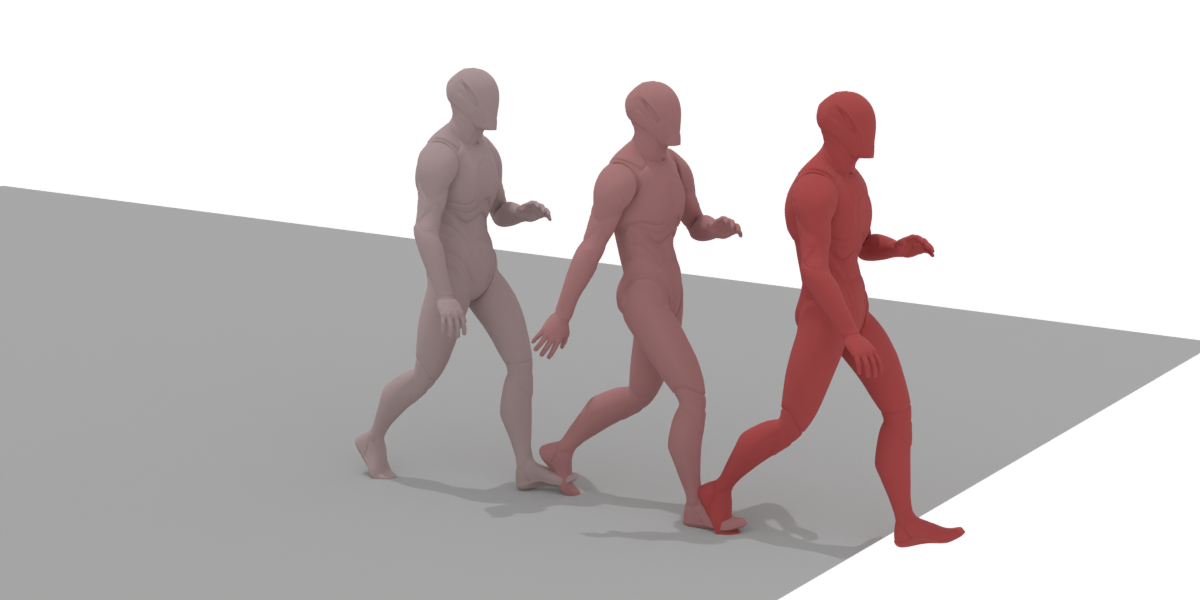}
        \caption{\rev{Out-of-distribution style: \textit{Holding a soda can}}}
    \end{subfigure}
    \hfill
    \begin{subfigure}{0.48\linewidth}
        \centering
        \includegraphics[height=3.9cm]{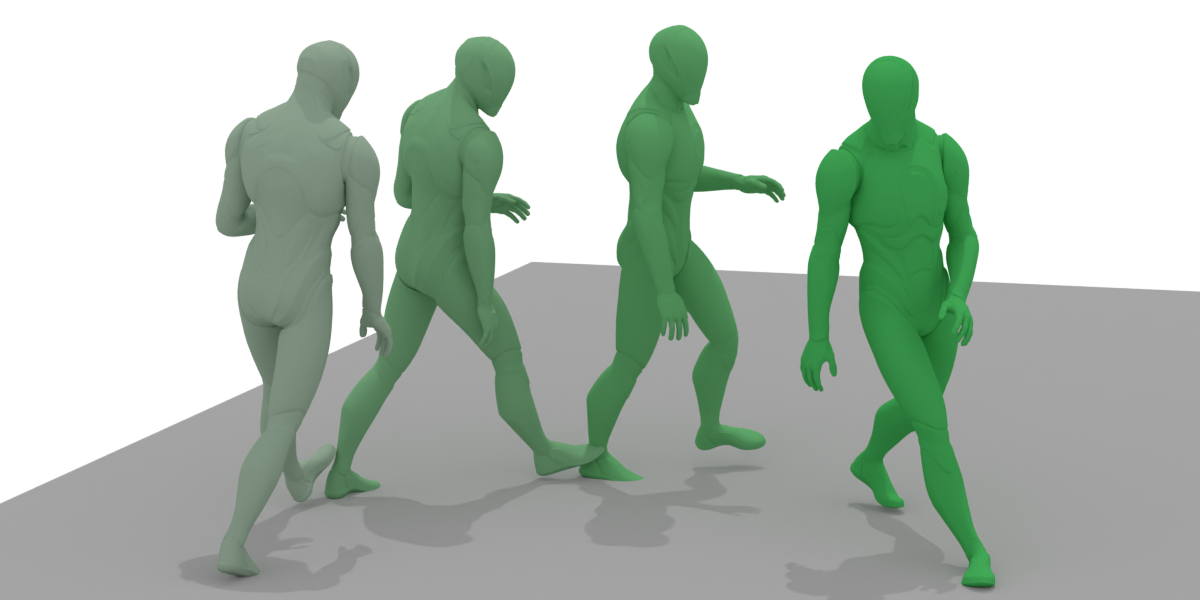}
        \caption{\rev{Ours}}
    \end{subfigure}

    \vspace{4pt}

    \begin{subfigure}{0.48\linewidth}
        \centering
        \includegraphics[height=3.9cm]{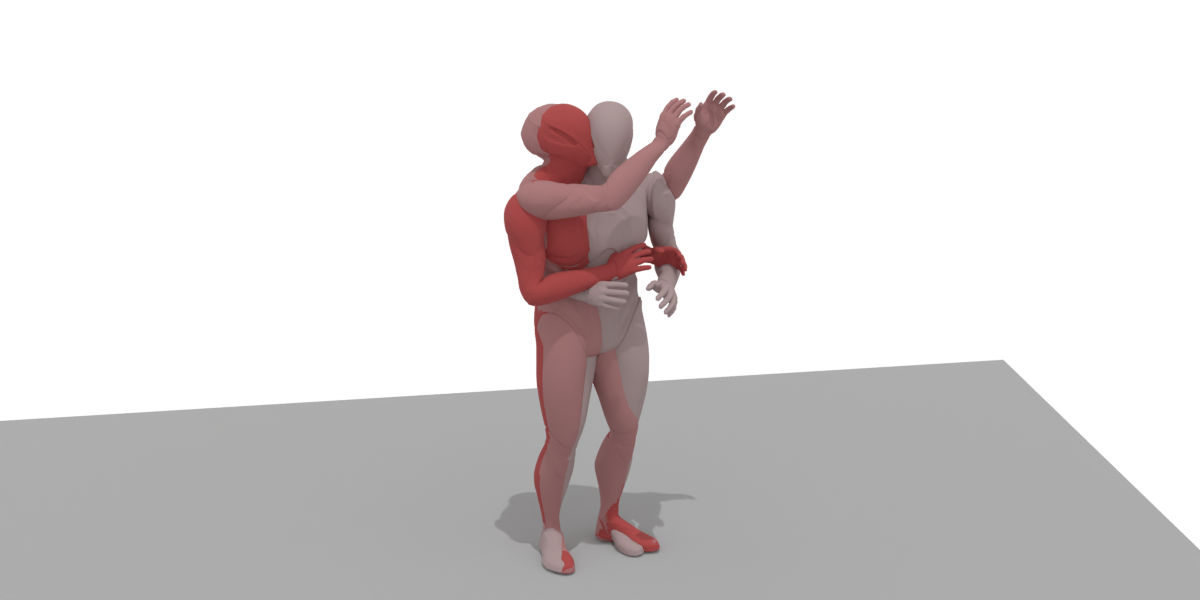}
        \caption{\rev{Out-of-distribution style: \textit{Taking a photo}}}
    \end{subfigure}
    \hfill
    \begin{subfigure}{0.48\linewidth}
        \centering
        \includegraphics[height=3.9cm]{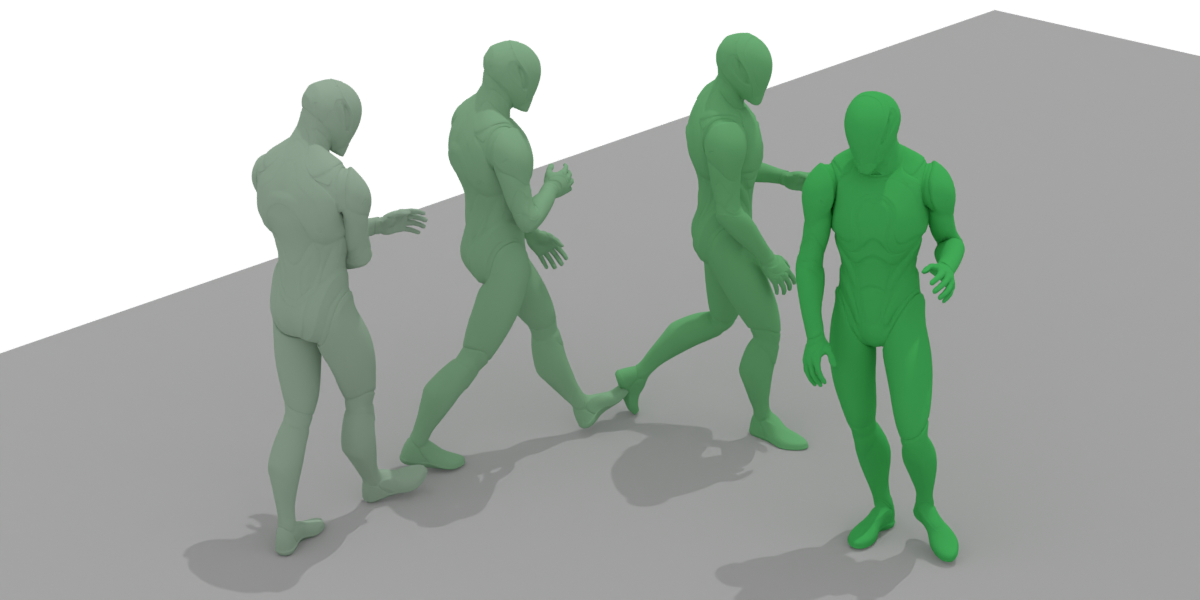}
        \caption{\rev{Ours}}
    \end{subfigure}
    \caption{Generation with \rev{three} style reference\rev{s} outside the 100STYLE
    dataset, given the text prompt \textit{``walking clockwise in a circle''}.
    \rev{The references are ordered from the most to the least successful transfer.
    \textit{Reaching and waving} (a) and \textit{Holding a soda can} (c) transfer
    successfully: the raised waving arm and the left arm carried bent against the
    chest both appear in our outputs (b, d), while the circular path asked for by
    the prompt is preserved. \textit{Taking a photo} (e) is a failure case: the reference
    raises both hands to head height, and while our output (f) does move the arms
    it never adopts that two-handed pose, which is what carries the style.
    Together these illustrate that transfer in this setting is hit or miss rather
    than uniformly degraded, as discussed in Section~\ref{sec:generalization}.}}
    \label{fig:ood_walk_clockwise}
\end{figure}

\rev{Transfer here is markedly less reliable than within 100STYLE, and it fails unevenly rather than
uniformly: with no change of settings, the same model reflects some external references about as clearly
as an in-domain style while reflecting others only partly, or barely at all. Which of the two happens is
difficult to anticipate from the reference alone, so the practical character of this setting is hit or
miss rather than a uniform loss of quality. The three references of
Figure~\ref{fig:ood_walk_clockwise} span that range, all driven by the same content prompt.
\textit{Reaching and waving} and \textit{Taking a photo} are similar in kind
--- both are stationary upper-body gestures with almost no locomotion of their own --- and \textit{Taking
a photo} differs mainly in being the faster and more articulated of the two, yet only one of them
transfers. That two such close references behave so differently is precisely why we describe this setting
as hit or miss rather than as a predictable loss of fidelity. The failures are graceful in one respect:
the content prompt continues to be followed and the motion remains plausible, so what is lost is the
stylistic character of the reference rather than the motion itself, and the result reads as a muted
version of the intended style.}

\rev{We attribute this to the style encoder's pretraining distribution: its clips are locomotion styles
from a single capture source (Section~\ref{pretraining_se}), so a reference whose body proportions,
retargeting or motion vocabulary fall outside that range is not well represented in the style latent
space, leaving the adaptation module little to inject. Accordingly, we claim generalization to unseen
styles only within the distribution the style encoder was trained on, and regard references beyond it as
a current limitation rather than a supported use case. Widening that distribution --- pretraining the
style encoder on non-locomotion and multi-subject data --- is the most direct remedy, and is limited
today by the scarcity of stylized motion data of sufficient quality.}

\begin{figure}[H]
    \centering
    \revcolor
    \textbf{\small Text prompt:} \textit{Walking forward then turning left}\\[2pt]
    \begin{subfigure}[b]{0.23\linewidth}
        \centering
        \includegraphics[width=\linewidth]{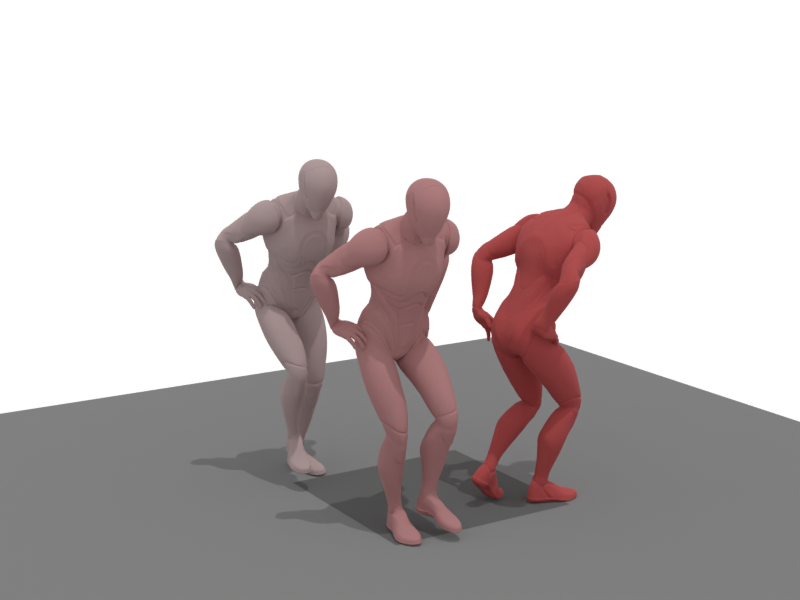}
        \caption{\rev{\textit{Chicken}}}
    \end{subfigure}
    \hfill
    \begin{subfigure}[b]{0.23\linewidth}
        \centering
        \includegraphics[width=\linewidth]{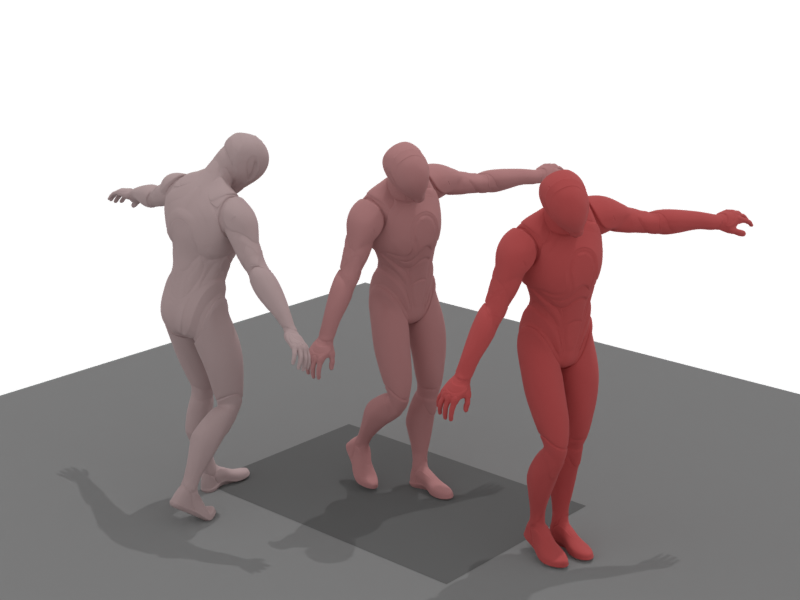}
        \caption{\rev{\textit{Aeroplane}}}
    \end{subfigure}
    \hfill
    \begin{subfigure}[b]{0.23\linewidth}
        \centering
        \includegraphics[width=\linewidth]{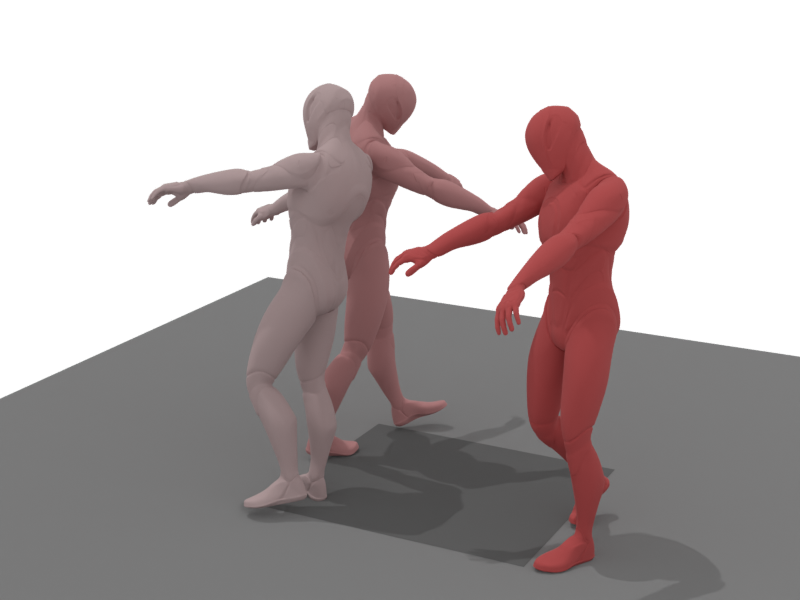}
        \caption{\rev{\textit{Zombie}}}
    \end{subfigure}
    \hfill
    \begin{subfigure}[b]{0.23\linewidth}
        \centering
        \includegraphics[width=\linewidth]{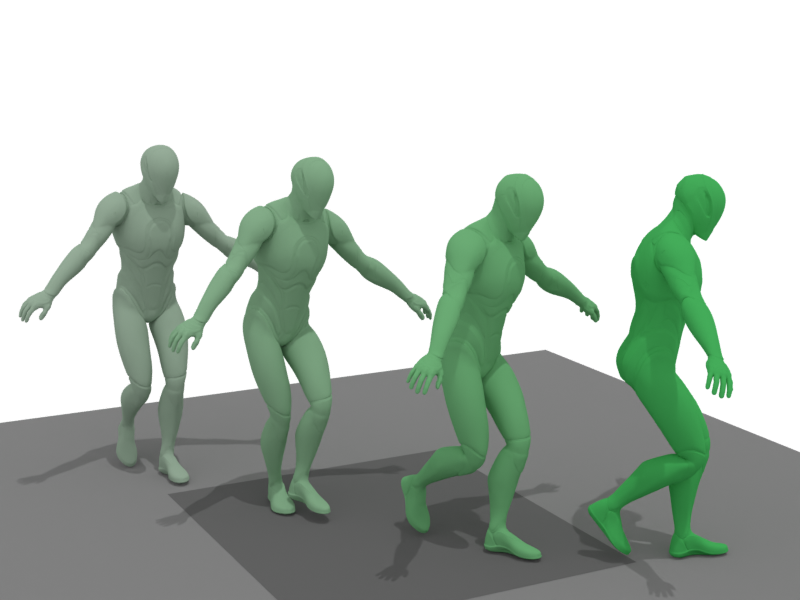}
        \caption{\rev{Ours}}
    \end{subfigure}
    \caption{\rev{Body-part style mixing with three style references, supplying the style embedding
    of the legs (a), of both arms (b) and of the spine (c). Our model generates a single motion (d)
    carrying all three at once: the crouched, bent-knee gait of \textit{Chicken}, the arms held out
    away from the body of \textit{Aeroplane}, and the forward-hunched torso of \textit{Zombie}.
    Colors transition from light to dark to indicate earlier to later frames.}}
    \label{fig:bodypart_three}
\end{figure}

\begin{figure}[H]
    \centering
    \revcolor
    \textbf{\small Text prompt:} \textit{Walking forward then turning left}\\[2pt]
    \begin{subfigure}[b]{0.19\linewidth}
        \centering
        \includegraphics[width=\linewidth]{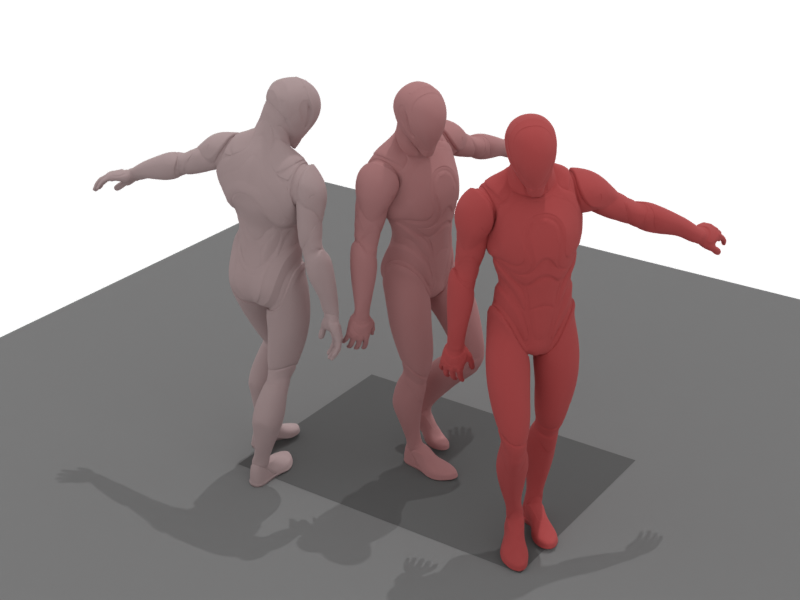}
        \caption{\rev{\textit{RaisedLeftArm}}}
    \end{subfigure}
    \hfill
    \begin{subfigure}[b]{0.19\linewidth}
        \centering
        \includegraphics[width=\linewidth]{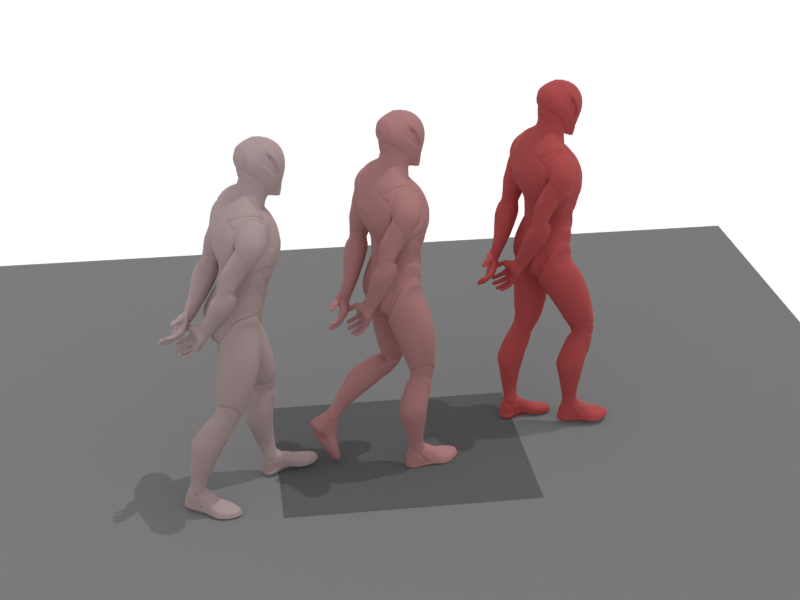}
        \caption{\rev{\textit{ArmsBehindBack}}}
    \end{subfigure}
    \hfill
    \begin{subfigure}[b]{0.19\linewidth}
        \centering
        \includegraphics[width=\linewidth]{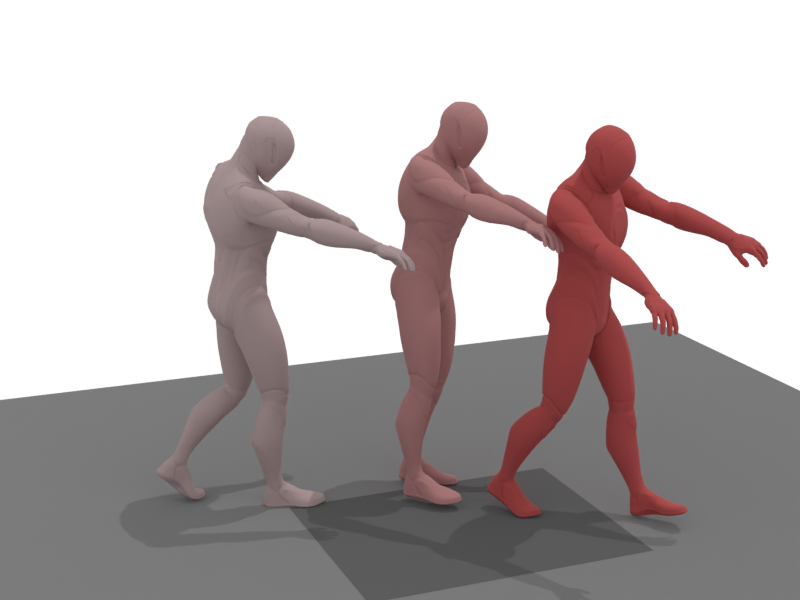}
        \caption{\rev{\textit{Zombie}}}
    \end{subfigure}
    \hfill
    \begin{subfigure}[b]{0.19\linewidth}
        \centering
        \includegraphics[width=\linewidth]{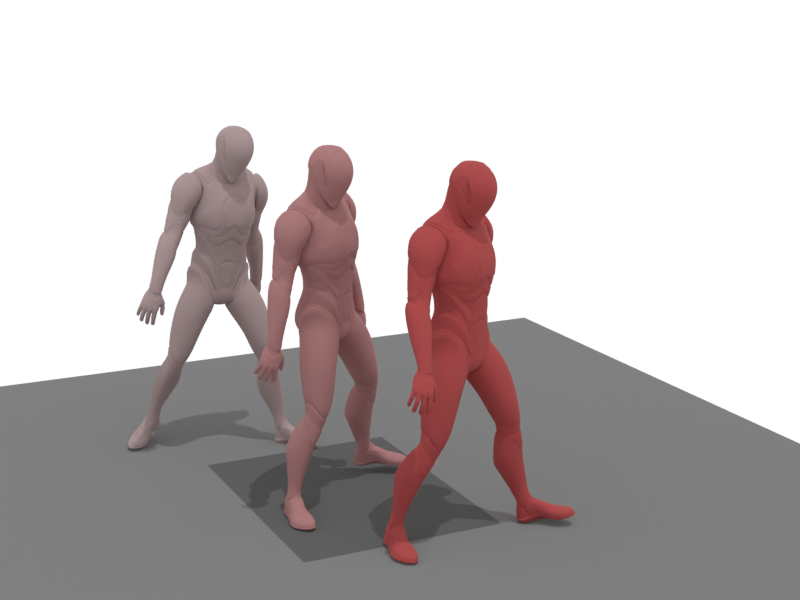}
        \caption{\rev{\textit{LegsApart}}}
    \end{subfigure}
    \hfill
    \begin{subfigure}[b]{0.19\linewidth}
        \centering
        \includegraphics[width=\linewidth]{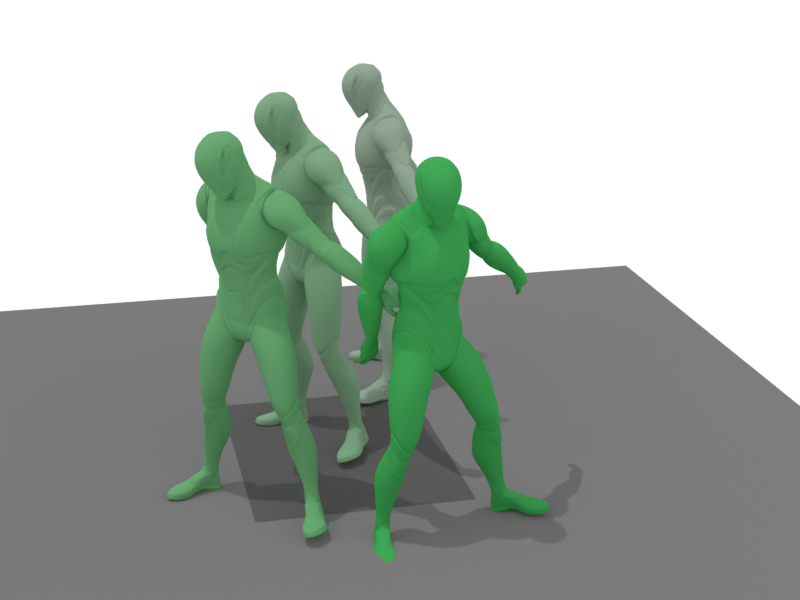}
        \caption{\rev{Ours}}
    \end{subfigure}
    \caption{\rev{Body-part style mixing with four style references, supplying the left arm (a), the
    right arm (b), the spine (c) and the legs (d) --- the two arms drawn from two different clips.
    The generated motion (e) is correspondingly asymmetric between the two arms, one carried up and
    away from the body while the other stays drawn in behind the torso, over the wide stance of
    \textit{LegsApart} and the forward lean of \textit{Zombie}. No single style reference provides
    this combination.}}
    \label{fig:bodypart_four}
\end{figure}

\rev{\subsection{Body-part style mixing}\label{sup:vis-bodypart}}
\rev{These are the two compositions discussed in Section~\ref{sec:bodypart_mixing}, in which the
components of the style code are drawn from different reference clips and assembled before injection.
The part groups are those of Section~\ref{sup:bodypart}.
Figure~\ref{fig:bodypart_three} composes three references, one per part group.
Figure~\ref{fig:bodypart_four} composes four, assigning \emph{different} styles to the left and right
arms, so that the generated motion is asymmetric between the two arms.
Animated versions of both are in the supplementary video.}

\begin{figure}[H]
    \centering
    \revcolor
    \begin{subfigure}{0.48\linewidth}
        \centering
        \includegraphics[height=3.9cm]{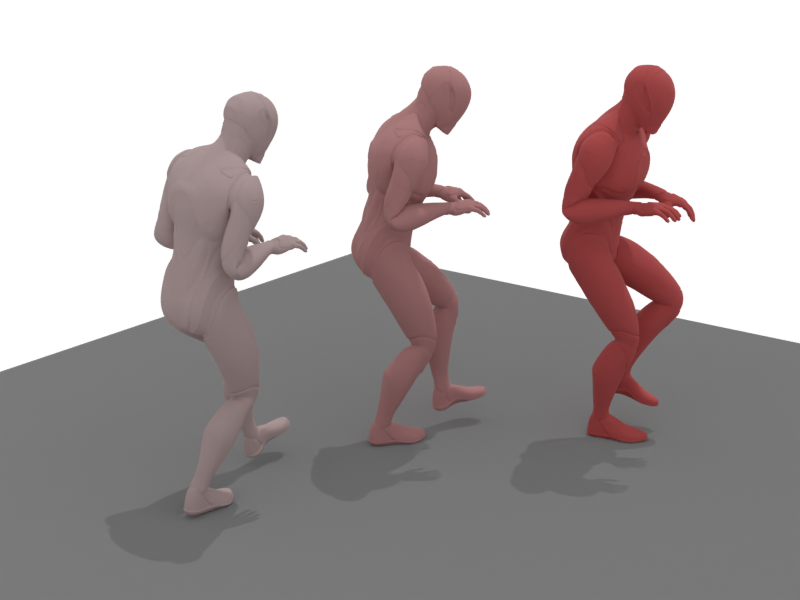}
        \caption{\rev{\textit{Dinosaur} style reference}}
    \end{subfigure}
    \hfill
    \begin{subfigure}{0.48\linewidth}
        \centering
        \includegraphics[height=3.9cm]{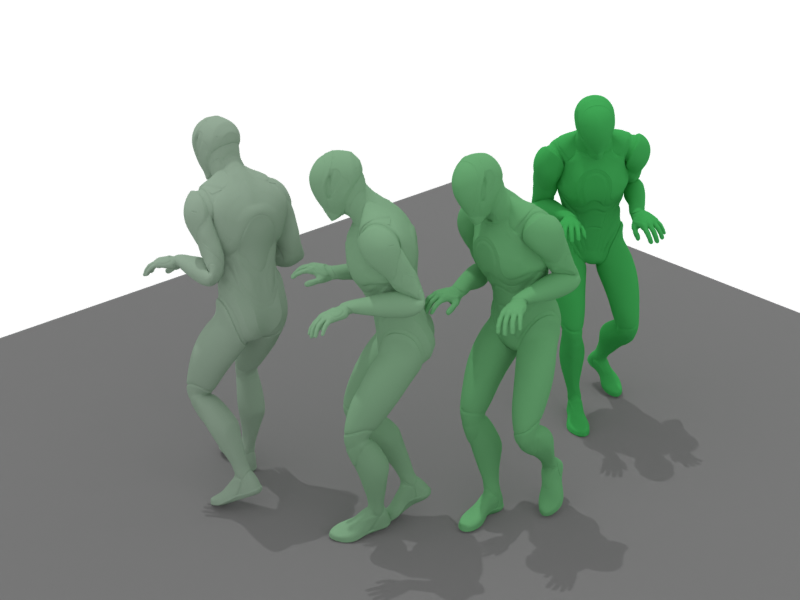}
        \caption{\rev{SASI, ``backpedaling''}}
    \end{subfigure}

    \vspace{4pt}

    \begin{subfigure}{0.48\linewidth}
        \centering
        \includegraphics[height=3.9cm]{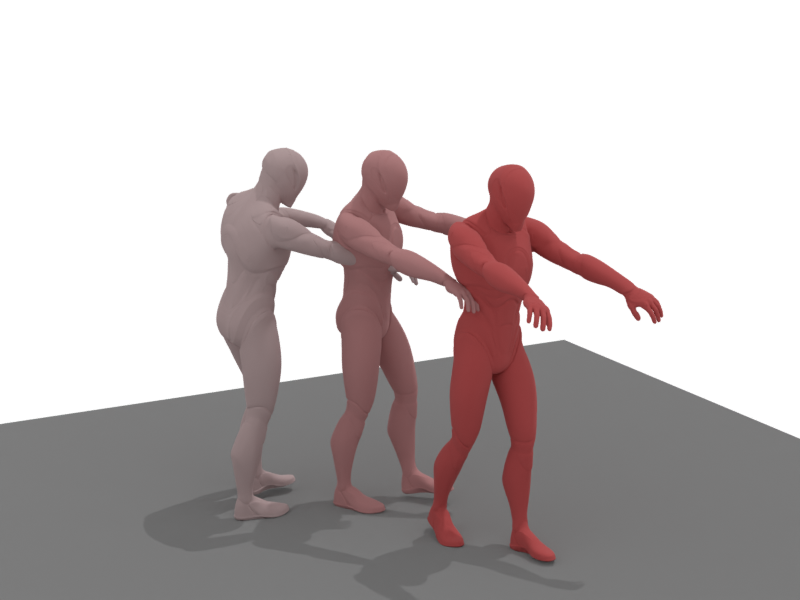}
        \caption{\rev{\textit{Zombie} style reference}}
    \end{subfigure}
    \hfill
    \begin{subfigure}{0.48\linewidth}
        \centering
        \includegraphics[height=3.9cm]{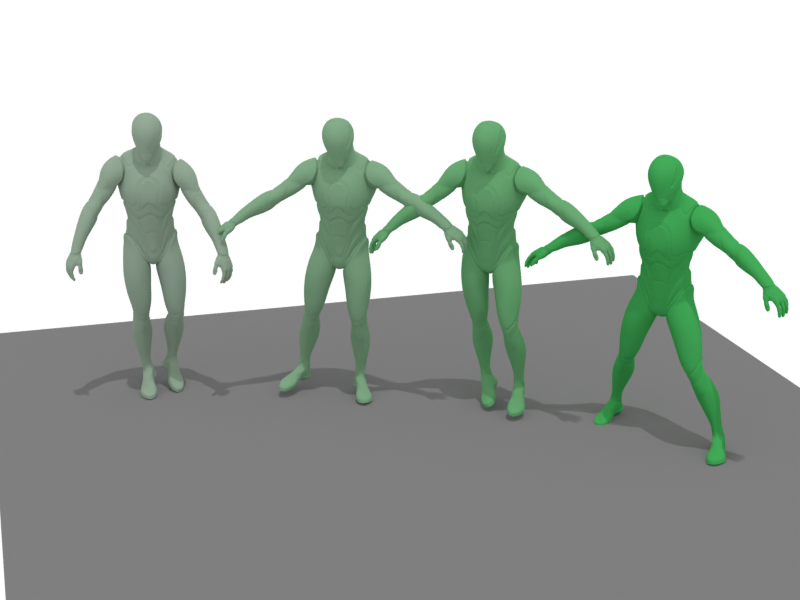}
        \caption{\rev{SASI, ``side hop to left''}}
    \end{subfigure}

    \vspace{4pt}

    \begin{subfigure}{0.48\linewidth}
        \centering
        \includegraphics[height=3.9cm]{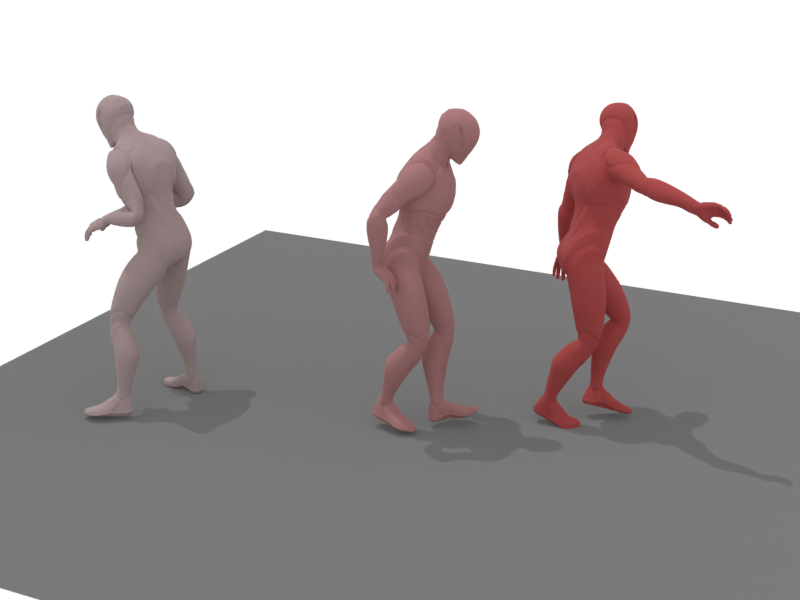}
        \caption{\rev{Style references, \textit{Dinosaur} $\rightarrow$ \textit{Chicken} $\rightarrow$
        \textit{Aeroplane}}}
    \end{subfigure}
    \hfill
    \begin{subfigure}{0.48\linewidth}
        \centering
        \includegraphics[height=3.9cm]{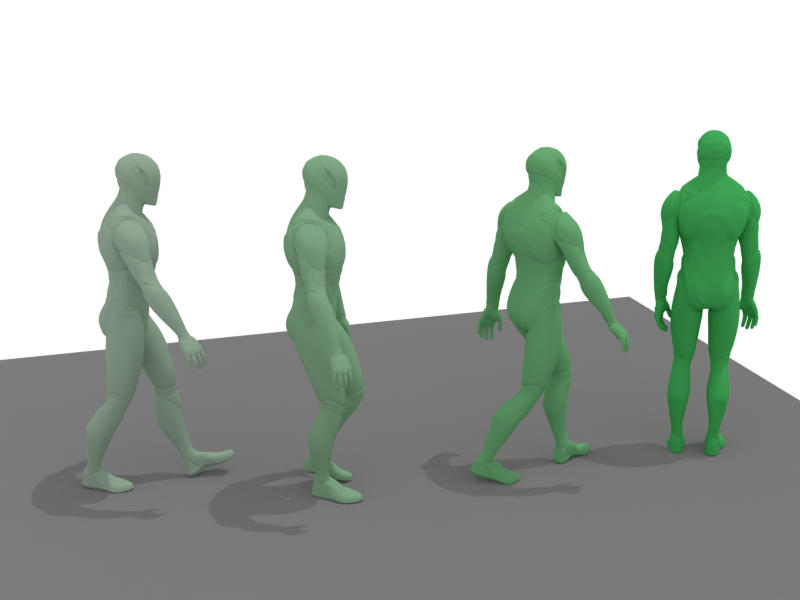}
        \caption{\rev{SASI, time-varying style}}
    \end{subfigure}
    \caption{\rev{Motions generated by the SASI variant, ordered from the most to the least successful
    transfer. In (b) the style is transferred faithfully: the crouch, the bent knees and the bent arms
    held in front of the chest of \textit{Dinosaur} (a) all appear, over the backpedal asked for by the
    prompt. In (d) the requested action is produced but the style is not: the arms hang at the sides
    and the torso stays upright, with none of the stiff forward-extended arms of \textit{Zombie} (c).
    (f) is the time-varying setting, for the prompt ``walking forward, then stopping and turning
    left'', with the three references of (e) applied in sequence; the output walks, stops and turns as
    asked, but stays close to a neutral style throughout and none of the three transitions becomes
    visible. The single-style cases (b, d) are consistent with the top-1
    \textit{SRA} of $0.6719$ reported for SASI in Table~\ref{tab:ablation_single_style}, and the third with the
    $70.3\%$ of time-varying sequences in which the requested switch never becomes detectable.
    Colors transition from light to dark to indicate earlier to later frames.}}
    \label{fig:sasi}
\end{figure}

\rev{\subsection{Style injection with SASI}\label{sup:vis-sasi}}
\rev{Figure~\ref{fig:sasi} shows the SASI variant of the injection-mechanism ablation in
Section~\ref{sec:ablation}, in which our SAM is replaced by the module of Wu \etal\ and retrained on the same
backbone, data and schedule. The three examples illustrate the behaviour behind the numbers reported
there: SASI transfers a single style well when it succeeds, but does not always pick the style up,
and it does not confine a reference to the interval it was assigned to when the style varies over
time.}

\begin{figure}[ht]
    \centering
    \revcolor
    \begin{subfigure}{0.48\linewidth}
        \centering
        \includegraphics[height=3.9cm]{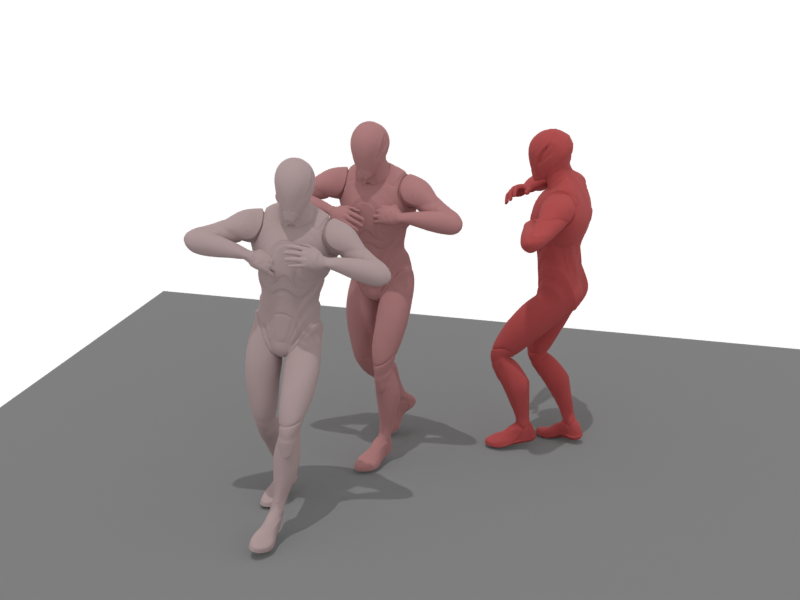}
        \caption{\rev{\textit{BeatChest} style reference}}
    \end{subfigure}
    \hfill
    \begin{subfigure}{0.48\linewidth}
        \centering
        \includegraphics[height=3.9cm]{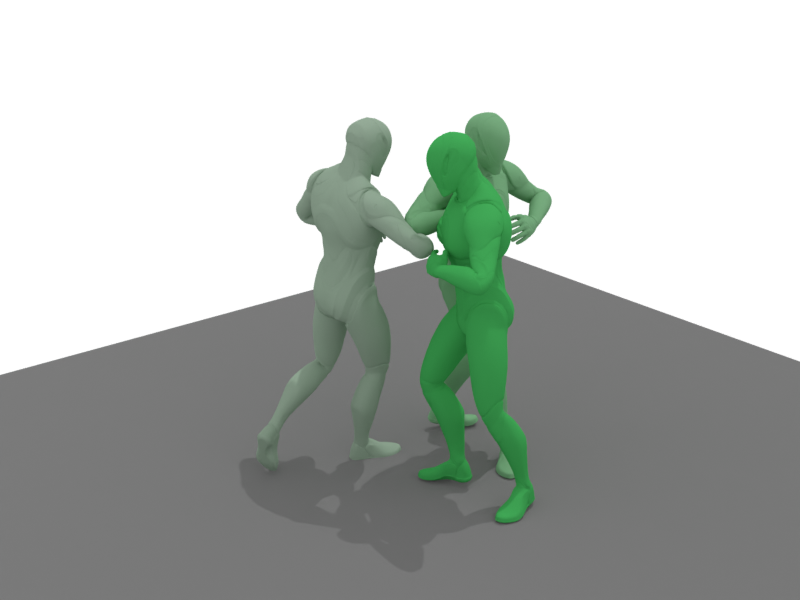}
        \caption{\rev{Ours}}
    \end{subfigure}
    \caption{\rev{A failure case on a fast in-domain style, for the text prompt
    \textit{``walking clockwise in a circle''}. The arm pose of the \textit{BeatChest} reference (a)
    changes sharply between the frames shown, from hands drawn in at the chest to elbows lifted and
    hands raised towards the face; the speed of that alternation is what carries the style. Our
    output (b) does move the arms, swinging them in and up towards the chest along much the same path,
    but gently: the trajectory is reproduced and the pace is not, so what is lost is the speed and
    amplitude of the striking motion rather than the motion itself. The walking
    content asked for by the prompt is still produced. This is the over-smoothing described in
    Section~\ref{sec:failure}, and it is a limitation of what the diffusion backbone
    reproduces rather than of what the style encoder recognizes. Colors transition from light to dark
    to indicate earlier to later frames.}}
    \label{fig:walk_beatchest}
\end{figure}

\rev{\subsection{Rapid or abrupt style references}\label{sup:vis-abrupt}}
\rev{Figure~\ref{fig:walk_beatchest} illustrates the second of the two failure modes discussed in
Section~\ref{sec:failure}. Unlike the references of
Section~\ref{sup:vis-ood}, \textit{BeatChest} is a 100STYLE style and is represented in the style
latent space; what the model fails to reproduce is its speed rather than its identity.}

\rev{\section{Temporal style transfer evaluation}\label{sup:localization}}

\rev{
Sequence-level \textit{SRA} cannot tell whether a style is confined to the interval it was assigned
to: a model that applies the first reference to the entire sequence already collects half of it.
We therefore evaluate temporal style transfer directly, on the $200$-frame protocol of Section~\ref{subsec:time-varying_style} with a single style switch at the mid-point $b$, and report both the $63$ in-domain and the $6$
out-of-domain styles in Table~\ref{tab:localization}. Three metric groups are used.
\textit{Segment SRA} classifies $[0,b)$ and $[b,N)$ separately, each against its own reference's
label. \textit{Leak} is the probability mass the classifier assigns to the neighbouring segment's
style inside a segment, normalized as $p_{\mathrm{other}}/(p_{\mathrm{own}}+p_{\mathrm{other}})$ and
averaged over both segments, so that $0.5$ means the two styles are indistinguishable within a
segment and sample pairs drawing the same style label are excluded.
\textit{Delay} is the signed offset between $b$ and the first sliding window ($41$ frames, stride
$4$) in which reference B outscores reference A, with sequences where B never wins counted in
\textit{Fail} instead of being averaged into the delay; for calibration, a hard cut between the two
reference clips is detected at $+2$ frames. The \textit{Boundary} columns give the peak velocity,
acceleration and jerk within $\pm3$ frames of $b$, each divided by that sequence's own median, so
the target is not $1$ but the value a genuine style change exhibits, given by the \textit{Real} row.

FlexMoGen is the only method whose two intervals score alike (in-domain, $0.5380$ against $0.5352$);
both baselines lose $0.17$ to $0.34$ between the first and the second interval, that is, they keep
producing the first style after the switch was requested. Leakage makes the mechanism explicit: the
baselines sit at $\approx0.49$, exactly the value obtained when a single style covers both intervals,
whereas ours is $0.1064$, i.e. within one of our segments the assigned style is roughly nine times as
likely as its neighbour, while for the baselines the two are indistinguishable.
The switch is detectable in $90.6\%$ of our sequences and lands within $1.5$ frames
($0.05$s) of where it was requested, against failure rates of $34.4\%$ and $53.4\%$ and offsets of
$6$ to $14$ frames for the baselines.

These margins carry over to the out-of-domain styles: our leakage ($0.1017$) and delay ($-1.45$) are
unchanged within their confidence intervals, and our failure rate drops to $6.8\%$, whereas SMooDi
keeps both intervals near chance. T2M+MP attains the nominally best
first-interval \textit{SRA} in this block ($0.4788$ against our $0.4620\pm0.0577$, a difference well
inside the interval), but it obtains it by copying poses from the reference, which is also why its
second interval still falls to $0.3528$ and its leakage stays at $0.4916$.

The boundary columns are a sanity check rather than a ranking. Each peak is normalized by the
sequence's own median, so what matters is the absence of an extreme value, not proximity to any
particular number: all three methods stay between $0.84$ and $1.10$ times the corresponding
\textit{Real} value in both blocks, and the columns therefore do not separate them. What they do
establish is that our earlier and more reliable switch is not bought with a discontinuity, our peaks
being at $0.96$ to $1.03$ times those of a genuine style change. They also have to be read together
with the failure rates, since a method that never switches is trivially continuous: SMooDi's
consistently lowest peaks ($0.84$ to $0.95$) reflect the absence of a transition rather than a
smoother one. The artifact the boundary window does expose is foot sliding, on which T2M+MP skates
in $82.8\%$ of boundary frames against $12.2\%$ for ours (in-domain).
}

\begin{table}
\revcolor
\centering
\small
\setlength{\tabcolsep}{4pt}
\caption{\rev{Temporal style transfer evaluation on time-varying style input, for the $63$ in-domain
and the $6$ out-of-domain styles, averaged over five evaluation repeats. Segment-wise \textit{SRA} is
top-1. \textit{Leak} measures style bleeding between the two intervals, and \textit{Delay} /
\textit{Fail} when and whether the requested switch happens at all. The boundary
columns are peak velocity, acceleration and jerk at the switch, normalized by each sequence's own
median; they are a sanity check on transition smoothness rather than a ranked metric, since any value
in the range of the \textit{Real} row is acceptable and only an extreme peak would indicate a visible
discontinuity. The $95\%$ confidence interval is at most $0.046$ for the in-domain columns and $0.21$
for the out-of-domain ones; for \textit{Delay} it reaches $0.7$ and $3.4$ frames respectively. We use
\textbf{bold} for the best score in the ranked columns, within each block.}}
\label{tab:localization}
\small
\begin{tabular*}{\linewidth}{@{\extracolsep{\fill}}lcccccccc@{}}
\toprule
\multirow{2}{*}{\textbf{Models}}
& \multicolumn{2}{c}{\textbf{Segment SRA$\uparrow$}}
& \multirow{2}{*}{\textbf{Leak$\downarrow$}}
& \multicolumn{2}{c}{\textbf{Transition}}
& \multicolumn{3}{c}{\textbf{Boundary (sanity check)}} \\
\cmidrule(lr){2-3} \cmidrule(lr){5-6} \cmidrule(lr){7-9}
& \textbf{Seg. 1} & \textbf{Seg. 2} &
& \textbf{Delay} & \textbf{Fail$\downarrow$}
& $\Delta v$ & $\Delta a$ & $\Delta j$ \\
\midrule
\multicolumn{9}{c}{\textit{In-domain styles} ($63$)} \\
\midrule
Real & 0.9993 & 0.9979 & 0.0003 & --- & --- & 1.3123 & 1.6530 & 1.8206 \\
T2M+MP & 0.5146 & 0.3403 & 0.4905 & $-6.21$ & 0.3441 & 1.2588 & 1.8210 & 1.8281 \\
SMooDi & 0.2908 & 0.1629 & 0.4982 & $-14.12$ & 0.5339 & 1.1318 & 1.3854 & 1.6817 \\
FlexMoGen (Ours) & \textbf{0.5380} & \textbf{0.5352} & \textbf{0.1064} & $\bm{-1.53}$ & \textbf{0.0943} & 1.2578 & 1.6020 & 1.7392 \\
\midrule
\multicolumn{9}{c}{\textit{Out-of-domain styles} ($6$)} \\
\midrule
Real & 1.0000 & 1.0000 & 0.0000 & --- & --- & 1.2702 & 1.6912 & 1.9049 \\
T2M+MP & \textbf{0.4788} & 0.3528 & 0.4916 & $-3.53$ & 0.2789 & 1.2635 & 1.7311 & 1.7901 \\
SMooDi & 0.2205 & 0.2092 & 0.4995 & $-16.62$ & 0.4610 & 1.2043 & 1.4576 & 1.7918 \\
FlexMoGen (Ours) & 0.4620 & \textbf{0.5143} & \textbf{0.1017} & $\bm{-1.45}$ & \textbf{0.0681} & 1.3059 & 1.6853 & 1.8102 \\
\bottomrule
\end{tabular*}\end{table}

\rev{\section{Long sequences with time-varying style}\label{sup:long-multistyle}}

\rev{The two evaluations in the main paper each vary one factor. Section~\ref{subsec:long_stylized_t2m} extends the sequence to
$400$ frames but applies a single style throughout; Section~\ref{subsec:time-varying_style} varies the style over time but fixes
the output at $200$ frames, the maximum training length. Neither of them shows that the two hold
together. This section repeats the time-varying protocol at $200$, $400$ and $800$ frames --- $6.7$,
$13.3$ and $26.7$ seconds at $30$\,fps --- changing nothing else.}

\rev{The protocol is that of Section~\ref{subsec:time-varying_style}: two style references per sequence, drawn from the six
out-of-domain styles of Table~\ref{tab:classification_styles}, the first assigned to the interval
$[0,b)$ and the second to $[b,N)$ with the boundary $b$ at the mid-point, evaluated over five repeats
of $320$ samples each. Because the two segments split the sequence, their duration scales with $N$:
$100$ frames each at $N=200$, and $400$ frames each at $N=800$, so at the longest setting a single
style interval is already twice the entire training length.
Table~\ref{tab:long-multistyle-seq} reports the sequence-level metrics and
Table~\ref{tab:long-multistyle-seg} the segment-level and transition measurements defined in
Section~\ref{sup:localization}. Both give our model only: Table~\ref{tab:varying_style_metric} and
Section~\ref{sup:localization} already establish that the baselines do not localize style at $200$ frames, and a localization that is absent at the training
length has nothing for a longer sequence to preserve.}

\begin{table}[ht]
\revcolor
\centering
\small
\setlength{\tabcolsep}{4pt}
\caption{\rev{Sequence-level metrics for time-varying style input at three output lengths, for the six
out-of-domain styles, averaged over five evaluation repeats. Each sequence carries two style
references split at the mid-point, so the duration of an individual style interval is half the value
in the \textit{Length} column. The $95\%$ confidence interval over the repeats is at most $0.46$ for
\textit{FID}, $0.10$ for \textit{MM Dist} and $0.024$ for every remaining column.}}
\label{tab:long-multistyle-seq}
\footnotesize
\setlength{\tabcolsep}{2pt}
\begin{tabular*}{\linewidth}{@{\extracolsep{\fill}}clcccccccc@{}}
\toprule
\multirow{2}{*}{\textbf{Length}}
& \multirow{2}{*}{\textbf{Model}}
& \multirow{2}{*}{\textbf{MM Dist$\downarrow$}}
& \multirow{2}{*}{\textbf{R-precision$\uparrow$}}
& \multirow{2}{*}{\textbf{FID$\downarrow$}}
& \multirow{2}{*}{\textbf{CLIP score$\uparrow$}}
& \multicolumn{3}{c}{\textbf{SRA$\uparrow$}}
& \multirow{2}{*}{\textbf{Foot Skating Ratio$\downarrow$}} \\
\cmidrule(lr){7-9}
& & & & & & \textbf{Top-1} & \textbf{Top-2} & \textbf{Top-3} & \\
\midrule
\multirow{2}{*}{$200$ ($6.7$\,s)}
 & Real             & 1.2474 & 0.9525 & 0.2720 & 0.7468 & 0.9994 & 1.0000 & 1.0000 & 0.0001 \\
 & FlexMoGen (Ours) & 3.1295 & 0.6613 & 2.0563 & 0.5855 & 0.5284 & 0.6634 & 0.7309 & 0.0890 \\
\midrule
\multirow{2}{*}{$400$ ($13.3$\,s)}
 & Real             & 1.0910 & 0.9569 & 0.2872 & 0.7622 & 0.9988 & 1.0000 & 1.0000 & 0.0001 \\
 & FlexMoGen (Ours) & 3.0377 & 0.6731 & 2.4010 & 0.5971 & 0.5466 & 0.6688 & 0.7312 & 0.0842 \\
\midrule
\multirow{2}{*}{$800$ ($26.7$\,s)}
 & Real             & 1.0982 & 0.9563 & 0.2687 & 0.7688 & 0.9975 & 1.0000 & 1.0000 & 0.0001 \\
 & FlexMoGen (Ours) & 2.9750 & 0.6881 & 2.0165 & 0.6235 & 0.5053 & 0.6478 & 0.7178 & 0.0856 \\
\bottomrule
\end{tabular*}
\end{table}

\begin{table}[ht]
\revcolor
\centering
\small
\setlength{\tabcolsep}{4pt}
\caption{\rev{Temporal style transfer evaluation at three output lengths, for the same protocol and
the same metrics as Table~\ref{tab:localization}. \textit{Bnd.\ FS} is the foot-skating ratio restricted
to the frames around the boundary. The boundary columns are peak velocity, acceleration and jerk at
the switch, normalized by each sequence's own median; as before they are a sanity check
rather than a ranked metric, the target being the range of the \textit{Real} row. The $95\%$
confidence interval is at most $0.032$ for the segment \textit{SRA} and boundary columns, $0.012$ for
\textit{Leak}, $0.016$ for \textit{Bnd.\ FS}, $0.025$ for \textit{Fail}, and $1.15$ frames for
\textit{Delay}.}}
\label{tab:long-multistyle-seg}
\footnotesize
\setlength{\tabcolsep}{2pt}
\begin{tabular*}{\linewidth}{@{\extracolsep{\fill}}clccccccccc@{}}
\toprule
\multirow{2}{*}{\textbf{Length}}
& \multirow{2}{*}{\textbf{Model}}
& \multicolumn{2}{c}{\textbf{Segment SRA$\uparrow$}}
& \multirow{2}{*}{\textbf{Leak$\downarrow$}}
& \multicolumn{2}{c}{\textbf{Transition}}
& \multicolumn{3}{c}{\textbf{Boundary (sanity check)}}
& \multirow{2}{*}{\textbf{Bnd. FS$\downarrow$}} \\
\cmidrule(lr){3-4} \cmidrule(lr){6-7} \cmidrule(lr){8-10}
& & \textbf{Seg. 1} & \textbf{Seg. 2} &
& \textbf{Delay} & \textbf{Fail$\downarrow$}
& $\Delta v$ & $\Delta a$ & $\Delta j$ & \\
\midrule
\multirow{2}{*}{$200$ ($6.7$\,s)}
 & Real             & 1.0000 & 0.9988 & 0.0000 & ---      & ---    & 1.3426 & 1.6888 & 1.8934 & 0.0000 \\
 & FlexMoGen (Ours) & 0.5237 & 0.5331 & 0.1155 & $-1.17$  & 0.1026 & 1.2514 & 1.5735 & 1.7120 & 0.1339 \\
\midrule
\multirow{2}{*}{$400$ ($13.3$\,s)}
 & Real             & 0.9981 & 0.9994 & 0.0001 & ---      & ---    & 1.2891 & 1.6796 & 1.8886 & 0.0005 \\
 & FlexMoGen (Ours) & 0.5444 & 0.5487 & 0.1170 & $-1.02$  & 0.1047 & 1.2223 & 1.5584 & 1.6879 & 0.1322 \\
\midrule
\multirow{2}{*}{$800$ ($26.7$\,s)}
 & Real             & 0.9981 & 0.9969 & 0.0001 & ---      & ---    & 1.3140 & 1.6799 & 1.8650 & 0.0004 \\
 & FlexMoGen (Ours) & 0.5094 & 0.5012 & 0.1238 & $+0.30$  & 0.1025 & 1.2210 & 1.5506 & 1.6890 & 0.1244 \\
\bottomrule
\end{tabular*}
\end{table}

\rev{Nothing degrades with length. \textit{FID} is $2.06$, $2.40$ and $2.02$ at the three lengths,
flat within confidence intervals of $\pm0.38$ to $\pm0.46$, and the foot-skating ratio stays between
$0.084$ and $0.089$ throughout. The content metrics do improve slightly as the sequence lengthens
(\textit{MM Dist} $3.1295 \to 2.9750$, \textit{CLIP score} $0.5855 \to 0.6235$), but so does the
\textit{Real} row ($1.2474 \to 1.0982$ and $0.7468 \to 0.7688$), so this reflects how the T2M
evaluator behaves on longer clips rather than a gain of our model; what stays constant is the gap
between the two rows.}

\rev{Localization is what the question turns on, and it is unaffected. At every length the two
intervals score alike --- $0.5237$ against $0.5331$ at $200$ frames, $0.5444$ against $0.5487$ at
$400$, and $0.5094$ against $0.5012$ at $800$ --- so the model does not fall back on holding the first
style, which is the failure mode both baselines exhibit in Table~\ref{tab:localization}, where they lose
$0.17$ to $0.34$ between the two intervals already at $200$ frames. Leakage stays between $0.1155$ and
$0.1238$, against the $\approx0.49$ of the baselines, and the failure rate is unchanged at $0.102$ to
$0.105$. The switch still lands within $1.2$ frames of where it was requested at all three lengths,
even when the two intervals it separates are $400$ frames long. The boundary columns stay just below
the corresponding \textit{Real} values throughout, exactly as they do at $200$ frames in Table~\ref{tab:localization}, so the switch does not turn into a discontinuity as the sequence grows, and the boundary
foot-skating ratio is likewise flat.}

\rev{The one cost appears in style fidelity at the longest setting: top-1 \textit{SRA} falls from
$0.5466$ at $400$ frames to $0.5053$ at $800$, slightly outside the confidence intervals, and the two
segment-wise values fall with it. Top-3 \textit{SRA} is flat ($0.7309$, $0.7312$, $0.7178$). The style
is therefore still applied, and still applied to the interval it was assigned to, but over the longest
sequences the classifier's first choice is marginally less often the requested style.}

\rev{One part of the setting is not varied here: the number of transitions per sequence stays at one.
What varies with $N$ is the duration of each style interval, from $100$ to $400$ frames. Sequences
containing several short intervals in succession are not covered by this evaluation.}

\rev{\section{User study}\label{sup:user-study}}

Because the perceptual quality of generated motions is inherently subjective, we conducted a user study to complement our quantitative evaluation.
We generated 20 motion samples using diverse content prompts and style examples; for each sample, all models generated a motion conditioned on the same prompt and style reference.
Participants were presented with side-by-side animated renderings and asked to perform pairwise comparisons between our model and each baseline—T2M+MP, LoRA-MDM, and SMooDi—across three criteria: content preservation, style reflection, and overall motion quality.
For each participant, 15 samples were randomly selected, yielding 330 total responses from 22 participants.

\begin{figure}[t]
    \centering
    \includegraphics[width=0.62\linewidth]{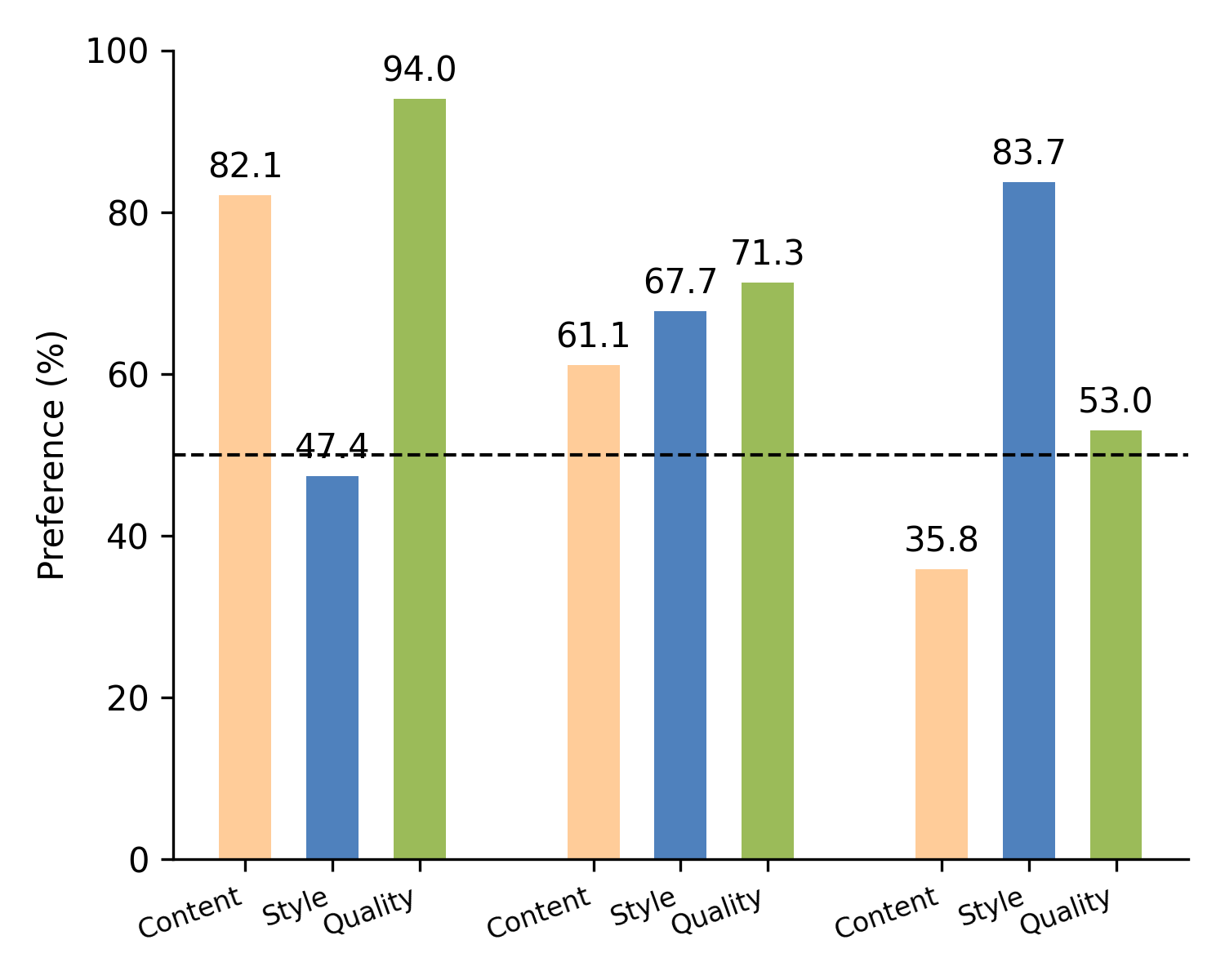}
    \caption{User study results (pairwise preference, \% favoring ours). Left: FlexMoGen vs.\ T2M+MP. Middle: FlexMoGen vs.\ SMooDi. Right: FlexMoGen vs.\ LoRA-MDM. Criteria are content preservation, style reflection, and motion quality.}
    \label{fig:user_study}
\end{figure}

As shown in Figure~\ref{fig:user_study}, FlexMoGen is preferred over all baselines across nearly all criteria.
Against SMooDi, FlexMoGen is preferred on all three axes: content preservation (61.1\%), style reflection (67.7\%), and motion quality (71.3\%).
Against T2M+MP, FlexMoGen achieves substantially higher preference on content (82.1\%) and quality (94.0\%); T2M+MP is marginally preferred on style (47.4\% for ours) because it largely copies poses from the reference clip regardless of the content prompt—consistent with its high \textit{SRA} but poor \textit{R-Precision} in Sections~\ref{subsec:long_stylized_t2m} and~\ref{subsec:time-varying_style}.
Against LoRA-MDM, FlexMoGen is strongly preferred on style reflection (83.7\%) and comparable on motion quality (53.0\%), while LoRA-MDM is preferred on content preservation (35.8\% for ours); this is consistent with LoRA-MDM's strong content metrics but near-zero \textit{SRA} scores in Table~\ref{tab:metric}, which indicate that it preserves text semantics at the expense of style fidelity.
Overall, the user study corroborates our quantitative findings: FlexMoGen achieves the best balance between content fidelity, style reflection, and motion quality.

\section{Pseudo code}\label{sup:pseudocode}

This section provides the pseudo code for both the training and inference procedures of our FlexMoGen framework.

\begin{algorithm}[H]
\caption{\textbf{FlexMoGen training}}
\label{alg:style_t2m}
\begin{algorithmic}[1]
\Require Pretrained motion encoder $E_m$ and motion decoder $D_m$ (frozen), pretrained T2M backbone $G_\theta$ (frozen), pretrained style VAE $SE$ (frozen), style adaption module (SAM) parameters $\{W_s^{K,\ell}, W_s^{V,\ell}\}_{\ell=1}^L$, noise schedule $\{\beta_t\}_{t=1}^T$
\Require Training batch of content motions $x_c$, style motions $x_s$, text prompts $c$, and diffusion steps $T$.
\While{not converged}
    \State Sample $(x_c, x_s, c)$ from training data
    \State Encode content motion: $z_0 \gets E_m(x_c)$
    \State Sample diffusion step $t \sim \mathcal{U}\{1,\dots,T\}$ and noise $\epsilon \sim \mathcal{N}(0, I)$
    \State Add noise to latent: $z_t \gets \sqrt{\bar{\alpha}_t} \, z_0 + \sqrt{1-\bar{\alpha}_t}\,\epsilon$
    \State Encode text: $e_c \gets \text{TextEncoder}(c)$
    \State Encode style example: $(\mu_s, \sigma_s) \gets SE(x_s)$; \quad and sample $s \sim \mathcal{N}(\mu_s, \sigma_s)$
    \For{each transformer layer $\ell = 1 \dots L$ in $G_\theta$}
        \State Compute queries, keys, values (content-only):
        \State \hspace{0.8em}$Q^\ell \gets W_Q^\ell h^\ell,\quad K^\ell \gets W_K^\ell h^\ell,\quad V^\ell \gets W_V^\ell h^\ell$
        \State Inject style via SAM:
        \State \hspace{0.8em}$\Delta K^\ell \gets W_s^{K,\ell} s,\quad \Delta V^\ell \gets W_s^{V,\ell} s$
        \State \hspace{0.8em}$\tilde{K}^\ell \gets K^\ell + \Delta K^\ell,\quad \tilde{V}^\ell \gets V^\ell + \Delta V^\ell$
        \State Apply self-attention with relative position encoding:
        \State \hspace{0.8em}$h^{\ell+1} \gets \text{Attn}(Q^\ell, \tilde{K}^\ell, \tilde{V}^\ell)$
    \EndFor
    \State Predict clean latent: $\hat{z}_0 \gets G_\theta(z_t, t, e_c, s)$
    \State Compute reconstruction loss:
    \State $\mathcal{L}_{\text{recon}} \gets \Vert \hat{z}_0 - z_0 \Vert$
    \State Compute style preservation loss:
    \State $\mathcal{L}_{\text{sty}} \gets \mathcal{L}_{KL}((\mu_{\hat{x}}, \sigma_{\hat{x}}), (\mu_s, \sigma_s))$,
    \State where $(\mu_{\hat{x}}, \sigma_{\hat{x}}) \gets SE(D_m(\hat{z}_0))$
    \State Compute the total loss:
    $\mathcal{L} \gets \mathcal{L}_{\text{recon}} + \lambda_{\text{sty}} \mathcal{L}_{\text{sty}}$
    \State Update only SAM parameters $\{W_s^{K,\ell}, W_s^{V,\ell}\}$  via backpropagation
\EndWhile
\end{algorithmic}
\end{algorithm}
\clearpage
\begin{algorithm}[H]
\caption{\textbf{FlexMoGen inference (stylized text-to-motion generation)}}
\label{alg:cosmo_inference}
\begin{algorithmic}[1]
\Require Pretrained motion decoder $D_m$, pretrained T2M backbone $G_\theta$, pretrained style VAE $SE$, SAM parameters $\{W_s^{K,\ell}, W_s^{V,\ell}\}_{\ell=1}^L$, noise schedule $\{\beta_t\}_{t=1}^T$
\Require Text prompt $c$, style motion $x_s$, number of diffusion steps $T$
\State Encode text: $e_c \gets \text{TextEncoder}(c)$
\State Encode style example: $(\mu_s, \sigma_s) \gets SE(x_s)$
\State Set deterministic style code: $s \gets \mu_s$
\State Initialize latent with Gaussian noise: $z_T \sim \mathcal{N}(0, I)$
\For{$t = T, T{-}1, \dots, 1$}
    \State Set $h^0 \gets z_t$ (with time and text embeddings fused into $h^0$)
    \For{each Transformer layer $\ell = 1 \dots L$ in $G_\theta$}
        \State Compute queries, keys, values (content-only):
        \State \hspace{0.8em}$Q^\ell \gets W_Q^\ell h^\ell,\quad K^\ell \gets W_K^\ell h^\ell,\quad V^\ell \gets W_V^\ell h^\ell$
        \State Inject style via SAM:
        \State \hspace{0.8em}$\Delta K^\ell \gets W_s^{K,\ell} s,\quad \Delta V^\ell \gets W_s^{V,\ell} s$
        \State \hspace{0.8em}$\tilde{K}^\ell \gets K^\ell + \Delta K^\ell,\quad \tilde{V}^\ell \gets V^\ell + \Delta V^\ell$
        \State Apply self-attention (with relative position encoding inside $\text{Attn}$):
        \State \hspace{0.8em}$h^{\ell+1} \gets \text{Attn}(Q^\ell, \tilde{K}^\ell, \tilde{V}^\ell)$
    \EndFor
    \State Predict clean latent: $\hat{z}_0 \gets G_\theta(z_t, t, e_c, s)$
    \State Update latent using a DDIM sampler :
    \State \hspace{0.8em}$z_{t-1} \gets \text{SampleStep}(z_t, \hat{z}_0, t)$
\EndFor
\State Decode motion from final latent:
\State \hspace{0.8em}$\hat{x} \gets D_m(z_0)$
\State \Return stylized motion $\hat{x}$
\end{algorithmic}
\end{algorithm}

\section{Implementation details}\label{sup:impl}

All models were trained on a single \textit{NVIDIA RTX 6000 Ada Generation} GPU. The text-to-motion model was trained for 60{,}000 iterations, and the style encoder was trained for 40{,}000 iterations, each requiring approximately 10 hours. The Style Adaptation Module (SAM) was trained for 20{,}000 iterations using \textit{bfloat16} precision and took around 8 hours.

The diffusion model was trained with 1000 steps, and we sample with 100 steps using DDIM.

\rev{
\subsection{Parameter count of SAM}
\label{sup:params}
The style code is $d_s=6\times64=384$ dimensional (five body parts and a root component, Section~\ref{pretraining_se}), and the frozen backbone has width $d_\text{model}=512$ with $L=7$ attention layers.
Each of $W_s^{K}$ and $W_s^{V}$ is learned separately at every layer, and each is realized as a
projection $d_s\!\to\!d_\text{model}$ followed by a zero-initialized gate
$d_\text{model}\!\to\!d_\text{model}$, so that SAM leaves the pretrained model unchanged at
initialization. Counting the biases of both linear maps, the trainable total is
\begin{equation*}
2L\left(d_s d_\text{model} + d_\text{model}^{2} + 2d_\text{model}\right) = 6{,}436{,}864 ,
\end{equation*}
that is $6.44$M parameters. The backbone contributes $14{,}330{,}855$ learned parameters, all of them
frozen, together with $5.12$M entries of fixed sinusoidal position tables that are never learned. SAM
is therefore $31\%$ of the $20.77$M parameters of the deployed model, and it is the only part of it
that is trained: the text-to-motion backbone, the motion VAE and the style encoder are all used exactly
as pretrained. For comparison, a ControlNet-style branch that duplicates the whole backbone trains
$16.1$M parameters, about $2.5$ times as many as SAM, and the SASI adapter trains $14.25$M:
$11.18$M in its per-layer text cross-attention, injection and gate blocks, and a further $3.06$M in
the temporal style encoder it learns in the same stage, where SAM instead reuses a frozen
$0.48$M style encoder.
}

\section{Inference time}\label{sup:inference}

We also compare the inference time of our model against baseline methods. As shown in Table~\ref{tab:infernce}, our approach achieves the shortest inference time among all models.
\begin{table}[ht]
    \centering
    \begin{tabular}{cc}
    \toprule
       \textbf{Models}  &  \textbf{Inference time (s)}\\
       \midrule
       T2M+MP  & 0.08 \\
       LoRA & 0.26 \\
       SMooDi & 0.16 \\
       Ours & 0.07 \\
       \bottomrule
    \end{tabular}
    \caption{Inference time comparison among our model and the baselines.}
    \label{tab:infernce}
\end{table}

\section{Dataset}
\label{sup:dataset}
The details on the datasets are as follows:
\begin{table}[ht]
\centering
\begin{tabular}{lccc}
\toprule
\textbf{Dataset name} & \textbf{Number of clips} & \textbf{Number of hours} & \textbf{Number of annotations}\\
\midrule
Internal dataset &  3089 & 6.4 & 7226 \\
\textit{100STYLE} & 16391  &  20.0 & 37118 \\
\bottomrule
\end{tabular}
\end{table}

Some examples of activities of the dataset include
\begin{table}[ht]
    \centering
    \begin{tabularx}{\linewidth}{c|X}
    \toprule
        Categories & Details \\
        \midrule
        Workout routines & hopping, squatting, skipping, jogging, push-ups, jumps, cartwheel \\
        Locomotion & Idling, spinning, walking / running forward / backward / sideways, turning left / right \\
        Transition & walking / running to a stop, from walking to running, starting to walk / run \\
    \bottomrule
    \end{tabularx}
    \label{tab:placeholder}
\end{table}

The chosen 63 styles and 6 out-of-distribution styles are as follows:
\begin{table}[H]
    \centering
    \caption{Classification of all styles for training and evaluation}
    \label{tab:classification_styles}
    \begin{tabularx}{\linewidth}{c|X}
    \toprule
       \textbf{Categories}  & \textbf{Styles} \\
       \midrule
       63 styles for training  & Aeroplane, Akimbo, Angry, ArmsAboveHead, ArmsBehindBack, ArmsBySide, ArmsFolded, Balance, BentForward, BentKnees, Cat, Chicken, Crouched, Depressed, Dinosaur, DragLeftLeg, DragRightLeg, DuckFoot, Elated, FairySteps, Flapping, GracefulArms, HandsBetweenLegs, HandsInPockets, Heavyset, HighKnees, LawnMower, LegsApart, March, Monk, Morris, Neutral, Old, OnHeels, OnPhoneLeft, OnPhoneRight, OnToesBentForward, OnToesCrouched, PendulumHands, Penguin, PigeonToed, Proud, Quail, RaisedLeftArm, RaisedRightArm, Roadrunner, Robot, Rocket, Rushed, ShieldedLeft, ShieldedRight, Star, Stiff, Superman, Swimming, SwingArmsRound, SwingShoulders, Teapot, WalkingStickLeft, WalkingStickRight, WhirlArms, WideLegs, Zombie\\
       \midrule
       6 extra out-of-distribution styles included in evaluation & BeatChest, FlickLegs, LeanLeft, LeanRight, Lunge, Strutting \\
       \bottomrule
    \end{tabularx}
\end{table}
\clearpage

\section{Skeleton and body-part division}\label{sup:bodypart}

\begin{figure}[ht]
    \centering
    \begin{subfigure}[b]{0.2\linewidth}
        \centering
        \includegraphics[width=\linewidth]{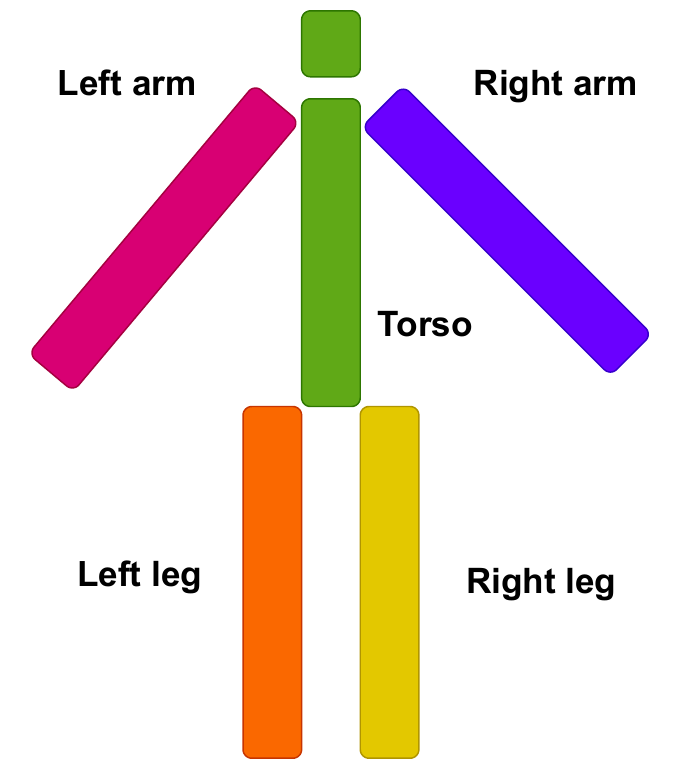}
        \caption{Body part definition}
        \label{fig:body_part_def}
    \end{subfigure}
    \hspace{1cm}
    \begin{subfigure}[b]{0.4\linewidth}
        \centering
        \includegraphics[width=\linewidth]{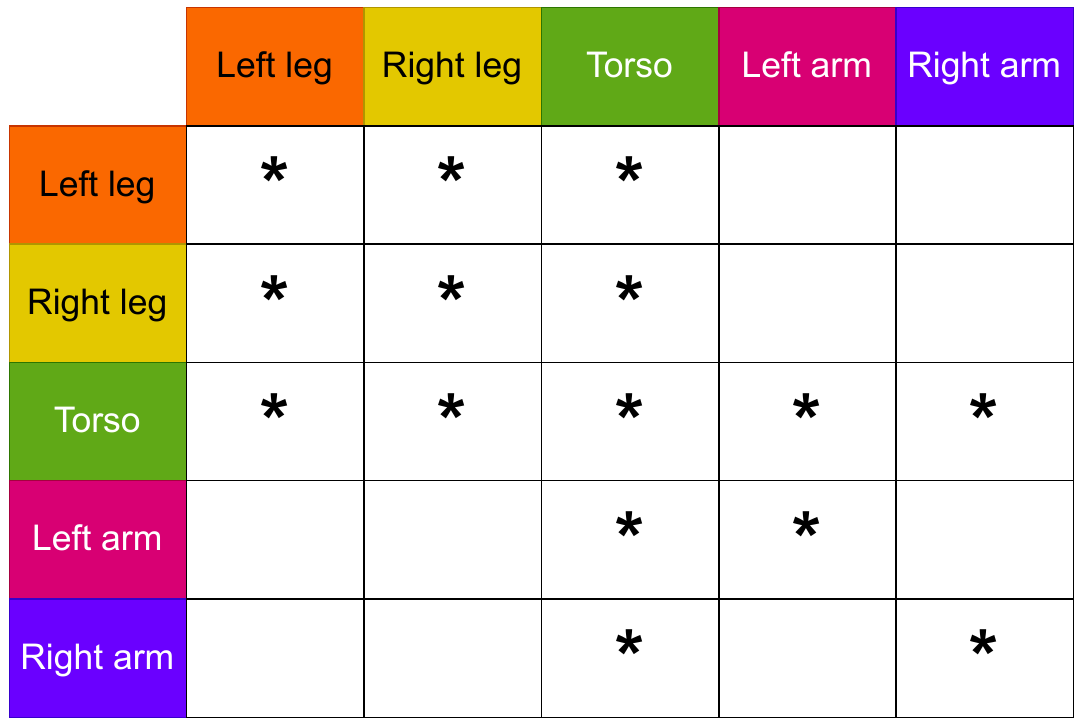}
        \caption{Body part correlation}
        \label{fig:body_part_corr}
    \end{subfigure}
    \caption{(a) Body part definition and (b) body part correlation.}
    \label{fig:body_part_def_corr}
\end{figure}

As described in the main text, we divide the skeleton into five body parts (Figure~\ref{fig:body_part_def}). When computing per-body-part style embeddings, we also mask out interactions between parts that should not influence each other. To achieve this, we employ a learned positional encoding that estimates correlation weights between every pair of body parts, and we mask out connections that are determined to be independent. The resulting part-dependence graph is visualized in Figure~\ref{fig:body_part_corr}.

\begin{figure}[H]
    \centering
    \revcolor
    \begin{subfigure}[t]{0.49\linewidth}
        \centering
        \includegraphics[width=\linewidth]{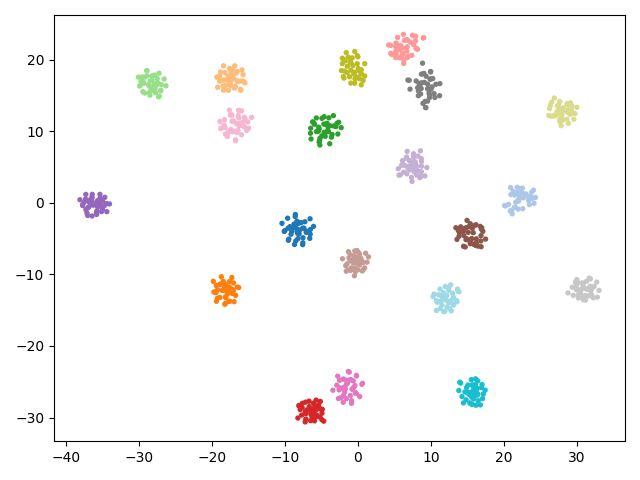}
        \caption{\rev{\textbf{w} per-body-part embedding}}
        \label{fig:t-sne-spatial}
    \end{subfigure}
    \hfill
    \begin{subfigure}[t]{0.49\linewidth}
        \centering
        \includegraphics[width=\linewidth]{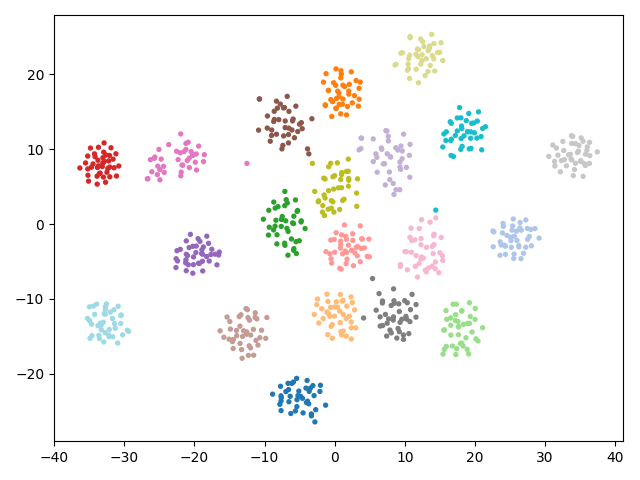}
        \caption{\rev{\textbf{w/o} per-body-part embedding}}
        \label{fig:t-sne-global}
    \end{subfigure}

    \caption{\rev{t-SNE of the style codes of 20 styles, three of them out of distribution. Both
    variants form one cluster per style; the per-part code (a) forms tighter clusters with wider
    margins between them than the variant without the per-part division (b).}}
    \label{fig:t-sne}
\end{figure}

\rev{\subsection{Effect on the style latent space}\label{sup:tsne}}
\rev{Figure~\ref{fig:t-sne} visualizes the style codes of 20 styles, three of them out of
distribution, using t-SNE. Both variants recover one cluster per style, so the per-part division is
not what makes styles separable in the first place. What changes is how cleanly they are separated:
with the per-part code (Figure~\ref{fig:t-sne-spatial}) the clusters are markedly tighter and stand
apart with wide margins, whereas without it (Figure~\ref{fig:t-sne-global}) they are diffuse enough
that several neighbouring styles almost touch. Since this code is what SAM injects at every layer, a
compact, well-separated code leaves less ambiguity between styles at injection time. This is the
mechanism behind the higher style fidelity visible in the supplementary video, and it is also what
keeps the individual part components distinct enough to be recomposed across references
(Section~\ref{sup:vis-bodypart}).}

\end{document}